%% file: ras25focchi.tex
\documentclass[preprint, 5p, times]{style/ras/elsarticle}

\input{includes/used_packages.tex}
\input{includes/acronyms.tex}

\input{includes/math_commands.tex}

\input{includes/newcommands.tex}

\begin{document}


\begin{frontmatter}

  \title{Pseudo-Kinematic Trajectory Control and Planning of Tracked Vehicles}
  \author[1,3]{Michele Focchi\corref{cor1}}

  \author[2]{Daniele Fontanelli}
  \ead{daniele.fontanelli@unitn.it}

  \author[2]{Davide Stocco}
  \ead{davide.stocco@unitn.it}

  \author[1]{Riccardo Bussola}
  \ead{riccardo.bussola00@gmail.com}
  
  \author[1]{Luigi Palopoli}
  \ead{luigi.palopoli@unitn.it}
  \cortext[cor1]{Corresponding author}

  \affiliation[1]{organization={Dipartimento di Ingegneria and Scienza dell'Informazione (DISI), University of Trento},
    addressline={via Sommarive 9},
    postcode={38123},
    city={Trento},
    country={Italy}}

    \affiliation[2]{organization={Dipartimento di Ingegneria Industriale (DII), University of Trento},
    addressline={via Sommarive 9},
    postcode={38123},
    city={Trento},
    country={Italy}}

    \affiliation[3]{organization={Dynamic Legged Systems, Istituto Italiano di Tecnologia (IIT), Genova},
    addressline={via San Quirico 19d},
    postcode={16163},
    city={Genova},
    country={Italy}
}

\begin{abstract}
  Tracked vehicles distribute their weight continuously over a large surface area (the tracks). This distinctive feature makes them the preferred choice for vehicles required to traverse soft and uneven terrain. From a robotics perspective, however, this flexibility comes at a cost: the complexity of modelling the system and the resulting difficulty in designing theoretically sound navigation solutions.
In this paper, we aim to bridge this gap by proposing a framework for the navigation of tracked vehicles, built upon three key pillars.
The first pillar comprises two models: a simulation model and a control-oriented model. The simulation model captures the intricate terramechanics dynamics arising from soil-track interaction and is employed to develop faithful digital twins of the system across a wide range of operating conditions. The control-oriented model is pseudo-kinematic and mathematically tractable, enabling the design of efficient and theoretically robust control schemes.
The second pillar is a Lyapunov-based feedback trajectory controller that provides certifiable tracking guarantees.
The third pillar is a portfolio of motion planning solutions, each offering different complexity-accuracy trade-offs.
The various components of the proposed approach are validated through
extensive simulations and experimental \DSrev{evaluations on two
  different robotic platforms, namely the MAXXII and the LIMO robots}.
\end{abstract}


\end{frontmatter}

\journal{Robotics and Autonomous Systems}


\section*{Supplementary Material}
{
  \begin{itemize}
    \item Video of experimental results is available at:\\
    \href{https://youtu.be/TyeZ0vRkI04}{\textit{youtu.be/TyeZ0vRkI04}}
    \item The code to reproduce the all the figures in Section~\ref{sec:flat_results} and~\ref{sec:slope_results} is open-source and available  at:  \\
    \href{https://github.com/mfocchi/tracked_robot_simulator.git}{\texorpdfstring{https://github.com/mfocchi/tracked\_robot\_simulator.git} }. The dataset of the experiments and the models is available through Git Large File Storage
    facility.\\    
  \end{itemize}
}
\makenomenclature
\nomenclature[00]{m}{Mass of the vehicle}
\nomenclature[01]{${}_bI$}{Tensor of inertia expressed in the body reference frame}
\nomenclature[02]{$I_{zz}$}{Rotational inertia of the vehicle along the $z$-axis}
\nomenclature[03]{${}_b{v}_x$}{Longitudinal velocity of the vehicle \gls{com} in the body reference frame}
\nomenclature[04]{${}_b{v}_y$}{Lateral velocity of the vehicle \gls{com} in the body reference frame}
\nomenclature[05]{${\omega_z}$}{angular velocity about $z$ axis in the body reference frame}
\nomenclature[05]{${}_b\boldsymbol{\omega}$}{angular velocity vector in  body reference frame}
\nomenclature[06]{$\omega_{w,{L/R}}$}{Left/right sprocket wheel angular velocity}
\nomenclature[071]{$r$}{Sprocket wheel radius}
\nomenclature[08]{$B$}{Distance between the tracks' center-lines}
\nomenclature[09]{$\varphi$}{Yaw angle of the base $x$ axis}
\nomenclature[10]{$\varphi_p$}{Yaw angle of a point on a track}
\nomenclature[11]{$t_p$}{time needed for a point of the track to reach a specific coordinate in the vehicle body frame starting from the initial contact with the terrain}
\nomenclature[12]{$x_{p}$}{Longitudinal coordinate in the body reference frame of a point of the left/right track}
\nomenclature[13]{$y_{p}$}{Lateral coordinate in the body reference frame of a point of the left/right track}
\nomenclature[16]{$\delta$}{Angle (with respect to the base $x$ axis) of the shear velocity vector of the infinitesimal track element}
\nomenclature[18]{$\mu$}{Friction coefficient}
\nomenclature[19]{$j_x$}{Shear-displacement in the inertial (world) reference frame along the $x$-axis}
\nomenclature[20]{$j_y$}{Shear-displacement in the  inertial (world) reference frame along the $y$-axis}
\nomenclature[20]{$j_{v,x}$}{Shear-velocity in the inertial (world) reference frame along the $x$-axis}
\nomenclature[20]{$j_{v,y}$}{Shear-velocity in the inertial (world) reference frame along the $y$-axis}
\nomenclature[21]{$j$}{Shear-displacement magnitude}
\nomenclature[22]{$\tau$}{Shear-stress magnitude}
\nomenclature[24]{$\vect{p}_c$}{Position of the center of mass in inertial frame}
\nomenclature[25]{$ \dot{\vect{p}}_c$}{Velocity of the center of mass in the inertial frame}
 \nomenclature[27]{$\tilde{\vect{p}}_c $}{Projection of the center of mass on the bottom track level}
\nomenclature[28]{$\vect{p}_g$}{Projection of the center of mass on the terrain level}
\nomenclature[29]{$\boldsymbol{\Phi}$}{ZYX Euler angles representing vehicle orientation}
\nomenclature[30]{$\vect{n}_t$}{Normal component of the terrain   frame}
\nomenclature[31]{$\vect{t}_x$}{Tangential component of the terrain  frame along the $x$-axis}
\nomenclature[32]{$\vect{t}_y$}{Tangential component of the terrain  frame along the $y$-axis}
\nomenclature[33]{$\vect{K}_{t,x}$}{Terrain linear stiffness}
\nomenclature[34]{$\vect{D}_{t,x}$}{Terrain linear damping}
\nomenclature[35]{$\vect{K}_{t,o}$}{Terrain torsional stiffness}
\nomenclature[36]{$\vect{D}_{t,o}$}{Terrain torsional damping}
\nomenclature[37]{$\beta$}{Auxiliary variable equal to $\varphi + \varphi^d$}
\nomenclature[38]{$\beta_{L/R}$}{Longitudinal slippage on the left/right track}
\nomenclature[39]{$s$}{Target slack for optimal control feasibility}
\nomenclature[41]{$v^d$}{Reference longitudinal velocity}
\nomenclature[42]{$\omega^d$}{Reference angular velocity}
\nomenclature[43]{$k_p$}{Lyapunov control cartesian gain}
\nomenclature[44]{$k_\varphi$}{Lyapunov control orientation gain}
\nomenclature[45]{$x^d, y^d, \varphi^d$}{Reference state (expressed in the inertial frame)}
\nomenclature[46]{${}_b{F}_{x,L/R}$}{Longitudinal force generated by the left/right track in the body frame}
\nomenclature[46]{$\sigma_{ij}$}{Normal pressure for the $ij$ patch}
\nomenclature[47]{${}_b{F}_{y,L/R}$}{Lateral force generated by the left/right track in the body frame}
\nomenclature[49]{${}_b{M}_{z,L/R}$}{Moment about the $z$-axis generated by the left/right track in the body frame}
\nomenclature[51]{$\vect{f}_{n,ij}$}{Normal force acting on the $ij$-th element of the left/right track}
\nomenclature[52]{$F_{t}$}{Overall traction force generated by the terrain}
\nomenclature[53]{$F_{y}$}{Overall lateral force generated by the terrain}
\nomenclature[55]{$\vect{F}_n$}{Normal force generated by the terrain}
\nomenclature[56]{$\vect{M}_{n}$}{Moment generated by the normal terrain force }
\nomenclature[57]{$T_f$}{Duration of the planned trajectory}

\setlength{\nomitemsep}{-0.5\parsep}
{\small\printnomenclature[1cm]}

Note that unless otherwise stated, bold uppercase letters denote matrices in $\Rnum^{3 \times 3}$, while bold lowercase letters represent vectors in $\Rnum^{3}$ throughout the paper. The subscript $L/R$ refer to the left/right track, respectively. Components marked with the subscripts $ij$ denote the $i$-th row and $j$-th column of the tracks' discretized footprint.

\section{Introduction}
\label{sec:introduction}
\input{sections/1_introduction.tex}

\section{Dynamic modeling of tracked vehicles}
\label{sec:modeling}
\input{sections/2_dyn_modeling}

\section{Kinematic modeling}
\label{sec:tracked_model}
\input{sections/3_kin_modeling}

\section{Slippage identification}
\label{sec:slippage_identification}
\input{sections/4_identification}

\section{Control design}
\label{sec:control}
\input{sections/5_control}
\section{Slippage-aware motion planning (2D)}
\label{sec:planning}
\input{sections/6_planning}

\input{sections/7_results}

\input{sections/8_conclusions}

\section*{Acknowledgments}

The authors would like to thank \emph{Davide Dorigoni} and \emph{Francesco Biral} (University of Trento) for their help in modeling the terramechanics of the tracked vehicle.

\section*{Appendix}
\input{sections/9_appendix}

\bibliographystyle{style/ras/elsarticle-num}
\bibliography{references/references}


\end{document}

%% file: includes/used_packages.tex
\usepackage{booktabs}
\usepackage{moreverb}
\usepackage{url}
\usepackage{amsmath, amssymb, amsfonts}
\DeclareMathAlphabet{\altmathcal}{OMS}{cmsy}{m}{n}
\usepackage{multicol,lipsum}
\usepackage{bm}
\usepackage{color, colortbl}
\usepackage{svg}
\usepackage{booktabs}
\usepackage{multirow}
\usepackage{makecell}
\usepackage{acro}
\usepackage{nomencl}

\usepackage[normalem]{ulem} 
\usepackage{cancel}
\usepackage{setspace}
\usepackage{eucal} 
\usepackage{enumitem} 
\usepackage{xcolor}
\usepackage{mathtools}
\usepackage[hidelinks]{hyperref}
\hypersetup {
	pdftoolbar = true,
	colorlinks = false,
	linkcolor = black,
	citecolor = black,
	urlcolor = black,
}
\usepackage{caption}
\usepackage{glossaries}
\usepackage{siunitx}
\newcommand{\USI}[1]{\si[per-mode=symbol]{#1}}
\newcommand{\SSI}[2]{\SI[per-mode=symbol]{#1}{#2}}
\newcommand{\RSI}[3]{\lbrack\num{#1}{,}\ \num{#2}\rbrack\,\si[per-mode=symbol]{#3}}

%% file: includes/acronyms.tex
\newacronym{dofs}{DoFs}{Degrees of Freedom}
\newacronym{com}{CoM}{Center of Mass}
\newacronym{cop}{CoP}{Center of Pressure}
\newacronym{zmp}{ZMP}{Zero Moment Point}
\newacronym{fddp}{FDDP}{Feasible Differential Dynamic Programming}
\newacronym{cp}{CP}{Capture Point}
\newacronym{grfs}{GRFs}{Ground Reaction Forces}
\newacronym{vhsip}{VHSIP}{Variable Height Springy Inverted Pendulum}
\newacronym{ik}{IK}{Inverse Kinematic}
\newacronym{ocp}{OCP}{Optimal Control Problem}
\newacronym{nlp}{NLP}{Nonlinear Program}
\newacronym{ltv}{LTV}{Linear Time Varying}
\newacronym{lti}{LTI}{Linear Time Invariant}
\newacronym{td}{TD}{Touch Down}
\newacronym{lc}{LC}{Landing Controller}
\newacronym{qp}{QP}{Quadratic Program}
\newacronym{rl}{RL}{Reinforcement Learning}
\newacronym{mpc}{MPC}{Model Predictive Control}
\newacronym{fwp}{FWP}{Feasible Wrench Polytope}
\newacronym{micp}{MICP}{Mixed-Integer Convex Program}
\newacronym{srbd}{SRBD}{Single Rigid Body Dynamics}
\newacronym{to}{TO}{Trajectory Optimization}
\newacronym{ddp}{DDP}{Differential Dynamic Program}
\newacronym{dof}{DoFs}{Degree of Freedom}
\newacronym{ode}{ODE}{Ordinary Differential Equation}
\newacronym{msd}{MSD}{Mass-Spring-Damper}
\newacronym{pwbc}{pWBC}{projection-based Whole Body Control}
\newacronym{sp}{SP}{Support Polygon}
\newacronym{imu}{IMU}{Inertial Measurement Unit}
\newacronym{icr}{ICR}{Instantaneous Center of Rotation}
\newacronym{rt}{RT}{Real Time}
\newacronym{slip}{SLIP}{Spring Loaded Inverted Pendulum}
\newacronym{eom}{EoM}{Equation of Motions}
\newacronym{sqp}{SQP}{Sequential Quadratic Programming}
\newacronym{mic}{MIC}{Mixed-Integer Convex}
\newacronym{cmaes}{CMA-ES}{Covariance Matrix Adaptation Evolution Strategy}
\newacronym{ara}{ARA*}{Anytime Repairing A*}
\newacronym{pca}{PCA}{Principal Component Analysis}
\newacronym{cpg}{CPG}{Central Pattern Generator}
\newacronym{wbc}{WBC}{Whole-Body Control}
\newacronym{pd}{PD}{Proportional-Derivative}
\newacronym{nmpc}{NMPC}{Nonlinear Model Predictive Control}
\newacronym{wbopt}{WBOpt}{Whole Body Optimization}
\newacronym{hc}{HC}{Hunt and Crossley's}
\newacronym{kv}{KV}{Kelvin-Voigt's}
\newacronym{wllsr}{WLLSR}{Weighted Linear Least Squared Regression}
\newacronym{mae}{MAE}{Mean Absolute Tracking Error}
\newacronym{lip}{LIP}{Linear Inverted Pendulum}
\newacronym{nn}{NN}{Neural Network}
\newacronym{drl}{Deep-RL}{Deep Reinforcement Learning}
\newacronym{td3}{TD3}{Twin Delayed Deep Deterministic Policy Gradient}
\newacronym{ddpg}{DDPG}{Deep Deterministic Policy Gradient}
\newacronym{rpe}{RPE}{Relative Percentual Error}
\newacronym{grl}{GRL}{Guided Reinforcement Learning}
\newacronym{e2e}{E2E}{End-to-end Reinforcement Learning}
\newacronym{rbf}{RBF}{Radial Basis Function}
\newacronym{ugv}{UGV}{Unmanned Ground Vehicles}


%% file: includes/math_commands.tex
\newcommand{\Rnum}{\mathbb{R}} 

\newcommand{\vect}[1]{{\mathbf #1 }}

\newcommand{\mat}[1]{\ensuremath{\begin{bmatrix}#1\end{bmatrix}}}	
\newcommand{\x}{\ensuremath{\times}}

\def\x{{\mathbf x}}

\newcommand{\mcos}[1]{\cos\left(#1\right)}
\newcommand{\msin}[1]{\sin\left(#1\right)}

\newcommand{\matan}[2]{\text{atan2} \left( #1, #2\right)}



%% file: includes/newcommands.tex
\usepackage{resizegather}

\newcommand\BibTeX{{\rmfamily B\kern-.05em \textsc{i\kern-.025em b}\kern-.08em
T\kern-.1667em\lower.7ex\hbox{E}\kern-.125emX}}

\newcommand{\ie}{{i.e.},\ }
\newcommand{\eg}{{e.g.},\ }
\newcommand{\etal}{{\textit{et~al.}}}

\makeatletter
\newcounter{definition*}
\newenvironment{definition*}[1][htb]
{\renewcommand{\ALG@name}{Definition}
	\let\c@algocf\c@megaalgorithm
	\begin{algorithm*}[#1]%
	}{\end{algorithm*}}
\makeatother

\makeatletter
\newcounter{definition}

\makeatother

\definecolor{sfahmi_blue}{RGB}{0.19,0.51,0.74}
\definecolor{LightBlue}{RGB}{0.4,0.4,1}

\newcommand{\DSrev}[1]{#1}
\newcommand{\DSdel}[1]{}

%% file: sections/1_introduction.tex
Tracked mobile robots are known for their resistance to slippage, thanks to the large contact area between their tracks and the ground. This makes them well-suited for a wide range of robot-assisted agricultural applications, including weed control, seeding, fertilisation, pest management, and crop harvesting. 
However, this flexibility comes at a cost: the complex interaction between the vehicle and the ground makes the system dynamics highly intricate, the parameters difficult to estimate, and the system itself challenging to control.

Most of the difficulty lies in the so-called skid-steering
mechanism. Due to the lateral rigidity of the tracks, when a tracked
vehicle follows a curved path, the entire track assembly must rotate
at the same angular velocity as the vehicle. In other words, since the
vehicle cannot rely on the pure rolling assumption of differential
drive systems or a separate steering mechanism à la Ackermann, it is
forced to skid its tracks on the ground when turning. The motion of
skid-steering vehicles is determined by the two longitudinal track
forces and the lateral friction force. These vehicles control their
heading similarly to differentially driven systems — by varying the
relative speeds of the left and right tracks. However, the lateral
friction force depends on both the linear and angular velocities of
the vehicle, resulting in challenging and time-varying
constraints~\cite{Shiller1993}.

To fully exploit the potential of these vehicles, we need accurate and practical models that capture the terramechanics interaction, along with control strategies that can take full advantage of such models.

\DSrev{\textit{Applications.} \gls{ugv} are often deployed in
  precision agricultural settings~\cite{carpio2020navigation} to
  autonomously perform tasks such as
  spraying~\cite{lochan2024advancements,abhiram2022autonomous},
  inspection~\cite{pal2024novel}, and
  harvesting~\cite{park2023human}. Several mobile robotic platforms
  have been developed for these purposes, leading to significant
  improvements in productivity~\cite{gil2023low}.  Enhancing control
  accuracy is crucial in those applications where stringent tracking
  requirements must be met, particularly when automating routine or
  complex tasks in orchards~\cite{taseElbou25} and vineyards under
  varying terrain conditions.  In orchards, in particular, the
  environment is densely populated with irregularly distributed
  obstacles and narrow pathways. In such settings, tolerance for
  errors in trajectory planning and execution is considerably lower
  than in open agricultural fields.}


\begin{figure*}[htb]
    \centering
    \includegraphics[width=0.83\columnwidth]{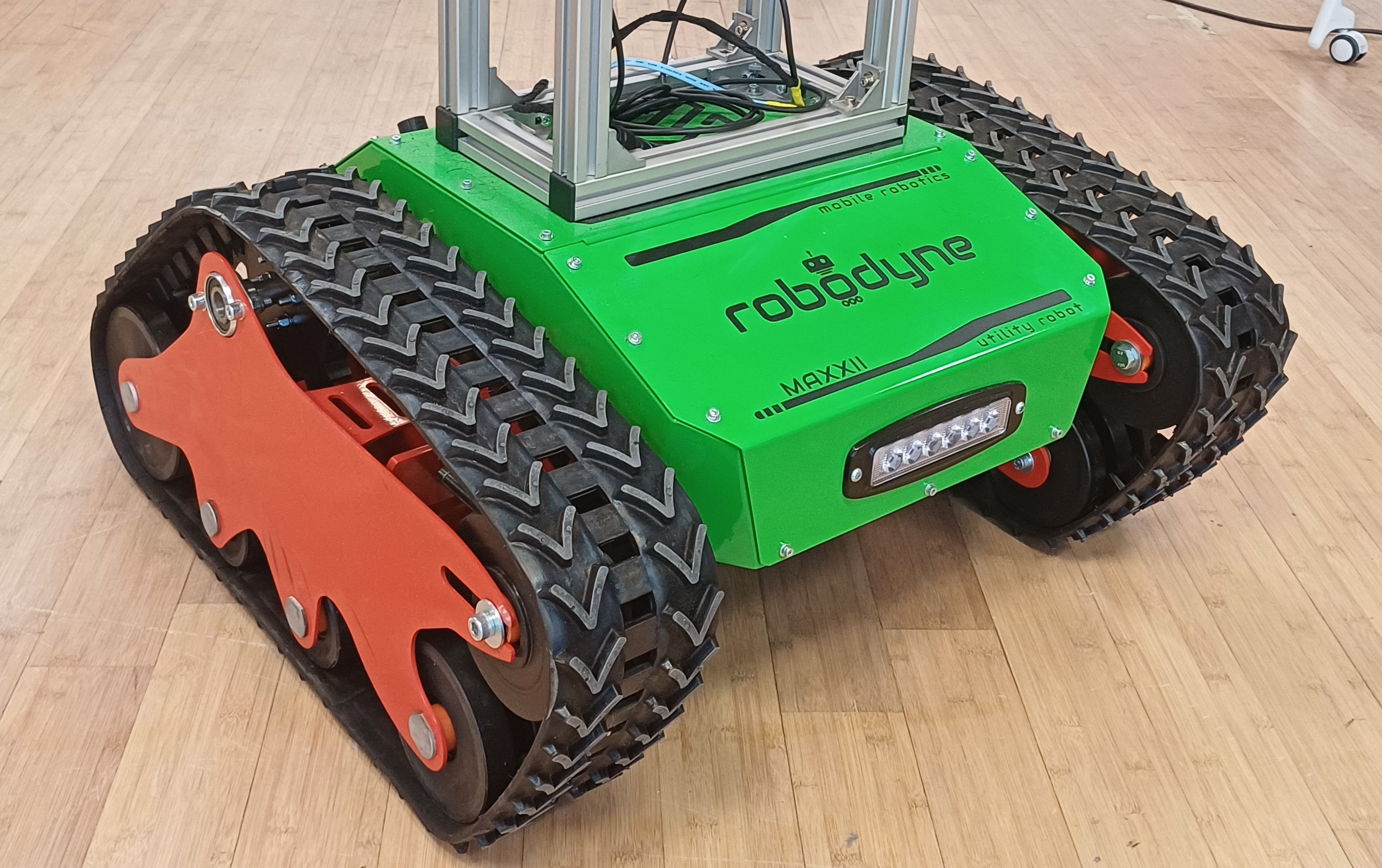}
    \includegraphics[width=0.7\columnwidth]{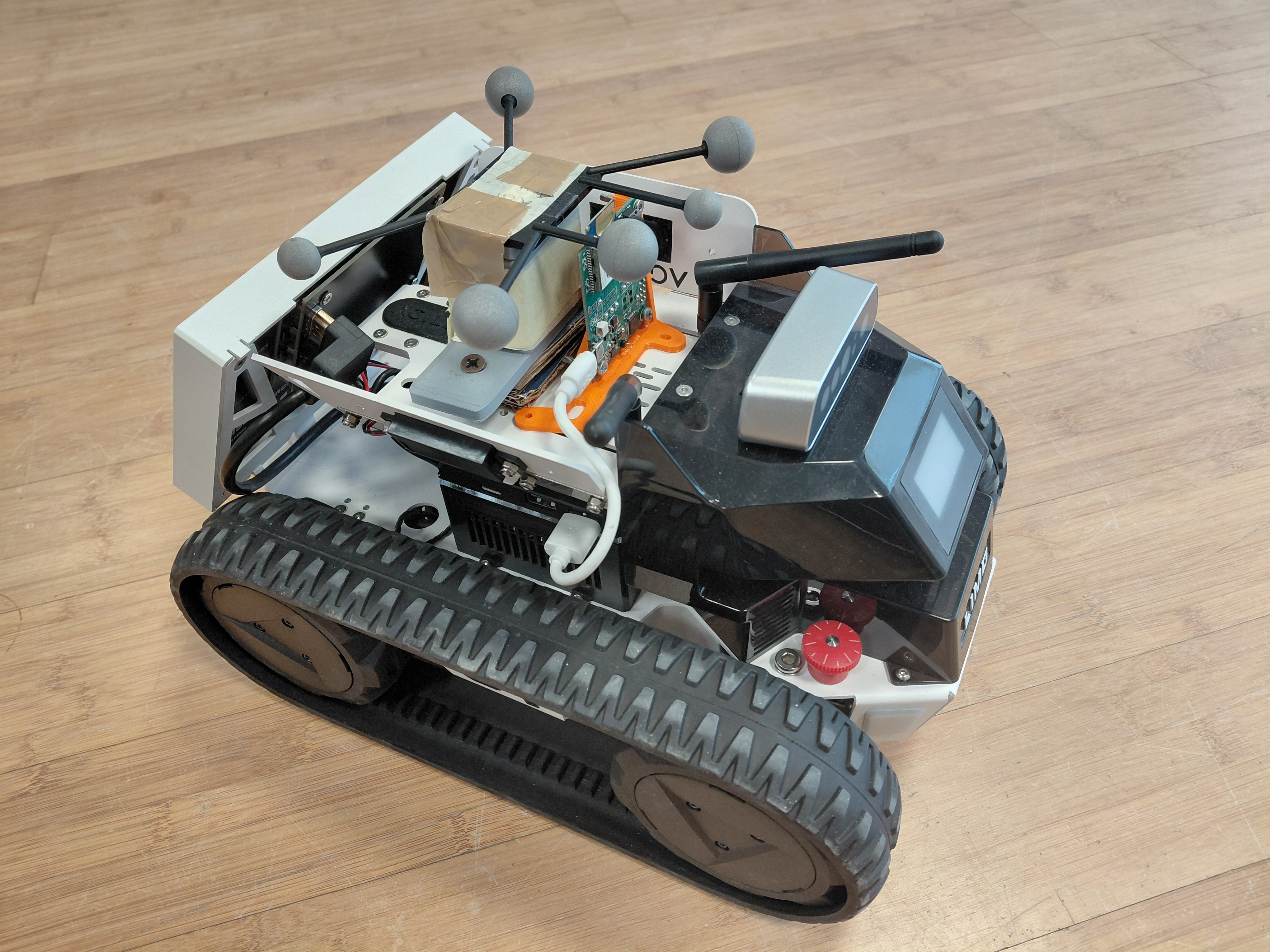}
  \caption{\DSrev{(left)} Picture of the MaxxII tracked vehicle \DSrev{\cite{maxxii_webpage} and (right) of the LIMO robot \cite{limo_webpage}} used for the experiments.}
  \label{fig:robots}
\end{figure*}

\paragraph{Related work}

Soil mechanics is often characterised by empirical shear
stress-slippage relationships, derived from massive amounts of
experimental data~\cite{Bekker, wong2022theory}. To obtain this
relationship, one must determine the deformation of soil elements and,
using nonlinear stress-strain relationships (i.e., the soil model),
calculate a complex 3D stress field resulting in soil-track
interaction forces. Due to the difficulty of deriving a full
theoretical model of soil-track interaction, simplified analytical
soil models are typically developed to capture the key physics, often
under the assumptions of homogeneous soil and a uniformly distributed
load along the tracks. At the macroscopic level, the interaction
forces between the track and the soil result in motion and
slippage. Although traction and lateral forces can be used in the
integration of dynamic models~\cite{Ahmadi2000}, track slip can also
be associated to purely kinematic effects by comparing the
odometry-based track velocity to the sprocket wheel's speed.

In this direction, different authors proposed an extended Kalman
filter~\cite{Ahn97, Zhao2024} or a least squares
approach~\cite{Song2004} to identify, in real time, the soil-track
slip parameters. Moosavian \etal{}~\cite{Moosavian2008} proposed
approximating the \emph{longitudinal} slip as an exponential function
of the \textit{turning radius}, resulting in an approximation that
proved effective at low speeds.

In the papers mentioned above, longitudinal slip estimation is
compensated through a feed-forward controller using a kinematic model,
where the angular speed of the driving sprockets is calculated at each
time step and sent to the low-level controller as a set point.
Piccinini \etal{}~\cite{Piccinini2023} take a different avenue, using
neural networks to predict lateral slippage for wheeled vehicles.
\DSrev{On a similar line, more recently, \cite{lupu2025magicvfm} proposed learning a rapidly tunable model of residual dynamics and disturbances, where a deep neural network maps terrain features and vehicle states to the current actuation matrix.}
		
Control solutions based solely on \emph{kinematic} models are
commonplace in the literature on nonholonomic vehicle
control. Differential‑drive vehicles are often modelled using a
nonlinear kinematic formulation — such as the well-known unicycle — where
trajectory control can be achieved via Lyapunov‑based
strategies~\cite{Kanayama1990} or feedback
linearisation~\cite{deluca99}. In contrast to differential‑drive
robots, the vast majority of control solutions for tracked vehicles
rely on \emph{dynamic} models~\cite{Ahmadi2000, Zou2018, Strawa2021,
Sabiha2022}, owing to the difficulty of separating kinematic and
dynamic effects.

Initial efforts to compensate for the system dynamics and apply
kinematic control laws accounting for the system nonholonomic nature
rooted on nonlinear feedback control~\cite{Fierro1997}. Zou
\etal{}~\cite{Zou2018} employed nonlinear feedback to convert the
dynamic control challenge into a kinematic one. More recently,
Sabiha \etal{}~\cite{Sabiha2022} combined back-stepping with sliding
mode control to design a trajectory tracking controller accounting for
track slippage. However, since their robot is operated at very low
\DSrev{advancing} speeds, slippage effects are minor and the advantages of their
strategy are not evident.

The complexity of modelling the interaction between track and terrain
has pushed toward the adoption of \emph{pseudo-kinematic} models,
where the dynamic effects are embedded into a small set of
parameters~\cite{Martinez2005}. Given the difficulty of accurately
modelling the complex slip behaviour between the tracks and the
terrain, pure kinematic approaches have been recently presented based
on the \gls{icr}~\cite{pentzer2014model}\footnote{The instantaneous
  centre of rotation is defined as the point in the horizontal plane
  about which the motion of the vehicle can be represented by a pure
  rotation and no translation occurs.}. Yamamuchi et
al.~\cite{yamamuchi} proposed various simplified longitudinal and
lateral slippage models for both turning and straight-line
motions. They used approximations to reduce terrain-dependent
parameters and employed regression analysis trained on inertial
data. The resulting models were used to achieve slip-compensated
odometry on loose and weak slopes. However, the use of
pseudo-kineamtic models in control is still an open problem and the
advantages are still to be shown.

The lessons learned in the literature mentioned above is that, given
the measurement of the vehicle velocity and the track drive sprocket
speeds, it is possible to determine key parameters of the soil model.
An important research direction is to find ways to integrate this
knowledge into control algorithms and motion planning that adapt to
the the soil conditions, \ie{} slippage-aware control and motion
planning.

\paragraph{Paper contribution and summary}

The purpose of this paper is to develop a reliable framework to
support robust navigation of tracked vehicles across a range of
terrains (of different slipperiness) and slopes. The framework
comprises three main components: 1. Accurate system models that
capture the complex effects arising from terramechanics dynamics;
2. Motion control strategies that leverage these models to ensure
high-fidelity trajectory tracking; 3. Scalable motion planning methods
that efficiently generate feasible trajectories, taking into account
the vehicle’s dynamic limitations.

\noindent{\bf Contributions to system modelling.}
Our modelling contribution is twofold. First, we introduce a
simulation model applicable to both flat and inclined terrains. This
model captures realistic terrain-vehicle interactions using a
distributed parameter approach, accounting for both uniform and
non-uniform load distributions. Its primary aim is to reduce the
sim-to-real gap by enabling the exploration of operating scenarios
that would be difficult or costly to replicate in a laboratory
setting.

Second, we propose a control-oriented pseudo-kinematic model, which
significantly simplifies the design of trajectory-following
controllers and motion planners. Our approach is inspired by Moosavian
\etal{}~\cite{Moosavian2008}, who model both lateral and longitudinal
slippage as pseudo-kinematic effects. In their formulation, lateral
slippage depends solely on the turning radius, however upon closer
analysis, this model exhibits a critical limitation: a singularity due
to a vertical asymptote when the turning radius approaches half the
vehicle’s width.  We address this issue by adopting an isomorphic
parametrisation that relates longitudinal slippage directly to the
sprocket wheel speeds. Additionally, we introduce a function
approximator, based on machine learning techniques, to estimate the
slippage parameters for the pseudo-kinematic model. The model has been
validated using experimental data collected from a MaxxII
robot~\cite{maxxii_webpage} (Fig.~\ref{fig:robots}(left)).

\noindent{\bf Contribution to trajectory control design.}
We propose a trajectory controller based on a pseudo-kinematic model
of a tracked vehicle that includes lateral slippage, inspired by
\cite{Zou2018} that incorporates feed-forward compensation for
longitudinal slippage. The slippage parameters are estimated via the
function approximators introduced earlier. The controller design is
grounded in a Lyapunov-based framework, which provides theoretical
guarantees on tracking performance by dynamically compensating for
lateral slippage during motion. \DSrev{ The enhanced controller
  performance was experimentally validated against a standard baseline
  unicycle controller on the LIMO tracked robot
  platform~\cite{limo_webpage} (see Fig.~\ref{fig:robots}(right)).}

\noindent{\bf Contribution to motion planning.}
We present a set of motion planning techniques with varying levels of
sophistication and evaluate their performance in conjunction with the
trajectory control algorithm proposed in this paper. The first class
of methods is based on Dubins manoeuvres and G1 curves
(clo\-tho\-ids). \DSrev{In particular,} Dubins manoeuvres \DSrev{comprises} circular arcs and
straight-line segments, \DSrev{and} are known to be optimal for nonholonomic
vehicles travelling at constant speed~\cite{dubins1957curves}. G1
curves, characterised by linearly varying curvature, offer a good
approximation of minimum-time paths for car-like vehicles subject to
curvature limits and longitudinal and lateral acceleration
constraints~\cite{FREGO201718}.  These curve-based techniques enable
the efficient generation of paths through a large number of
waypoints~\cite{clothoids,PastorelliDSFP25}, making them well-suited
for rapid trajectory planning. However, due to discrepancies between
the unicycle model and the dynamics of tracked vehicles, the resulting
trajectories may not be exactly trackable. As an alternative, we
propose a slippage-aware motion planner based on numerical
optimisation, which significantly improves tracking accuracy, albeit
at the cost of increased computational effort.

Finally, we present a thorough comparative analysis of the tracking
performance achieved by the slippage-aware controller and planner
\DSrev{against} \DSdel{versus} commonly used approaches based on
Dubins paths and clothoids. The improvements in accuracy are evaluated
through simulations using the aforementioned distributed parameter
models.  Table~\ref{tab:soa} summarizes the overall improvements of
our method with respect to the other methods (listed in chronological
order).

\begin{table*}[ht!]
	\centering
	\caption{  \DSrev{State of the art approaches comparison} }
       \resizebox{\textwidth}{!}{ \DSrev{
    \begin{tabular}{l c c c c c c  }
    \toprule
  	\textbf{Paper} & \textbf{Slip Prediction Method} & \textbf{Control  type } & \textbf{Exp. Validation}  & \textbf{Non Unif. Load} &   \textbf{Slopes} & \textbf{Slip aware Planning}\\
      \midrule            
   \textbf{Ours}       &  Regression, Decision Trees  & Kin. Based Control (Lyapunov)&  \checkmark &  \checkmark     &\checkmark  &\checkmark \\
    \cite{Shiller1993} &     Terramechanics equation       & Open Loop            &             &                 &\checkmark  &\checkmark \\
    \cite{Ahn97}       &     Ext. Kalman Filter       & Open Loop           &             &                 &            &            \\
   \cite{Ahmadi2000}  &       Terramechanics equation     & Feedback  Linearization      &             &                 &            &            \\
     \cite{Song2004}  &      Newton Raphson/LS      &      Open Loop       &             &                 &            &            \\
 \cite{Moosavian2008} &     Regression, exponentials &Kin. Based (Lyapunov)  & \checkmark  &                 &            &            \\
 \cite{Zou2018}      &   Terramechanics eq.&  Feedback lin. + Kin. Based Control (Lyapunov)&        &                 &           &            \\
 \cite{Strawa2021}   & Newton–Raphson (NR)/ Unsc. Kalman Filter& dynamic state feedback + state comp.&          &                 &           &            \\
 \cite{Sabiha2022} &    Terramechanics& Kin. back-stepping  +  Sliding Mode Dynamic Control & \checkmark &                &           &            \\
 \cite{Zhao2024} &      Sliding Mode Observer   &      Open Loop  &    \checkmark  &                 &            &            \\
     \cite{Martinez2005}& Estim. of Track's \gls{icr} via Generic Algorithms &Open Loop  &    \checkmark  &            &            &            \\
 \cite{lupu2025magicvfm}  &   Visual Foundation Model + deep neural network &Non-linear Adaptive Controller  & \checkmark  &                 &  \checkmark  &            \\
 \bottomrule              
	\end{tabular}
}
}    
\label{tab:soa}
\end{table*}

\paragraph{Outline}

The remainder of the paper is organised as follows. Section~\ref{sec:modeling} introduces the distributed parameter terrain models developed for flat and sloped surfaces \DSrev{that will be used as basis for the simulations in Sections~\ref{sec:flat_results} and~\ref{sec:slope_results}}. Section~\ref{sec:tracked_model} provides a brief overview of the pseudo-kinematic model for tracked vehicles \DSrev{upon which the controller will be synthesized}. Section~\ref{sec:slippage_identification} describes the machine learning framework for slippage identification and details the experimental setup used to collect data with the MaxxII robot. Section~\ref{sec:control} presents the design of the slippage-aware tracking controller, while Section~\ref{sec:planning} outlines the optimization-based slippage-aware planner together with other state-of-the art planning strategies. Sections~\ref{sec:flat_results} and~\ref{sec:slope_results} report the results of simulations and comparative evaluations on flat and sloped terrain, respectively, and discuss the limitations of the proposed approach. Experimental results with the LIMO robot are also reported in Section \ref{sec:hw_exps}. Finally, Section \ref{sec:conclusions}  draws the conclusions highlighting directions for future work.

%% file: sections/2_dyn_modeling.tex

We now introduce two dynamic models of a tracked vehicle: (a) a model
for simulating motion on flat terrain; and (b) an extension of this
model to 3D sloped terrain, incorporating progressively more realistic
terrain interactions, including the modeling of uniform and nonuniform
load distributions on the tracks.
\DSrev{These models will be used to generate the simulation results on flat  (Section \ref{sec:flat_results}) and sloped  (Section \ref{sec:slope_results}) surfaces, respectively.  The assumptions undertaken for each model together with their features are summarized in Table \ref{tab:model_assumptions}. }

\begin{table}[htp]
\centering
	\caption{\DSrev{Model Assumptions}} 
      \resizebox{\columnwidth}{!}{  
      \DSrev{
         \begin{tabular}{l c c c c   }
        \toprule
         \textbf{Model}                           &  \textbf{Terrain Interaction}  & \textbf{Slip Distrib.}        &   \textbf{Load Distrib.}       & \textbf{Terrain} \\
         \midrule                      
        Dynamic  \eqref{eq:2D_distributed_model}  &     Distributed                & Uniform       & Uniform                             & Flat   \\
        Dynamic  \eqref{eq:3D_distributed_model}  &     Distributed                & Uniform       & Non-uniform                          & Sloped   \\
        Kinematic  \eqref{eq:tracked_vehicle_kinematics}  & Lumped                 &  -            &   -                                    & Sloped   \\ 
     \bottomrule              
	\end{tabular}}
    }
  	\label{tab:model_assumptions}
\end{table}


\subsection{Dynamic model for flat terrain}
\label{sec:dynamic_modeling}

The dynamic model of a tracked vehicle for flat terrain corresponds to that  presented in~\cite{wong2022theory}, which is  reported here for completeness. This model carries the following \textbf{assumptions}: (a)~horizontal terrain with respect to gravity (\ie{}no terrain slopes); (b)~uniform slippage distribution; (c)~uniform load distribution under both tracks; (d)~no load transfer between tracks during motion due to accelerations.

The system is modelled through the state variables $x$, $y$, and
$\varphi$ (yaw angle with respect to the $z$-axis), and assumes a
constant robot height from the ground. To build the model, we first
need to compute the forces involved -- specifically, the interaction
forces between the vehicle and the terrain. To this end, we employ the
terramechanics framework (brush-bristle model) presented
in~\cite{wong2022theory}, which describes the relationship between
shear stress\footnote{In the context of a brush slippage model, shear
  stress refers to the force per unit area acting parallel to the
  surface of the material, causing it to deform or slide. This model
  simulates the interaction between a brush and a surface, where the
  brush's bristles deform and slide against the surface, generating
  shear stress.} and shear displacement as
\begin{equation}
  \tau = c + \sigma \mu\left(1 - e^{-j/K}\right) \text{,}
  \label{eq:shear_stress}
\end{equation}
where $\mu$ is the friction coefficient, $c$ is the terrain cohesion
coefficient that represents adhesion and between the track and the
ground, $\sigma$ is the normal pressure, and $K$ represents the shear
deformation modulus, i.e., the slope of the shear stress curve at the
origin in the case $c= 0$, which indicates the magnitude of the shear
displacement required to develop the maximum shear stress\footnote{ A
  higher shear deformation modulus $K$ indicates that a soil will
  resist more shear stress before deforming.  In hard soils, the shear
  modulus is generally higher, meaning they can withstand larger shear
  stresses without significant deformation.  It is worth noting that
  Coulomb's law of friction represents a special case of equation
  \eqref{eq:shear_stress}. According to Coulomb's law, the maximum
  frictional force is instantly mobilized as soon as any relative
  motion occurs between the track and the ground. This corresponds to
  setting the value of $K$ to zero.}.  On non-deformable terrains with
negligible adhesion between the track and the ground, the cohesive
forces of the ground are not present, hence term $c$ can be neglected.
To compute the shear stress $\tau$ we need the magnitude of the shear
displacement (also called shear strain) $j$.  First let's consider an
arbitrary point on either the left or the right track defined by the
coordinate pair $(x_p, y_p)$ expressed in the body reference frame of
the tracked vehicle.

\begin{figure}[t]
  \centering
  \includegraphics[width = 1.0\columnwidth]{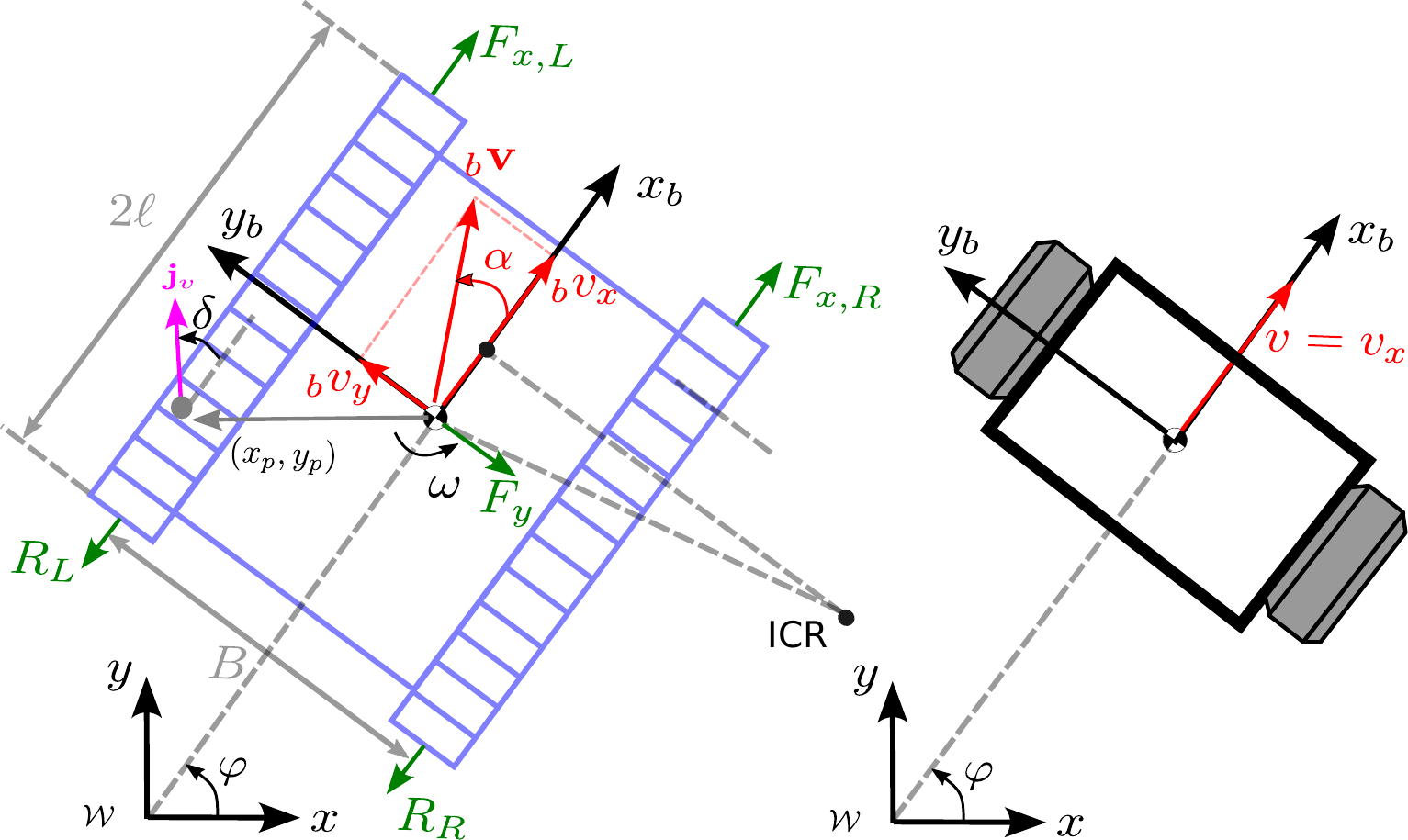}
  \caption{Top view of a differentially steered tracked mobile robot (left) and its unicycle approximation (right). Standard definitions for frames and variables are also provided for both models.}
  \label{fig:standard_definitions}
\end{figure}

We first present the development for the left track, as the same
reasoning applies to the right one.  Throughout this work, unless
specified otherwise, vectors are assumed to be expressed in the
inertial frame. A depiction of the tracked vehicle and its unicycle
approximation, together with the standard definitions for frames and
variables, is shown in Fig.~\ref{fig:standard_definitions}.
The shear velocity ${}_b\vect{j}_v \in \Rnum^2$ can be defined as the
difference between absolute velocity of a point on the left track
(expressed in the body reference frame) and the input sprocket wheel
speed:
\begin{equation}
  {}_b\vect{j}_v = \prescript{}{b}{\mat{ j_{v,x} \\
      j_{v,y}}}=\mat{
    {}_bv_x-\omega_z y_p  -\omega_{w,{L}} r\\
    {}_bv_y+\omega_z x_p} ,
\end{equation}
where ${}_bv_x$ is the longitudinal, ${}_bv_y$ the lateral velocity in
the body reference frame, $\omega_z$ is the angular velocity w.r.t.
the vertical axis, $\omega_{w,{L}}$ is the left sprocket wheel speed, and $r$ is the sprocket wheel radius (see Fig.~\ref{fig:standard_definitions}). 
The time required
for a point on the track to reach the coordinate $x_p$ from its
initial contact with the terrain is \DSrev{computed as}
\begin{equation}
 t_{p} = \int_{x_p}^{\ell} \frac{1}{\omega_{w,{L}}r} \, \mathrm{d}x_p = \frac{\ell - x_p}{\omega_{w,{L}}r} \text{,}
\end{equation}
where $\ell$ is the track semi-length. 
The angle that represents the angular displacement of the vehicle during the time interval  $t_{p}$, can be determined by the integration of the
angular velocity (assumed constant) with respect to time $t_p$ that it takes for the point ($x_p$, $y_p$) to travel from the initial point
of contact at the front of the track
\begin{equation}
  \varphi_{p} = \omega_z t_p\, \mathrm{d}x_p = \frac{\omega_z(\ell - x_p)}{\omega_{w,{L}}r} \text{.}
\end{equation}
If \DSrev{the ground is deformable and non-cohesive,} the adhesion between the track and the ground is
negligible. \DSrev{Therefore,} the shear displacement, in the inertial frame, can be obtained via integration of
the shear-velocities \DSdel{(mapped to the inertial frame)}
\begin{equation}
 j_{x} = \int_{x_p}^{\ell} \frac{ {}_wj_{v,x} }{\omega_{w,{L}}r} \, \mathrm{d}x_p \text{,} \quad
 j_{y} = \int_{x_p}^{\ell} \frac{ {}_wj_{v,y} }{\omega_{w,{L}}r} \, \mathrm{d}x_p \text{,}
  \label{eq:shear_displacement}
\end{equation}
where
\begin{equation}
 \prescript{}{w}{\mat{ j_{v,x}  \\  j_{v,y}}}  = 
 \mat{ \cos(\varphi_{p} + \varphi) & -\sin(\varphi_{p}+\varphi) \nonumber \\
 \sin(\varphi_{p}+\varphi) & \cos(\varphi_{p}+\varphi)}
\prescript{}{b}{ \mat{ j_{v,x}  \\  j_{v,y}} } \text{,}
  \label{eq:mapping_to_world}
\end{equation}
\DSrev{and} where $\varphi$ is the heading angle of the robot. This integral can be solved analytically and yields the following expressions for the shear-displacement components
\begin{align}
 j_{x} &= \dfrac{{}_bj_{v,y}\cos(a) + b\sin(a) - (v_y + \omega_z\ell)\cos(\phi)-b\sin(\phi)}{\omega_z} \nonumber  \text{,} \\
 j_{y} &= \dfrac{{}_bj_{v,y}\sin(a) - b\cos(a) - (v_y + \omega_z\ell)\sin(\phi)-b\cos(\phi)}{\omega_z} \nonumber
   \text{,} \\
 a &= \frac{\omega_z(\ell-x_p)}{\omega_{w,{L}} r} + \phi \text{,} \qquad b = {}_bv_x-\omega_z y_p \text{,}
\end{align}
whose magnitude is
\begin{equation}
 j = \sqrt{j_{x}^{2}+j_{y}^{2}}  \text{.} \nonumber
\end{equation}
By applying~\eqref{eq:shear_stress}, the magnitude $\tau(x_p, y_p, {}_bv_x, \omega_z, \omega_{L/R}) \in \Rnum$ of the shear stress
 at
$(x_p, y_p)$ is obtained, where $\omega_{L/R}$ are the wheel sprocket
speed of the left/right track.
To obtain the lateral and longitudinal components of the shear stress
vector, the shear-velocity is used. Indeed, the direction of the local
contact force for a point is opposite to its shear velocity
vector. Hence, by denoting with $\delta$ the angle with respect to the
base frame $x$ axis (see Fig. \ref{fig:standard_definitions} and refer
to~\cite{wong2022theory} for details on the computation of $\delta$),
we have
\begin{align}
 \sin(\delta) = \frac{{}_bj_{v, y}}{\|{}_b\vect{j}_{v}\|} \text{,} \quad \text{\DSrev{and}} \quad \cos(\delta) = \frac{{}_bj_{v, x}}{\|{}_b\vect{j}_{v}\|} \text{.}
\end{align}

\begin{figure}[t]
  \centering
  \includegraphics[width = 0.7\columnwidth]{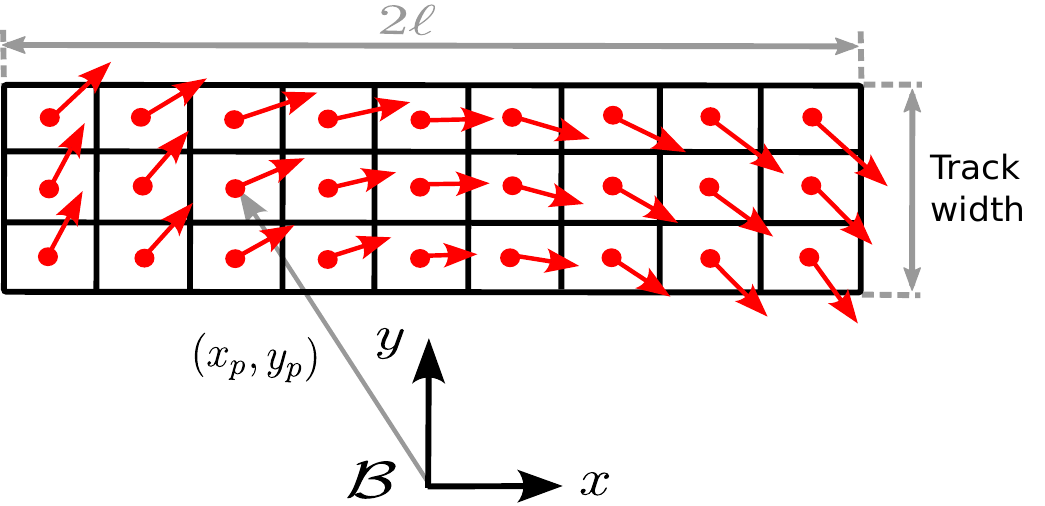}
  \caption{Left track discretisation and vector-field of the shear velocity  for a right turn.}
  \label{fig:left_track_discretization}
\end{figure}

As a consequence, the infinitesimal forces and moments exchanged
between the track and the ground by the terramechanics interactions
are obtained from the shear stress on infinitesimal contact areas
$\mathrm{d}A$ of the track, \ie{}\footnote{Note that because of the
  definition of $\delta$ the infinitesimal forces are defined in the
  body frame. Defining forces and dynamics in the body frame is
  instrumental to have a time-invariant inertia tensor, which
  simplifies the dynamic equations (when considering the non-flat
  terrain case).}
\begin{equation}
  \begin{split}
       &{}_b\mathrm{d}F_{\mathrm{x}} = -\tau  \cos(\delta) \, \mathrm{d}A \text{,} \\
      &{}_b\mathrm{d}F_{\mathrm{y}} = -\tau  \sin(\delta) \, \mathrm{d}A \text{,} \\
      &{}_b\mathrm{d}M_{\mathrm{z}} = -y_p \, {}_b\mathrm{d}F_{\mathrm{x}} + x_p \, {}_b\mathrm{d}F_{\mathrm{y}} \text{,}
   \end{split}
\label{eq:terra_mechanics_interactions1}
\end{equation}
Since the integration over the whole tracks' area
cannot be solved analytically, we discretise each track into $n$ patches of area $A_p$
 (refer to Fig.~\ref{fig:left_track_discretization}), and the
total forces and moments acting on the  left (or right) track are
calculated as the summation of every contribution as
\begin{equation}
 {}_bF_{\mathrm{x}} = \sum_{i=1}^{n}  {}_b\mathrm{d}F_{\mathrm{x}, i}, \quad
 {}_bF_{\mathrm{y}} = \sum_{i=1}^{n}  {}_b\mathrm{d}F_{\mathrm{y}, i}, \quad
 {}_bM_{\mathrm{z}} = \sum_{i=1}^{n}  {}_b\mathrm{d}M_{\mathrm{z}, i}.
  \label{eq:terra_mechanics_interactions2}
\end{equation}
Under the uniform load assumption, the normal pressure $\sigma$ used
to compute the shear stress in Eq.~\eqref{eq:shear_stress} remains
constant across all contact patches and can be directly determined
from the total load as
\begin{equation}
\sigma_{ij} = \frac{mg}{2A_t}, \quad \forall i,j \, \text{,}
\label{eq:sigma_uniform}
\end{equation}
where $A_t = n A_p$ represents the contact area of each track. Then,
for a tracked vehicle of mass $m$ with a moment of inertia $I_{zz}$
about a vertical axis (passing through the \gls{com}\footnote{We
  assume that the \gls{com} coincides with the origin of the body
  frame.}), the equations of motion can be written in the body frame
(c.f. Appendix for a detailed
derivation) as
\begin{align}
\begin{split}
  & m({}_b\dot{v}_x - {\omega_z}~{}_bv_y) = \underbrace{{}_bF_{\mathrm{x}, L} + {}_bF_{\mathrm{x}, R}}_{F_t} - \underbrace{({}_bR_L + {}_bR_R)}_{R_t}  \text{,} \\
  & m({}_b\dot{v}_y+{\omega_z}~{}_bv_x) = \underbrace{{}_bF_{\mathrm{y}, L} + {}_bF_{\mathrm{y}, R}}_{F_y} \text{,} \\
  & I_{zz}\dot{\omega_z} = \underbrace{ {}_bM_{\mathrm{z}, L} + {}_bM_{\mathrm{z}, R}}_{M_t}  - \underbrace{\frac{B}{2} ({}_bR_R + {}_bR_L)}_{M_r} \text{,} 
  \end{split}
  \label{eq:2D_distributed_model}
\end{align}
where 
$F_t$ is the total traction force, $R_t$ is the total
\emph{longitudinal} resistance force to the track movement (\ie{}due
to rolling friction), $F_y$ the total \emph{lateral} resistance force,
$M_t$ is the turning moment, $M_r$ the turning resistance moment and
$B$ is the distance between the two centrelines of the
tracks. Eq.~\eqref{eq:2D_distributed_model} represent the force
equilibrium in each of the three \gls{dof} of the vehicle. We
parametrise the 3D configuration space and its tangent mapping as
\begin{equation}
 \boldsymbol{\eta} = \mat{x \\ y \\ \varphi} \in \Rnum^2 \times S \text{,} \quad
 \dot{\boldsymbol{\eta}} = \mat{\dot{x} \\ \dot{y} \\\dot{\varphi} } = \mat{\vect{R}(\varphi) & \vect{0}\\ \vect{0}^\top & 1 }\mat{{}_bv_x \\ {}_bv_y \\ \omega_z} \text{,}
 \label{eq:generic_kinematics}
\end{equation}
where $\varphi$ is the vehicle heading angle (about the $z$-axis, which
is pointing up), and $x$ and $y$ its Cartesian position in absolute
coordinates (see Fig.~\ref{fig:standard_definitions}) and $\vect{R} \in \Rnum^2$ is a rotation matrix.
The symbol $\mathbb{R}^n$ represents the Euclidean space of dimension
$n$, $S$ the set of Euler angles defined on the interval
$[-\pi, \pi]$. 


\subsection{Dynamic model for sloped terrain}
\label{sec:dyn_model_sloped}
\begin{figure}[t]
  \centering
  \includegraphics[width = 1.0\columnwidth]{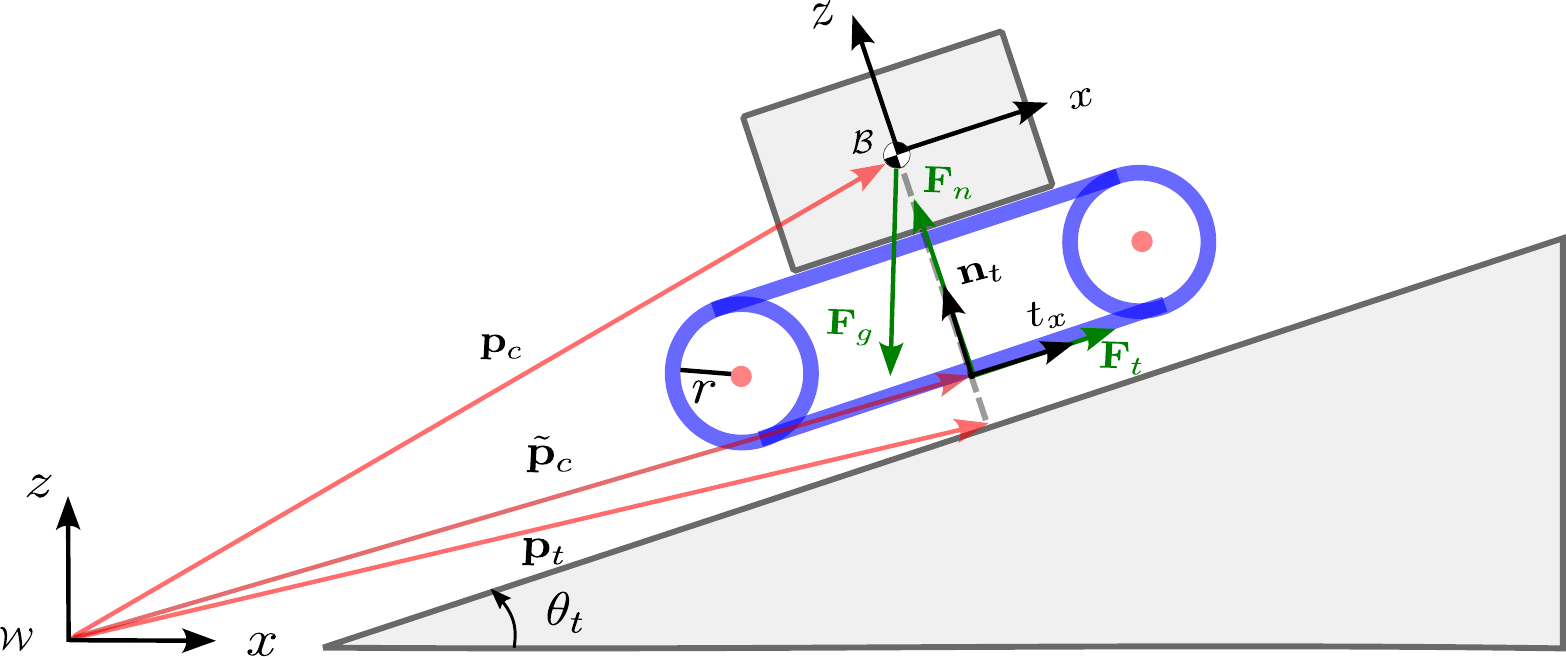}
  \caption{Side view of a differentially steered tracked mobile robot on a slope highlighting the standard definitions for frames and variables. $\mathcal{W}$ is the inertial frame, $\mathcal{B}$ is the body frame.}
  \label{fig:standard_definitions3D}
\end{figure}

In this section, we present the model of the tracked vehicle in a 3D
environment, \ie{}moving on sloped terrains. We model the robot as a
single rigid body, therefore the configuration space becomes 6D (see
Fig.~\ref{fig:standard_definitions3D} for the standard definitions)
and the dynamics will be described by the Newton-Euler
equations.
Extending the model~\eqref{eq:2D_distributed_model} to 3D implies a
higher number of kinematic \gls{dof} (from $3$ to $6$); however, the
motion of the robot is constrained on the terrain profile, reverting
back the number of unconstrained \gls{dof} to be 3. For the sake of
simplicity, we choose to parametrise the orientation with Euler
Angles, selecting the ZYX sequence, a widely adopted choice in
robotics.  Then the Newton-Euler equations
for the vehicle, written in a local body frame attached to the \gls{com} (see Appendix for details on body computation), are
\begin{align}
\begin{split}
 &m\left({}_b\dot{\vect{v}} + {}_b\boldsymbol{\omega} \times {}_b\vect{v}\right) =
 \underbrace{m~{}_w\vect{R}_b^T\mat{0 \\ 0 \\ -g}}_{{}_b\vect{F}_{g}} + {}_w\vect{R}_b^T \vect{F}_n + \mat{F_t + R_t\\F_y\\0}, \\
 &{}_b\vect{I}~{}_b\dot{\boldsymbol{\omega}} + {}_b\boldsymbol{\omega} \times {}_b\vect{I} ~{}_b\boldsymbol{\omega} =
 {}_w\vect{R}_b^T \vect{M}_n + \mat{0 \\ 0 \\ M_t+M_r} ,
 \end{split}
  \label{eq:3D_distributed_model}
\end{align}
where ${}_b\dot{\vect{v}}$ is the velocity of the vehicle's \gls{com}, ${}_w\vect{R}_b$ is the robot orientation,  
its inertia tensor ${}_b\vect{I}$, and ${}_b{\boldsymbol{\omega}}$, ${}_b\dot{\boldsymbol{\omega}}$ the angular velocity and acceleration vectors, respectively. 
With respect to Eq. \eqref{eq:2D_distributed_model}  new forces appeared, that previously were canceling each other:    the gravity force vector ${}_b\vect{F}_g$ and the
terrain normal force  $\vect{F}_n$ with its moment $\vect{M}_n$ (about the \gls{com}). 
As in the flat-terrain case $F_t$, $F_y$ are the traction and the lateral resistance force, respectively,
$R_t$ is the advancing resistance force, $M_t$ is traction moment, and $M_r$ is the rolling-resistance torque.  

\textbf{Remark:} It is useful to highlight that the interaction with the terrain is now decomposed into two components: (a) the tangential component ($F_t$, $F_y$), which results from terramechanics effects and is computed as described in Section \ref{sec:dynamic_modeling}; and (b) the normal component ${}_b{\vect{F}}_n$, which represents the reaction force generated by the track’s penetration into the terrain. The normal force is perpendicular to the terrain surface and ensures that the terrain constraint is satisfied. Its computation depends on whether the load distribution is assumed to be uniform or non-uniform. In the case of a uniform load distribution, the point of application is fixed in the middle of the tracks. In the nonuniform case, the application point may vary, but it is constrained to lie within the convex hull of the tracks.

\noindent \subsubsection{Uniform load distribution}
We consider first the case of \emph{uniform} load
distribution. This enables us to obtain a much simpler model
formulation where only a resultant force (and moment) is
representative of the overall \emph{normal} interaction with the
terrain. We model terrain compliance with a spring-damper model with
linear/torsional stiffness
$\vect{K}_{t,x}, \vect{K}_{t,o} \in \Rnum^{3 \times 3}$ and damping
$\vect{D}_{t,x}, \vect{D}_{t,o}\in \Rnum^{3 \times 3}$ and insert the
resulting contact force/moment $\vect{F}_{n}, \vect{M}_n \in \Rnum^3$
into the dynamics~\eqref{eq:3D_distributed_model}. We consider
the contact \emph{unilateral} and allow the force to be generated only
along the direction of the normal $\vect{n}_t$ to the contact (see
Fig.~\ref{fig:standard_definitions3D}). 
\begin{equation}
 \vect{F}_{n} =
  \begin{cases}
 \vect{n}_t \vect{n}_t^{\top}(\vect{K}_{t,x}(\vect{p}_t -
 \tilde{\vect{p}}_c ) - \vect{D}_{t,x} \dot{\vect{p}}_c) &
 \vect{n}_t^\top(\vect{p}_t - \tilde{\vect{p}}_c ) >0 , \\
 \vect{0} & \text{otherwise} ,
  \end{cases}
\end{equation}
where $\vect{p}_t$ is the projection of the \gls{com} at the terrain level and $\tilde{\vect{p}}_c$ 
its projection at the track level, respectively. $\tilde{\vect{p}}_c$ is obtained evaluating the terrain mesh under  $\vect{p}_t$.
The spring/damper generates a force whenever there is
penetration of $\tilde{\vect{p}}_c$beenbeen inside the terrain, zero otherwise. 
A similar
approach has been taken for the rotational moment, with the difference that the tangential
components along $\vect{t}_x$ and $\vect{t}_y$ will be accounted for
in the dynamics
\begin{equation}
 \vect{M}_{n} = \underbrace{[\vect{I} - \vect{n}_t
   \vect{n}_t^\top]}_{\vect{N}}
 \left(\vect{K}_{t,o}\vect{e}\left({}_w\vect{R}_b,{}_w\vect{R}_t(n_t)\right)
   - \vect{D}_{t,o} \boldsymbol{\omega}\right) ,
\end{equation}
where $\vect{e}(\cdot)$ is the function that computes the difference 
in orientation of the robot ${}_w\vect{R}_b$ 
w.r.t. the terrain ${}_w\vect{R}_t(n_t)$
 and
$\vect{N}$ is a projection operator which maps into the
$\vect{t}_x-\vect{t}_y$ plane. Note that there is no unilateral
constraint on the moments.

\noindent \subsubsection{Nonuniform load distribution} 
Instead of assuming that the load is evenly spread, we consider a more
realistic distribution that better captures the actual force
variations across the contact surface. The terrain is still modeled
with a spring-damper model. Since the track is not supposed to be
deformable, the stiffness of the terrain is the only determining one
(being the two stiffness in series).
As for the case of terramechanics interactions, the tracks' contact surface is
discretised in patches, computing, for each patch, \textit{also} the linear force due to the
penetration with the terrain  (there is no need to compute moments
because the patch size is small with respect to the track dimension). The
load distribution, \ie{}the distribution of the normal forces over the
track surface, will naturally emerge from the integration of the
dynamics.

\begin{figure}[t]
  \centering
  \includegraphics[width = 1.0\columnwidth]{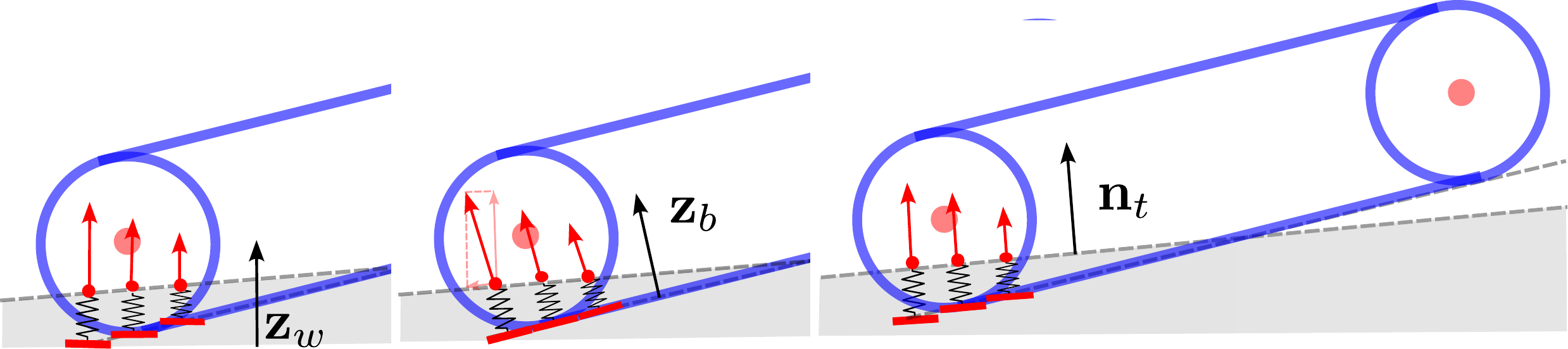}
  \caption{Different projection directions for the ground reaction
    forces: (left) along the $z$-axis of the inertial frame, (middle)
    along the $z$-axis of the body frame, (right) along the terrain
    normal $\mathbf{n}_t$.}
  \label{fig:projections}
\end{figure}

As in the case of uniform load, we need to: (a) evaluate the mesh
elevation at the position of the contact patches; (b) compute the
linear forces $\vect{f}_{n,ij}$ from the terrain penetration for all the
patches (c) compute the normal pressure $\sigma_{ij}$ on each patch
dividing $\vect{f}_{n,ij}$ for the patch area $A_p$ and (d) compute the
terramechanics interactions for that loading condition from
Eq.~\eqref{eq:terra_mechanics_interactions1} and
Eq.~\eqref{eq:terra_mechanics_interactions2}. For each patch, by
forward kinematics computations, we get the position $\vect{p}_{ij}$
and velocity $\dot{\vect{p}}_{ij}$ of each patch of the track (in an
inertial frame $\mathcal{W}$), and the projection $\vect{p}_{t,ij}$ of
$\vect{p}_{ij}$ onto the terrain mesh using a \emph{ray casting}
algorithm~\cite{ray_casting_open3d}\DSrev{.}
The choice of the projection direction has an impact on the
resulting terrain forces $\vect{f}_{n,ij}$. Possible options are:
(a) the inertial frame $z$-axis, (b) the base frame $z$-axis, and
(c) the terrain normal $\mathbf{n}_t$ (see Fig.~\ref{fig:projections}).
We found that proper load transfer was only achieved when the projection was performed along the world frame $z$-axis.
Therefore, considering also the terrain damping, the force interaction $\vect{f}_{n,ij}$ for each patch will be
\begin{align}
    \rho&= \vect{z}^{\top}_{w} \left( \vect{p}_{t, ij}-  \vect{p}_{ij} \right) \\
 \dot{\rho}&= \vect{z}_{w}^\top \dot{ \vect{p}}_{ij}  \\
 \vect{f}_{n,ij}&=  \vect{z}_{w} \left(\mathbb{I}_{\rho > 0} \vect{K}_{t,ij} \rho -\mathbb{I}_{\dot{\rho} < 0} \vect{D}_{t,ij}\dot{\rho} \right) A_t
    \label{eq:terrain_forces}
\end{align}
where $\rho$ and $\dot{\rho}$ is the terrain penetration and its time derivative, respectively, and $\mathbb{I}_{\rho > 0}$, $\mathbb{I}_{\dot{\rho}< 0}$ are appropriate indicator functions that model the unilaterality of the contact. Note that both terrain stiffness $\vect{K}_{t,ij}$ and damping $\vect{D}_{t,ij}$ for the patch ($ij$) are normalised by the track area $A_t$.

\paragraph{Velocity-dependent terrain stiffness}
In real conditions the terrain is deformable and stiffens  with the robot speed due to increased compression, 
as well as with its position relative to the track transversal middle line. We model this phenomena by computing the $ij$-th patch stiffness as follows
\begin{align}
 \vect{K}_{t,ij} = \vect{K}_{t} \left[1 + \vect{K}_{t,p}~{}_bv_x \left( {}_b\vect{p}_{ij,x} + \mathrm{sign}({}_bv_x)\ell \right)\right] \text{,}
  \label{eq:patch_stiffness}
\end{align}
where $\vect{K}_{t}$ is the nominal terrain stiffness, $\vect{K}_{t,p}$ is the rate
of variation of terrain stiffness with robot speed and
${}_b\vect{p}_{ij,x}$ is the $x$-axis coordinate of the $(i,j)$ patch
in the body frame. More precisely, when the robot is moving forward,
in the back of the tracks ${}_b\vect{p}_{ij,x} = -\ell$ there is no stiffness
variation while there a maximum increase in the front.  The moment for
each patch is computed as
\begin{align}
 \vect{m}_{n,ij} = \left(\vect{{p}}_{ij} - \vect{p}_c \right) \times \vect{f}_{n,ij} \text{.}
  \label{eq:terrain_moments}
\end{align}
Note that the computation of the normal pressure $\sigma_{ij}$ in~\eqref{eq:sigma_uniform} should be replaced by
\begin{equation}
\sigma_{ij} = \frac{{}_w\vect{R}_b^\top \vect{f}_{n,ij}}{A_p} \quad \forall i,j \, \text{,}
\label{eq:sigma_distributed}
\end{equation}
where $A_p$ is the patch area. The sum of the contributions for all patches in both tracks gives us the total ground force $\vect{F}_n$ and moment $\vect{M}_n$. 


\subsubsection{Static friction}
\noindent To compute the $R_t$ term (static friction)
in~\eqref{eq:2D_distributed_model}
and~\eqref{eq:3D_distributed_model}, we start by recalling that \DSrev{the} static
friction represents the resistance to rolling (opposite to the
motion), thus we compute it depending on the loading of each patch
\begin{align}
 R_t = \sum_{ij} \vect{f}_{n,{ij}} \, c_r \, \mathrm{sign}({}_bv_x) \text{,}
  \label{eq:rolling_friction}
\end{align}
where $c_r$ is the rolling friction coefficient. 

\subsection{\DSrev{Numerical integration of the equations of motion}}
After solving for ${}_b\dot{\vect{v}}$,
${}_b\boldsymbol{\dot{\omega}} \in \Rnum^3$ in
eq~\eqref{eq:3D_distributed_model}, we can then integrate the states
with a generic implicit or explicit Runge-Kutta method, i.e.,
\begin{equation}
   \begin{array}{l}
 \vect{F}\left(\vect{x}_{k} + h_k\displaystyle\sum_{j=1}^{s}a_{ij}  \vect{K}_j, ~\vect{K}_i, t_{k} + c_i h_k\right) = \vect{0} , \\
      \vect{x}_{k+1} = \vect{x}_{k} + h\displaystyle\sum_{i=1}^{s} b_i \vect{K}_i , 
  \end{array}
\end{equation}
for $i = 1, \dots, s$ stages, and $k = 0, 1, \dots, m$ time steps of
size $h_k$, where $b_i$ is the $i-th$ weight relative to $\vect{K}_i$
stage estimated slope, $a_{ij}$ are the coefficients of the
Runge-Kutta matrix, and $c_i$ are the nodes and $h$ the size of the
time-step interval~\cite{stocco2024matrix, Hairer1996}. Notice that~\eqref{eq:3D_distributed_model} can be
rewritten as an implicit first-order \gls{ode} to be integrated as
\begin{equation}
 \vect{F}(\vect{x}, \dot{\vect{x}}, t) = 
 \begin{bmatrix}
\dot{ \vect{p}}_{c, k+1}- \vect{f}_{\mathrm{dyn}}({}_w\vect{R}_b {}_b\vect{v}_{c,k}) \\
 \dot{\Phi}_{k+1} - \vect{T}(\boldsymbol{\Phi})^{-1}{}_b\boldsymbol{\omega}_k
  \end{bmatrix} \text{.}
\end{equation}
It is worthwhile to note that $\vect{T}(\cdot)\in \Rnum^3$ is a
nonlinear mapping from Euler Rates $\dot{\boldsymbol{\Phi}}$ to angular velocities.

%% file: sections/3_kin_modeling.tex

In this section, we will present a pseudo-kinematic model that captures the terramechanics interactions in few lumped parameters that will be used for the synthesis of the control law \DSrev{in Section \ref{sec:control}}. Before presenting the pseudo-kinematic model we first introduce the \emph{unicycle} model which is instrumental for the next derivation.

\subsection{Unicycle model}
A differentially driven mobile robot can be modeled as shown in
Fig.~\ref{fig:standard_definitions}(right), where the speeds of the
two wheels can be independently controlled to provide a desired
longitudinal speed  ${}_bv_x$ (\ie{}along the body $x_b$ axis) and
a yaw rate $\omega_z$ about the vertical axis. In this derivation, we
assume a \emph{flat} ground. Let the state of the system be its pose
on a plane. If the pure rolling condition holds (\ie{}the sway speed
${}_bv_y$ is identically zero), then the kinematic equations are
notoriously given by
\begin{equation}
  \left\{\!\!\begin{array}{l}
    \dot{x} = {}_bv_x \mcos{\varphi} \\
    \dot{y} = {}_bv_x \msin{\varphi} \\
    \dot{\varphi} = \omega_z
  \end{array}\right.,
  \label{eq:unicycle_kinematics}
\end{equation}
where $x$ and $y$ are the Cartesian coordinates (expressed in the
world frame) and $\varphi$ is the angle between the robot body and the
inertial frame $x$-axis.
It is possible to relate the kinematic inputs ${}_bv_x$, $\omega_z$, to the
left and right sprocket wheel velocities of the tracks, namely
$\omega_{w,L}$ and $\omega_{w,R}$ respectively, by a linear mapping
\begin{equation}
  \mat{{}_bv_x \\ \omega_z} = \mat{\frac{r}{2} & \frac{r}{2} \\ -\frac{r}{B} & \frac{r}{B}} \mat{\omega_{w,L} \\ \omega_{w,R}} ,
  \label{eq:unicycle_mapping}
\end{equation}
where $r$ is the wheel sprocket radius and $B$ is the distance between
tracks.

\subsection{Pseudo-kinematic model for tracked vehicles}

Getting inspiration from~\cite{Zou2018, yamamuchi}, we intend to
extend~\eqref{eq:unicycle_kinematics}, \eqref{eq:unicycle_mapping} to \textit{tracked} vehicles,
lumping the terrain interactions into three parameters, namely: the
\emph{lateral} slippage angle $\alpha$ and the left and right track
\emph{longitudinal} slippages $\beta_L$ and $\beta_R$,
respectively. The kinematics equations are given by~\cite{Zou2018,
  Sabiha2022}, \ie{}
\begin{equation}
  \left\{\!\!\begin{array}{l}
    \dot{x} = \dfrac{{}_bv_x }{\mcos{\alpha}} \mcos{\varphi + \alpha} \\
    \dot{y} = \dfrac{{}_bv_x }{\mcos{\alpha}} \msin{\varphi + \alpha} \\[1.0em]
    \dot{\varphi} = \omega_z
  \end{array}\right. \text{,}
  \label{eq:tracked_vehicle_kinematics}
\end{equation}
where the lateral slip angle $\alpha$ represents the deviation of the
velocity vector with respect to the $x_b$ axis due to the lateral
slippage (i.e., due to a velocity component ${}_bv_y$ along the $y_b$
axis, as in Fig.~\ref{fig:standard_definitions}~(left) which derives
from the skid steering behaviour).  $\alpha$ can be computed from the
actual velocity measurements as
\begin{equation}
  \alpha = \tan^{-1} \left(\frac{{}_bv_y}{{}_bv_x}\right) \text{.}
  \label{eq:alpha_estimate}
\end{equation}

Note that the slip angle increases with the path curvature
(\ie{}decreases with the turning radius), it is positive (negative)
for a clockwise (counterclockwise) turn, and zero
for straight line motions. The value of the slip angle is also tightly
interconnected to the (forward) shift of the \gls{icr} during a
turning manoeuvre~\cite{Shiller1993}. On the other hand, the
\emph{longitudinal} slippage (\ie{}along $x_b$ axis, as in
Fig.~\ref{fig:standard_definitions}~(left)) \emph{directly} affects
the wheel speed by modifying the inputs to
Eq.~\eqref{eq:tracked_vehicle_kinematics} as
\begin{equation}
 \mat{{}_bv_x \\ \omega_z } = \mat{\frac{r}{2} & \frac{r}{2} \\ -\frac{r}{B} & \frac{r}{B}} \mat{\omega_{w,L} + \frac{\beta_L (\omega_{w,L}, \omega_{w,R})}{r}\\ \omega_{w,R}+ \frac{\beta_R(\omega_{w,L}, \omega_{w,R})}{r}}
 \label{eq:tracked_vehicle_mapping}
\end{equation}
From kinematics, track velocities w.r.t. \DSrev{the} ground can be related to ${}_bv_x$ and $\omega_z$ by
\begin{equation}
  \begin{aligned}
    & {}_bv_L^{t}={}_bv_x-\omega_z \frac{B}{2}, \quad & {}_bv_R^{t}={}_bv_x+\omega_z
    \frac{B}{2} \text{,}
  \end{aligned}
  \label{eq:track_speed}
\end{equation}
the longitudinal slippage is the relative speed between the track and the ground 
\begin{equation}
  \begin{aligned}
    & \beta_L ={}_bv_L^{t} -\omega_{w,L} r , \quad & \beta_R = {}_bv_R^{t} - \omega_{w,R} r \text{.}
    \label{eq:beta_estimate}
  \end{aligned}
\end{equation}
where the left/right input sprocket wheel velocities $\omega_{w,L}$
and $\omega_{w,R}$ are obtained by the encoder readings.  Whenever the
track velocity w.r.t. \DSrev{the} ground equals the input velocity the longitudinal
slippage is zero.  This is defined to be positive when the traction
effort produced assists the longitudinal motion (\ie{}outer turning
wheel) and negative otherwise (\eg{} inner turning wheel). Note that
while $\beta_L$ and $\beta_R$ affect directly the inputs and can be
compensated in a feed-forward fashion in~\eqref{eq:unicycle_mapping};
in the case of $\alpha$, there is an indirect dependency on the inputs
through the dynamics.
This motivates our choice of accounting for this effect directly in the controller design of Section~\ref{sec:control}.

\subsection{Embedding of pseudo-kinematic model and dynamic model}
The slippage parameters in the pseudo-kinematic
model~\eqref{eq:tracked_vehicle_kinematics} can be conceptually linked
to the dynamic models~\eqref{eq:2D_distributed_model},
\eqref{eq:3D_distributed_model}.  In
Fig.~\ref{fig:standard_definitions}, it can be noted that the presence
of the lateral slippage is directly linked to the presence of a
lateral velocity component ${}_bv_y$. This is due to the lateral
acceleration along the $y_b$ axis (i.e., body frame axis), generated
by the lateral force $F_y$. On its turn, the lateral force is due to
the effect of the lateral component of the shear velocity
${}_bj_{v,y}$ on the track surface through
\eqref{eq:shear_stress}. The longitudinal slippage, instead, can be
directly related to its longitudinal component ${}_bj_{v,x}$.

%% file: sections/4_identification.tex

In this section, the experimental identification of the lateral and
the longitudinal slippage are presented separately due to their
different nature. The purpose is to show a promising matching between
simulation and experiments providing a good validation of the
simulation environment.

\subsection{Hardware platform}

The hardware platform used in this study is the MaxxIIx~\cite{maxxii_webpage}
tracked robot shown in Fig.~\ref{fig:robots}(left). To enable test
ground-truth odometry, the robot is equipped with marker tags mounted
on top of its chassis, which are detected by an OptiTrack Motion
Capture System. Additionally, it features motor encoders on its
driving wheels, with wheel speed commands provided through a ROS2
driver interface. The experiments are executed on an onboard Intel NUC
11 Mini-PC with 16GB RAM.
The control loop operates at a frequency of \SSI{200}{\hertz}, as well
as the OptiTrack system. All tests are conducted in an indoor robotics
lab featuring a slippery parquet floor. The actual speed of each track
is determined by differentiating the robot's posture (position and
heading) over time and employing~\eqref{eq:track_speed}. Since
numerical differentiation can amplify noise, highly accurate and
smooth localization is required. To achieve this, we connected the
motion capture system via a low-latency wired network, capable of
providing reliable data at \SSI{200}{\hertz}. Furthermore, we applied
mild filtering to the position signal, which proved beneficial in
mitigating the effects of quantization noise introduced by the
differentiation.

\subsection{Longitudinal slippage}
\label{sec:long_slippage_ident}

Moosavian \etal{}~\cite{Moosavian2008} claim that for limited speeds
the slippage in tracks is almost independent from vehicle speed and
employ exponential to approximate the relationship between wheel
slippage and the \emph{turning radius}. Our findings highlight that
there is also a dependency on the vehicle's absolute velocity in the
range of speeds of our interest. Additionally, the definition of the
longitudinal slippage $\beta = 1-v_{t}/\omega_{w}r$ given
in~\cite{Moosavian2008} has a singularity (shown as a vertical
asymptote) when the turning radius $R$ is equal to half the distance
from the two wheels and the inner wheel speed becomes
zero~\cite{Moosavian2008, yamamuchi}.
%
A more meaningful function approximation should account for
dependencies on both the turning radius and the longitudinal speed
${}_bv_x$.  In this work, we propose an alternative parameterization
based on both wheel speeds, which, thanks
to~\eqref{eq:unicycle_mapping}, can be shown to be isomorphic to
(${}_bv_x$, $\omega_z$), therefore also to 
the turning radius $R={}_bv_x/\omega_z$. This approach has the added
advantage of avoiding singularities when the inner wheel speed crosses
zero. Hence, we propose to design a multiple-input regressor, based on
machine learning techniques, to estimate the values of $\beta_L$,
$\beta_R$ and $\alpha$ (see next Section) receiving as inputs the wheel sprocket speeds
$\omega_{w,L}$ and $\omega_{w,R}$
\begin{equation}
 \mat{\beta_L\\ \beta_R}  = f_{\beta}\left(\omega_{w,L},\omega_{w,R}\right).
\label{eq:fbeta}
\end{equation}
\DSrev{ We tested different types of regression method, namely
  decision trees, neural networks and pure \gls{rbf} interpolation.
  Table \ref{tab:learning_methods} summarizes the different machine
  learning methods used for slip parameter estimation, including the
  characteristics of the regression model and their pros and cons.
  The dataset has been split 80\% as train and 10\% as validation set
  and 10\% test set.  The validation dataset was employed to guide
  model training and prevent overfitting.  The performance evaluation
  was executed on the test set using the $r^2$ score.}
\begin{table*}[htp]
    \centering
    \caption{\DSrev{Comparison of learning methods for slippage parameters' estimation}}
    \label{tab:learning_methods}
    \DSrev{
        \resizebox{\textwidth}{!}{
        \begin{tabular}{l c p{8.5cm} p{7.5cm} }
        \toprule
        \textbf{Method} & \textbf{Training Time [s]} & \textbf{Advantages} & \textbf{Limitations} \\
        \midrule
        \gls{rbf} Interpolation 
        & 1.10 
        & Provides  geometrically interpretable model for low-dimensional problems; smooth interpolation behavior 
        & Requires loading of all data at runtime; scales poorly with dimensionality; limited generalization capability \\
         \midrule
        Decision Trees 
        & 2.68 
        & Interpretable and easy to visualize; effectively captures nonlinear relationships; insensitive to input feature scaling 
        & Unable to extrapolate beyond the training domain; prone to overfitting if not properly regularized \\
        \midrule
        Neural Network (NN) 
        & 17.00 
        & Highly scalable to large datasets; capable of modeling complex nonlinear mappings 
        & May fail to capture small-scale behaviors (e.g., zero crossings); prone to overfitting; requires extensive training data and tuning \\
        \bottomrule
    \end{tabular}}
    }
\end{table*}

\begin{figure}[t]
 \centering
 \includegraphics[width=0.47\columnwidth]{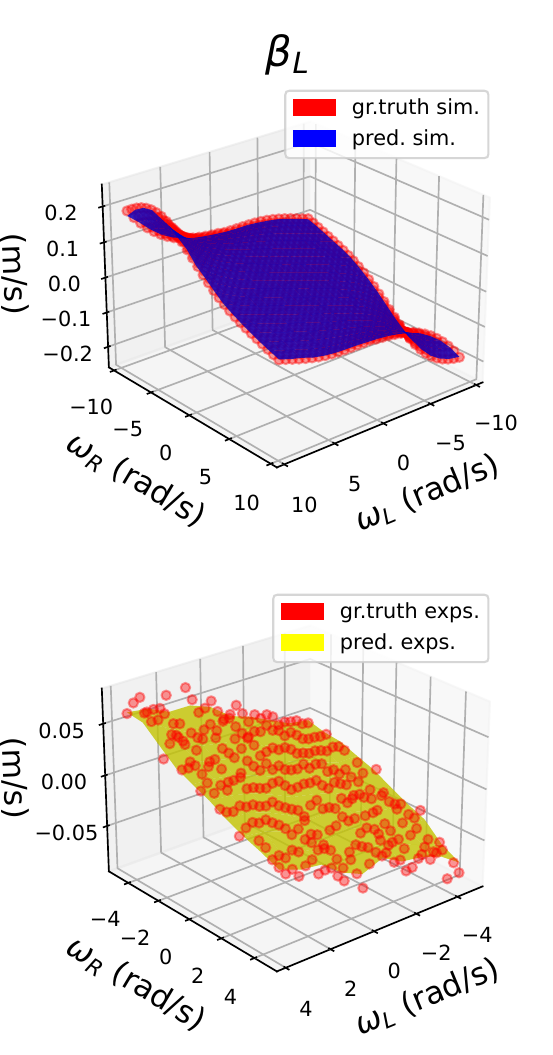}
 \includegraphics[width=0.47\columnwidth]{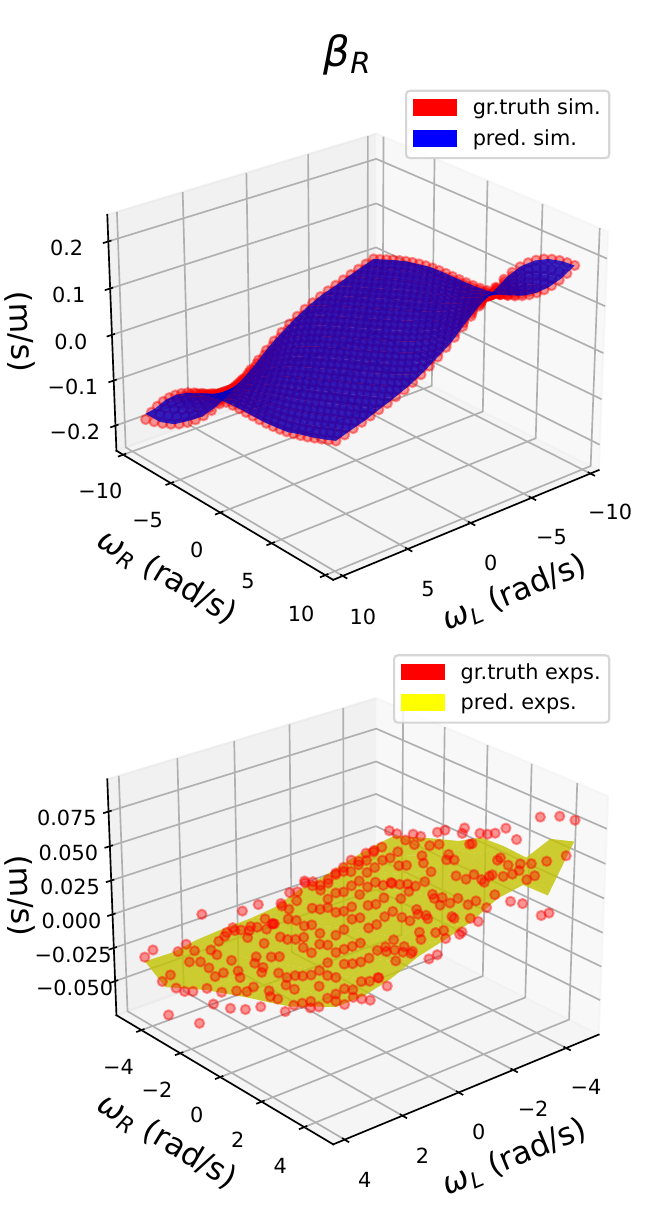}
 \caption{Fitting of the longitudinal slippage parameters $\beta_L$
   (left) and $\beta_R$ (right) as a function of the sprocket wheel
   speeds $\omega_{w,L}$ and $\omega_{w,R}$.  The top row reports
   simulation results, while the bottom row the experimental results.  The
   blue surface and the yellow surface are the fit of the regressor
   predictions for simulation and experiments, respectively, while the
   red dots are the acquired data. }
  \label{fig:beta_ident}
\end{figure}

\DSrev{For the Decision Tree models, the \texttt{CatBoost} library was employed. 
The model parameters were set as follows: a maximum depth of 6, 1000 iterations, a maximum of 31 leaves, and the Root Mean Squared Error (RMSE) as the loss function.
For the Neural Network, a Multi-Layer Perceptron (MLP) was implemented using the \texttt{PyTorch Lightning} framework. 
The network architecture consists of four fully connected (\texttt{Linear}) layers with dimensions $[5,\,128,\,256,\,128,\,3]$. 
Each layer, except the last, is followed by a ReLU activation function and a Dropout layer with a probability of 0.2. 
All input features were standardized by removing the mean and scaling to unit variance. 
The model was trained using the Mean Squared Error (MSE) loss function and a learning rate of $1\times10^{-4}$.}
The shortcoming of neural networks is that they do
not capture precisely the zero crossing of $\alpha$ for low angular
velocities (e.g., wheel velocities almost equal) and have an offset
around zero. Decision trees do not have this issue with the difference
that are not able to extrapolate meaningful results out of the
training region.  This may lead to inadequate compensation for
slippage; however, it is not an issue if we collect enough samples in
the velocity range of interest.  For the above arguments, we chose to
employ decision trees.  For the collection of the dataset for
identification, we uniformly sample wheel speeds in the set
$\RSI{-10}{10}{\radian\per\second}$ with increments
$\SSI{0.65}{\radian\per\second}$ and send them in open loop to the
robot.  In the accompanying video, we show the data collection
procedure.
We kept the speed constant for each pair
$(\omega_{w,L}, \omega_{w,R})$ for $2$~s to achieve a steady state,
estimated $\beta_L$, $\beta_R$ with~\eqref{eq:beta_estimate} and then
and averaged the estimates over the intervals. Since the regressor
provided jerky outputs when trained with a low number of samples, we
perform upsampling to increase the number of training samples by
interpolating a continuous and smooth function over a denser grid
(with $0.1$~rad resolution), using a \gls{rbf} interpolator with a
smoothing factor of $0.1$. 
We train the regressor on the data thus
acquired and we compare its predictions — sampled from a test set
distinct from the training set — against the corresponding
ground-truth values for the left and right tracks, respectively, in
both simulation and indoor experiments. The results of this comparison
is pictorially reported in Fig.~\ref{fig:beta_ident}.
The simulation was performed in an open-loop fashion using the
distributed parameter model~\eqref{eq:3D_distributed_model}, assuming
flat ground with a friction coefficient of $\mu = 0.1$ and shear
deformation modulus $K=0.001$ (these values were chosen to match the
experimental results).  The simulation results are shown in
Fig.~\ref{fig:beta_ident} (top row).  \DSrev{ Since the \gls{rbf} are
  able to manage high-dimensional inputs, it is possible to directly
  perform predictions directly querying the \gls{rbf} interpolator for
  a single input. The \gls{rbf} interpolator provides predictions very
  similar to the decision trees.  The drawback is that the full
  dataset should be loaded at runtime before each experiment, with a
  delay of about $1$~s.}

In the real experiments the speed range has been limited to the range
$\{-4.6, 4.6\}$~rad/s due to actuation limits of the robot platform
that prevented us to go at higher speed.  The experimental results are
shown in Fig.~\ref{fig:beta_ident} (bottom row). The strong agreement
between the regressor's predictions and the acquired data is
demonstrated both in simulation (coefficient of determination $R^2$
for $\beta_L = 0.999$, $\beta_R = 0.999$) and in real experiments
($R^2$ for $\beta_L = 0.91$, $\beta_R = 0.84$).

\subsection{Lateral slippage}
\label{sec:lat_slippage_ident}

\begin{figure}[t]
 \centering
 \includegraphics[width=1.0\columnwidth]{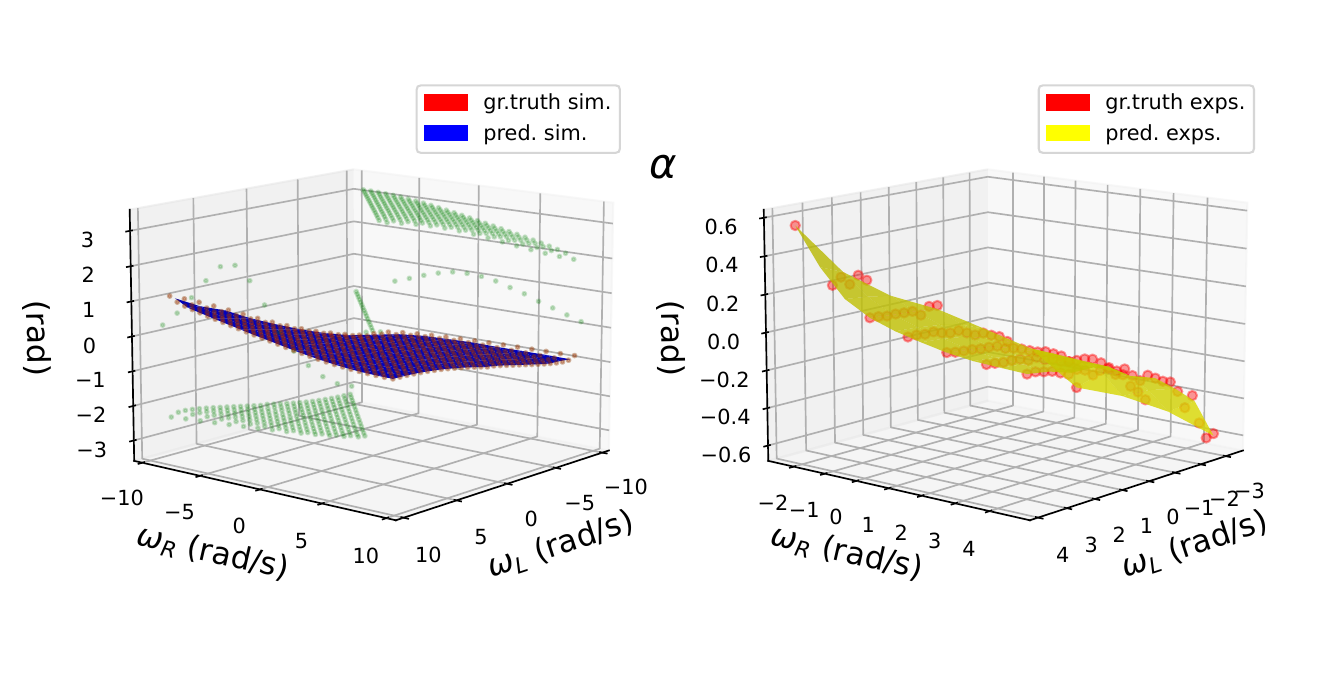}
 \caption{Fitting of the lateral slippage parameter $\alpha$ as a
   function of the sprocket wheel speeds $\omega_{w,L}$ and
   $\omega_{w,R}$ for simulation (left) and real experiments (right),
   respectively.  The blue and the yellow surface are the fit of the
   regressor predictions for simulation and experiments, respectively,
   while the red dots are the acquired data correspondent to positive
   longitudinal velocities ${}_bv_x > 0$ that are considered for the fitting. 
   The samples corresponding to both
   positive and negative longitudinal velocities are presented in
   green just for comparison.}
  \label{fig:alpha_ident}
\end{figure}

The same identification dataset is employed to identify  
the  function $f_{\alpha}$ to estimate the lateral slippage $\alpha$
\begin{equation}
 \alpha = f_{\alpha}\left(\omega_{w,L},\omega_{w,R}\right) \text{.}
\label{eq:falpha}
\end{equation}
As can be seen in the green plot in Fig.~\ref{fig:alpha_ident} (left)
there is a big discontinuity in the ground-truth data.
This can still be captured by the function approximator at the price
of low accuracy in the boundaries of the discontinuity. By inspecting
the data, the discontinuity happens when the longitudinal velocity ${}_bv_x$
changes sign (\eg{} from forward to backward motion). In this case,
due to the definition of $\alpha$, there is a flip of
$\pi\,\USI{\radian}$ (or $-\pi\,\USI{\radian}$), which is responsible
for the discontinuity. As mentioned, in the discontinuity region,
i.e., low speed, the identification accuracy decreases. To account for
this issue, we opted to split the training data to identify two
separate function approximators: one for ${}_bv_x > 0$ and another for
${}_bv_x<0$, this enabled us to have good $\alpha$ prediction also at a low
value of speed ${}_bv_x$. In Fig.~\ref{fig:alpha_ident}(left), we report the
identification results for the simulations (with $\mu = 0.1$ and
${}_bv_x > 0$), while, in Fig.~\ref{fig:alpha_ident}(right), the results for
real experiments show in both cases a good fit on the validation test
set (i.e., $R^2=0.999$ and $R^2=0.961$, respectively).

\begin{figure}[t]
 \centering
 \includegraphics[width=0.7\columnwidth]{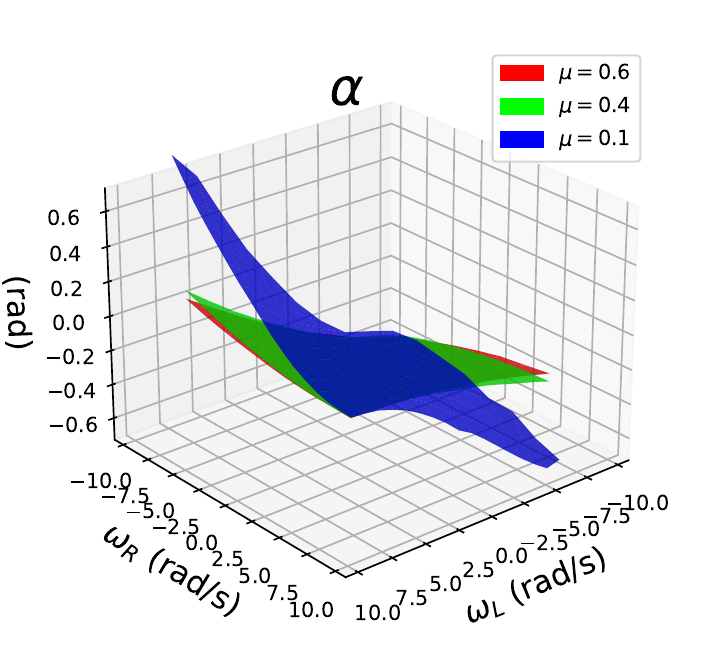}
 \caption{Simulation: Plots of lateral slippage $\alpha$ as a function
   of the sprocket wheel speeds $\omega_{w,L}$ and $\omega_{w,R}$ for
   a friction coefficient $\mu=0.1$ (blue plot) $\mu=0.4$ (green
   plot), and $\mu=0.6$ (red plot) for positive values of the
   longitudinal velocity ${}_bv_x$. }
  \label{fig:different_friction_ident_sim}
\end{figure}

In simulation, we repeated the identification for three different
values of the friction coefficient $\mu \in\{0.1, 0.4, 0.6\}$. The
results are reported in Fig.~\ref{fig:different_friction_ident_sim},
which shows that lower values of $\mu$, i.e., a more slippery surface,
reflect in higher values of $\alpha$ for the same speed range.
Note that the plots for $\mu = 0.4$ and $\mu = 0.6$ are very similar,
highlighting the highly nonlinear behavior of the system.

\subsection{Experimental validation}
\begin{figure}[t]
  \centering
  \includegraphics[width=1.0\columnwidth]{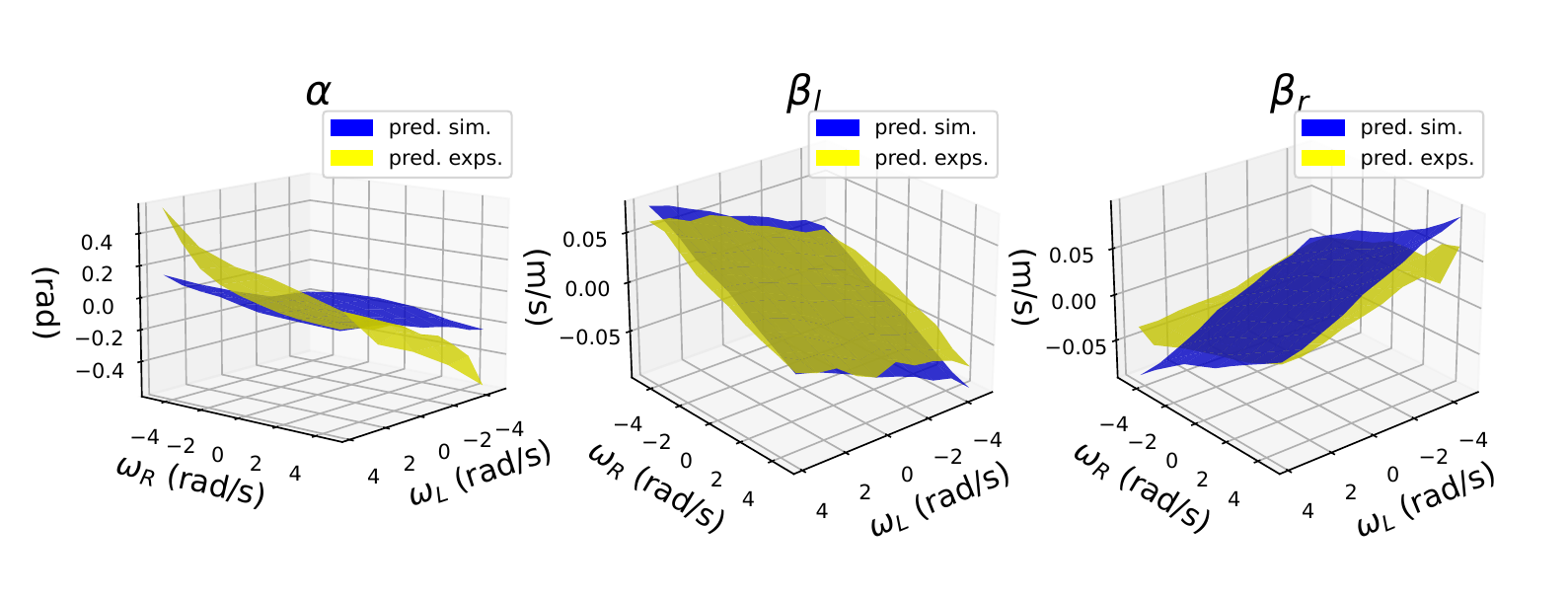}
  \caption{Comparison of the predictions of the slippage parameters
    $\alpha$ (left), $\beta_L$(middle), $\beta_R$(right) for
    simulation (blue) and experiments (yellow). }
  \label{fig:comparison}
\end{figure}

We compare in Fig. \ref{fig:comparison} the plots of the identified
surfaces  for simulation (blue) and experiments (yellow).
For a fair comparison the simulation data is shown in the same
(reduced) speed range $\{-4.6, 4.6\}$~rad/s as in the experiments.
The results present a good quantitative matching for $\beta$
confirming that the simulation accurately captures the trends observed
in real experiments.  However the matching is only qualitatively in
the case of $\alpha$.  We conjecture that this is due to the non
trivial rubber-rigid floor interaction, suggesting the need to perform
further outdoor experiments.  The validation of the simulation model
opens to simulation results only, while the experimental validation is
postponed to future work.

\subsection{Identification on slopes:  influence of gravity  on slippage parameters}
\label{sec:3d_ident}
When moving on slopes, the primary difference compared to flat terrain
is the influence of gravity, which affects the direction of $F_y$ that
may no longer be directed outward from the curve but rather opposite
to the component of gravity parallel to the terrain. As a result,
lateral slippage becomes proportional to
$\left\Vert F_{g, \mathbin{\!/\mkern-5mu/} }-[F_t, F_y, 0]^{\top}
\right\Vert$. Therefore, when operating on slopes, the identifications
presented in Sections~\ref{sec:long_slippage_ident}
and~\ref{sec:lat_slippage_ident} should be repeated, incorporating the
unit vector ${}_b\hat{\vect{g}}$ representing the direction of the
gravity vector (mapped in the based frame) as an additional input to
the function approximator, along with the wheel speeds
\begin{equation}
 \alpha = f_{\alpha}\left(\omega_{w,L},\omega_{w,R}, {}_b\hat{\vect{g}} \right), \quad   \mat{\beta_L\\ \beta_R}  = f_{\beta}\left(\omega_{w,L},\omega_{w,R}, {}_b\hat{\vect{g}} \right).
\label{eq:falpha_fbeta3D}
\end{equation}

This is in accordance to \cite{yamamuchi} that claims the robot rotation behavior changes with the robot inclination on the slope. 
The function approximator should be trained computing estimates
of longitudinal and lateral slippages~\eqref{eq:alpha_estimate},
\eqref{eq:beta_estimate} employing ${}_bv_x$, ${}_bv_y$ in the local (base) frame. The rationale behind this choice will become clear in Section~\ref{sec:3D_control}.
%
%
For training, we conducted the same tests as on flat terrain (\ie{} uniformly varying wheel speeds within their bounds) but with the robot moving on ramps. We collected samples by repeating simulations for ramps with different inclinations, ranging from 0 to $\SSI{-0.3}{\radian}$ in increments of 0.5, storing ${}_b\hat{\vect{g}}$ and the wheel speed values. For these tests, we set a friction coefficient of $\mu=0.6$ and employed the model in~\eqref{eq:3D_distributed_model}. The identification was performed only in \textit{simulation} for both the uniform and nonuniform load distribution models.

%% file: sections/5_control.tex

A Lyapunov-based control law design is presented in this section. We briefly subsume a trajectory following control law for the \emph{unicycle}-like model~\eqref{eq:unicycle_kinematics}, which is then adapted to the case of the tracked robot model~\eqref{eq:tracked_vehicle_kinematics}. The development is initially performed for a flat (horizontal) terrain assumption and will be extended to the 3D case in Section \ref{sec:3D_control}.

\paragraph{Useful Definitions} In the context of trajectory tracking,
we denote with  the $d$ superscript (\eg{} $(x^d, y^d, \varphi^d)$ or $(v^d, \omega_z^d)$)  desired quantities,
which act as reference for the system state $(x,\,y,\,\varphi)$. \textbf{To not overload the  notation, only for this section, we will consider }$\mathbf{v = {}_bv_x}$ and $\mathbf{v^d = {}_bv_x^d}$. The desired trajectory will be identified by the following virtual  model
\begin{equation}
 \begin{aligned}
   &\dot{x}^d = v^d \mcos{\varphi^d}\\
   &\dot{y}^d = v^d \msin{\varphi^d}\\
   &\dot{\varphi}^d = \omega_z^d\\
 \end{aligned}
 \label{eq:unicycle_desired}
\end{equation}
where the initial conditions $x^d(0)$, $y^d(0)$, $\varphi^d(0)$ are assigned and
$v^d(t)$ and $\omega_z^d(t)$ are two known functions.
It is therefore useful to define the error vector
\begin{equation}
  \vect{e} = \begin{bmatrix}
    e_x\\
    e_y\\
    e_\varphi
  \end{bmatrix} = \begin{bmatrix}
    x - x^d \\
    y - y^d \\
    \varphi - \varphi^d
  \end{bmatrix} .
  \label{eq:error}
\end{equation}
In the remaining of this section, we use the following definitions
\begin{equation}
  \begin{aligned} 
    &e_{xy} = \left\Vert \begin{bmatrix} e_x \\ e_y \end{bmatrix}\right\Vert =
    \sqrt{e_x^2 + e_y^2} , \quad
    &&\psi = \matan{e_y}{e_x} ,\\
    &e_{x} = e_{xy} \mcos{\psi} ,\quad &&e_{y} = e_{xy} \msin{\psi} .
  \end{aligned}
  \label{eq:err2}
\end{equation}

\subsection{Lyapunov-based control for a unicycle}
\label{ref:LyapBased}

The following control law for a unicycle-like vehicle will be used as
a reference for the tracked vehicle control.  Given the dynamics~\eqref{eq:unicycle_kinematics}, we aim to develop a control law that ensures the convergence of the actual trajectory to the desired one. To this end, consider the Lyapunov function
\begin{equation}
  V = \frac{1}{2}\left(e_x^2 + e_y^2\right) + \left(1 -
    \mcos{e_\varphi}\right) ,
  \label{eq:lyap}
\end{equation}
which is positive definite and radially unbounded for $e_\varphi \in [-\pi,\,\pi]$, and whose time derivative is
\begin{equation}
  \begin{aligned}
    \dot{V} = & e_x \dot{e}_x + e_y \dot{e}_y + \msin{e_\varphi} \dot{e}_\varphi = e_x \left(v \mcos{\varphi} - v^d \mcos{\varphi^d}\right) + \\
    &+ e_y \left(v \msin{\varphi}- {}_bv_x^d \msin{\varphi^d}\right) +
    \msin{e_\varphi} (\omega_z-\omega_z^d) .
  \end{aligned}
  \label{eq:lyapDer}
\end{equation}
By considering the auxiliary inputs $\delta v$ and $\delta \omega_z$ in
a back-stepping fashion, we have
\begin{align}
  v = v^d + \delta v, \quad \omega_z = \omega_z^d + \delta \omega_z ,
  \label{eq:control_input_unicycle}
\end{align}
and, by simple trigonometric manipulations and after substitution of
the dynamics of the error~\eqref{eq:error} along the system
dynamics~\eqref{eq:unicycle_kinematics}, leads to
\[
  \begin{aligned}
    \dot{V}\!\! = & - 2 v^d e_x \msin{\frac{e_\varphi}{2}}\msin{\frac{\beta} {2}}\!\! +\!\! 2 v^d e_y \msin{\frac{e_\varphi}{2}}\mcos{\frac{\beta} {2}}\!\! + \\
    &+ \delta v \left(e_x \mcos{\varphi} + e_y \msin{\varphi}\right) +
    \msin{e_\varphi} \delta \omega_z ,
  \end{aligned}
\]
with $\beta$ being defined as $\beta = \varphi + \varphi^d$.
Using~\eqref{eq:err2} and after some algebraic manipulation, we have,

\begin{equation}
  \begin{aligned}
    \dot{V} & = 2 v^d e_{xy} \msin{\frac{e_\varphi}{2}} \msin{\psi - \frac{\beta}{2}} + \msin{e_\varphi} \delta \omega_z + \\
    & + \delta v e_{xy} \mcos{\psi - \varphi} \text{,}
  \end{aligned}
  \label{eq:lyapDer1}
\end{equation}

and hence, by setting the auxiliary inputs
\begin{equation}
  \begin{aligned}
    \delta v & = - k_{p}e_{xy} \mcos{\psi-\varphi} \text{,}\\
    \delta \omega_z &= \frac{1}{ \msin{e_\varphi}} \left( - 2 v^d e_{xy} \msin{\frac{e_\varphi}{2}} \msin{\psi - \frac{\beta}{2}} \right) - k_{\varphi} \msin{e_{\varphi}} \\
    & = - v^d e_{xy} \frac{1}{\mcos{\frac{e_\varphi}{2}}} \msin{\psi -
      \frac{\beta}{2}} - k_{\varphi} \msin{e_{\varphi}} \text{,}
  \end{aligned}
  \label{eq:delta_control_unicycle}
\end{equation}
with $k_{p}>0$ and $k_{\varphi}>0$ design parameters, we have
\begin{equation}
  \dot{V} = -k_p e_{xy}^2 \cos^2 \left(\varphi - \psi\right) - k_{\varphi} \sin^2 \left(e_\varphi\right) \text{,}
\end{equation}
which is negative semi-definite. More precisely,
\begin{align*}
  \dot{V} = 0 ~ \text{ when } \begin{array}{c}
    e_{xy} = 0 \text{,} ~ e_{\varphi} = 0 \text{,}\\
    e_{\varphi} = 0 \mbox{, and } \varphi = \psi \pm \frac{\pi}{2}\text{.}
  \end{array}
\end{align*}
The first case is the desired equilibrium.  In the second case, by
applying~\eqref{eq:control_input_unicycle}, we have $\delta v = 0$ and
$
  \delta \omega_z =  - v^d e_{xy} \sin\left(\psi - \frac{\beta}{2}\right)
  = - v^d e_{xy} \sin\left(\frac{e_{\varphi}}{2} \pm
    \frac{\pi}{2}\right) = \pm v^d e_{xy} \text{,}
$
which is an equilibrium point only if $e_{xy} = 0$. This proves global
asymptotic stability by Lasalle theorem.

\subsection{Lyapunov-based control of the tracked vehicle}

\begin{figure}[t] \centering
 \includegraphics[width=1.0\columnwidth]{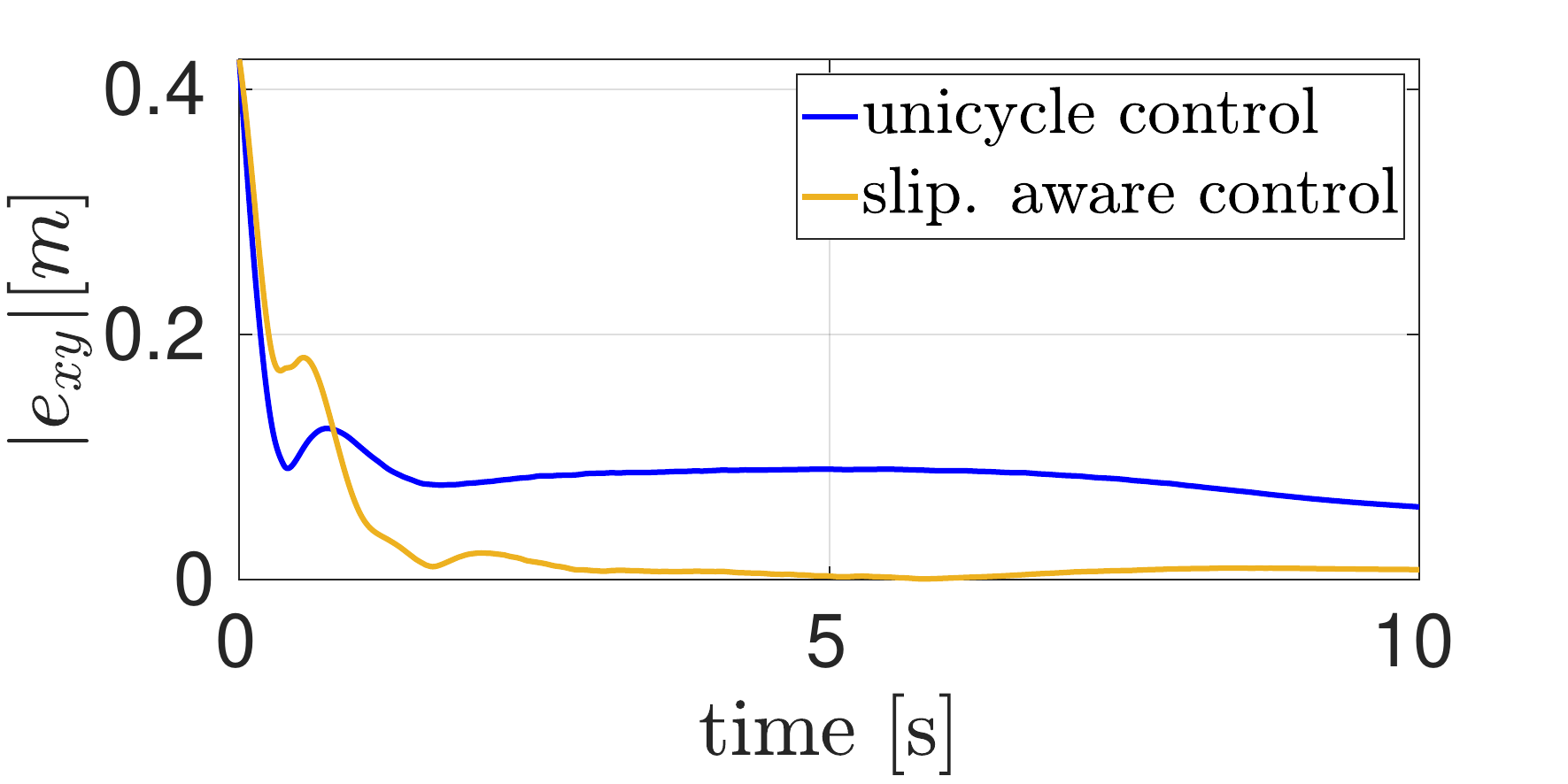}
  \caption{Simulation with pseudo-kinematic model. Cartesian tracking error for a spiral-like reference trajectory. 
  The blue line is the unicycle controller, the yellow line is the slippage-aware controller.}
   \label{fig:matlab_sim}
\end{figure}

Let us consider the dynamics~\eqref{eq:tracked_vehicle_kinematics}
where we modify the inputs~\eqref{eq:control_input_unicycle} to account for the lateral slippage
$\alpha$, \ie{}
\begin{equation}
  v = (v^d + \delta v) \mcos{\alpha}, \quad \omega_z = \omega_z^d + \delta
  \omega_z .
  \label{eq:control_input_tracked}
\end{equation}
We also modify the Lyapunov function~\eqref{eq:lyap} as follows
\begin{equation}
  V = \frac{1}{2}\left(e_x^2 + e_y^2\right) + \left(1 - \mcos{e_\varphi+\alpha^d}\right)
  \label{eq:lyap3}
\end{equation}
where $\alpha^d = f_{\alpha}\left(\omega^d_{w,L}, \omega^d_{w,R}\right)$, with $f(\cdot)$
that could be the function approximator defined in Section~\ref{sec:lat_slippage_ident}
and where the desired references $v^d$, $\omega_z^d$ are mapped  into $\omega^d_{w,L}, \omega^d_{w,R}$ and then to $\alpha^d$.
The function~\eqref{eq:lyap3} is positive
definite and has an equilibrium point in $e_{xy}=0$ and
$e_\varphi = -\alpha^d$, hence having a tracking reference for $\varphi$
in $\varphi^d - \alpha^d$ and giving up the possibility to track
$\varphi^d$.  This is somewhat expected form the
dynamics~\eqref{eq:tracked_vehicle_kinematics}: the effect of $\alpha$
is to make it impossible to simultaneously track $x^d$, $y^d$ and
$\varphi^d$. Indeed, if $\varphi^d = \varphi$, it is impossible to
generate a velocity that keeps $x$ and $y$ on $x^d$ and $y^d$, which
are generated by unicycle-like dynamics (hence, with $\alpha =
0$). However, from a practical point of view, it is not an issue to track
$\varphi^d - \alpha^d$ instead of $\varphi^d$, since an appropriate
reference for $\varphi^d + \alpha^d$ could be designed at the planning level, based on
desired values of $v^d$ and $\omega_z^d$. By choosing the two control
values as per Equation~\eqref{eq:control_input_tracked}, following the same
rationale of Section~\ref{ref:LyapBased}, we find that the derivative
of the Lyapunov function~\eqref{eq:lyap3}, obtained after some basic
trigonometric simplifications is given by
\begin{equation}
  \begin{aligned}
    \dot{V} & = e_x \dot{e}_x + e_y \dot{e}_y + \msin{e_\varphi +
      \alpha^d} \left(\dot{e}_\varphi + \dot{\alpha}^d\right) \\
    & = 2 v^d e_{xy} \msin{\frac{\alpha + e_\varphi}{2}} \msin{\psi -
      \frac{\alpha + \beta}{2}} + \\
    & + \delta v e_{xy} \mcos{\psi - (\alpha + \varphi)} +
    \msin{e_\varphi+\alpha^d} \left( \delta\omega_z
      +\dot{\alpha}^d\right) ,
\end{aligned}
\end{equation}
where again we set $\beta = \varphi + \varphi^d$. If we choose the new
auxiliary controllers as
\begin{equation}
  \begin{aligned}
    \delta v & = - k_{p} e_{xy} \mcos{\psi-(\varphi+\alpha)} , \\
    \delta \omega_z & = - v^d e_{xy}
    \frac{1}{\mcos{\frac{\alpha+e_\varphi}{2}}} \msin{\psi -
      \frac{\alpha+\beta}{2}} - k_{\varphi} \msin{e_{\varphi}+\alpha^d}
    - \dot{\alpha}^d ,
    \label{eq:delta_control_tracked}
  \end{aligned}
\end{equation}
we finally have
\begin{equation}
  \begin{aligned}
    &\dot{V} = -k_p e_{xy}^2 \cos^2\left(\psi - (\alpha +
      \varphi)\right) - k_\varphi \sin^2 \left(e_\varphi+\alpha^d\right).
  \end{aligned}
  \label{eq:lyapDer2_4}
\end{equation}
which is negative semidefinite. Using Lasalle arguments, we can prove
convergence to the equilibrium $e_{xy}=0$ and $e_\varphi = -\alpha^d$.
Note that for the estimation of $\alpha=f\left(\omega_{w,L}, \omega_{w,R}\right)$
in~\eqref{eq:control_input_tracked}, actual values should be used.
%
To show the global stability of the pseudo-kinematic model~\eqref{eq:tracked_vehicle_kinematics}
controlled by the designed controller~\eqref{eq:control_input_tracked},~\eqref{eq:control_input_tracked}
we performed a simulation (see Fig.~\ref{fig:matlab_sim}) where  the aim is to track a spiral-like reference trajectory (constant $v^d= 0.3 m/s$) of decreasing
radius of curvature ($\omega_z^d = 4t, t>0$) (\ie{} where $\dot{\alpha}^d$ is not null) starting with an
initial state $[x_0, y_0, \varphi_0]^\top = [0.3, 0.3, 0.2]^\top$ different from the initial reference state $[0., 0., 0.]^\top$. 
The figure shows the increased convergence performances given by the slippage-aware
controller that accounts for the slippage present in the model with respect to a standard
Lyapunov controller designed for a unicycle.

\paragraph{Limitations}
In practical applications, zero convergence of the tracking error is impossible due to the 
unavoidable mismatch between the pseudo-kinematic model and the real robot dynamics (or the more accurate dynamic model, in the case of simulation).   
Additionally, the parameters  $\alpha$  and $\beta_L$, $\beta_R$, estimated by the approximating functions, can be uncertain and terrain-dependent. In practice, they can be determined through visual odometry (\eg{} using relative measurements obtained by triangulation from inertially fixed features such as trees). The impact of uncertainty in the model and/or in the estimation of  $\alpha$ on tracking accuracy can be assessed by analyzing the size of the control-invariant set on the level set of the derivative of the Lyapunov function—specifically (\ie{} the set within which $\dot{V}$  is no longer negative definite) for a certain bound of uncertainty on the parameters. It can be shown that this set is convex and compact. However, computing the sensitivity of the invariant set as a function of the uncertainty in the slippage parameters is beyond the scope of this work and is left for future research. As an alternative, in a separate study~\cite{bussola24irim}, we analyzed how a set of tracked vehicles operating within a given area can collect noisy estimates of the slippage parameters and reach a consensus that enhances tracking accuracy.

\subsection{Controller extension to 3D}
\label{sec:3D_control}
The purpose of this section is to illustrate how the controller designed for a 2D environment could be extended to 3D. We assume the controller~\eqref{eq:control_input_tracked} is defined on local variables (\ie{} the body frame or Frenet frame). This is the reason in Section~\ref{sec:3d_ident} slippages have been estimated in a local (body) frame rather than in the world frame. In this case, it only suffices to map the tracking errors~\eqref{eq:error} from the world frame into the (local) body frame
\begin{align}
	{}_B\vect{e} = {}_{W}R_B^\top\mat{e_x\\ e_y\\0}, \quad
	{}_Be_\varphi = z_B^\top\mat{0\\0\\ \varphi -\varphi^d}
	\label{eq:frenet_error}
\end{align}
hence, the error $\vect{e}$ to be used in~\eqref{eq:control_input_tracked} should be $[{}_B\vect{e}_x, {}_B\vect{e}_y, {}_Be_\varphi]$.
No additional mapping is required since the outputs $v={}_bv_x$ and $\omega_z$ in~\eqref{eq:control_input_tracked} are already defined in the body frame. Therefore, 
equation~\eqref{eq:tracked_vehicle_mapping} can be directly inverted to compute the reference wheel speeds. The reference trajectory  $x^d, y^d, \varphi^d$ remains defined in 
the world frame. As a result, the modified controller is expected to naturally follow the terrain shape while tracking the reference.

%% file: sections/6_planning.tex
In this section, we will present  \textit{point to point} planning strategies (of different levels of complexity) that can be used to generate references
for the two  controllers presented in Section~\ref{sec:control}: (1) the unicycle  controller~\eqref{eq:control_input_unicycle} and (2) the slippage-aware controller~\eqref{eq:control_input_tracked}. 
The purpose of this section is not to be exhaustive, but rather to provide an introduction to the problem of motion planning for tracked vehicles. It presents two state-of-the-art approaches and introduces a new method that explicitly accounts for slippage. The advantages and limitations of each approach will be quantitatively evaluated in Section~\ref{sec:plan_eval}, in terms of performance degradation under slippage conditions. More precisely our goal is to compare:
\begin{enumerate}[label=\Alph*), itemsep=0pt, topsep=0pt, parsep=0pt, partopsep=0pt]
    \item a baseline planning strategy based on Dubins curves, which has the assumption of constant velocity. Dubins curves are piecewise  minimum time optimal curves that can be analytically derived \cite{Laumond1998},
    \item a planning strategy based on $G_1$ clothoids, with linearly variable curvature \cite{clothoids},
    \item a slippage-aware numerical strategy based on direct optimal control which exploits the pseudo-kinematic model \eqref{eq:tracked_vehicle_kinematics} of the tracked vehicle.
\end{enumerate}
As in the case of the slippage-aware controller, the slippage parameters are estimated from sprocket wheel speeds via the function approximators identified in Section~\ref{sec:slippage_identification}. For simplicity, we consider trajectory planning on a horizontally flat ground, the generalization to slopes being trivial as in the case of the controller. The planning based on Dubins and Clothoids trajectories has the advantage of relying on \emph{closed form} solutions that involve little computation time but rely on a unicycle model approximation, hence \textit{do not account} for slippage.

\subsection{Dubins planning}
Dubins \cite{dubins_paper} showed that the minimum-time trajectory for a mobile robot moving from an initial configuration  $\vect{c}_0$ =  ($x_0^d$, $y_0^d$, $\varphi_0^d$) to a final configuration $\vect{c}_f$ = ($x_f^d$, $y_f^d$,$\varphi_f^d$)  under the assumptions of (1) constant forward speed and (2) bounded path curvature (i.e., a minimum turning radius), consists of at most three segments selected from six basic motion primitives: straight line segments and arcs of circles.

To obtain the Dubins solution we only need to define the (constant) longitudinal velocity ${}_bv_x^d$ and the maximum curvature $k=\omega_z^d/{}_bv_x^d$.
This result has been obtained by solving an indirect optimal control problem, that in this specific case,  can be solved in \textit{closed} form. This provides the references for the robot position $x^d, y^d$ and orientation $\varphi^d$, for  the whole duration of the trajectory $T_f=L/{}_bv_x^d$, where ${}_bv_x^d$ is the chosen (constant) longitudinal speed and $L$ is the sum of the 3 arc lengths from the Dubins solution.  To determine the bounds on the maximum curvature $k$, we formulate and solve a linear program that computes the feasible range of $k$ given the desired velocity ${}_bv_x^d$ and the bounds on the sprocket wheel speeds. The Dubins solution provides also the corresponding vectors of ${}_bv_x^d$, $\omega_z^d$ that can be used as desired reference velocities for the controller. The drawbacks of this planning approach are that: 1) involves discontinuities in the time evolution of the curvature  and 2) requires a constant longitudinal velocity.

\subsection{Clothoids planning} Another class of minimum time curves that avoids the discontinuities in the 
curvature and assumes a linear variation of it, is the $G_1$ Clothoid \cite{clothoids}.
The curve is generated in curvilinear space, and the reference trajectory is obtained by mapping it to time,
assuming a given discretization step  $\mathrm{d}T$  and a specified longitudinal velocity.
The procedure is as follows: using the Clothoid library \cite{clothoids_library}, we generate the curve
on curvilinear coordinates $s$ with a total length $L$. We then discretize it into $N=L/({}_bv_x^d \mathrm{d}T)$ samples.
Then for each sample, we evaluate the curve $x(s)$, $y(s)$, $\varphi(s)$ and its derivatives   
$\mathrm{d}x(s)/\mathrm{d}s$, $\mathrm{d}y(s)/\mathrm{d}s$, $d\varphi(s)/\mathrm{d}s$,
assuming that we want to follow the curve at a constant speed ${}_bv_x^d=\mathrm{d}s/\mathrm{d}t$ the mapping is
\begin{align}
	\dot{x}& = \frac{\mathrm{d}x(s)}{\mathrm{d}s}{}_bv_x^d \text{,}  &x =\int \dot{x} \, \mathrm{d}t \text{,}\nonumber  \\
	\dot{y}& = \frac{\mathrm{d}y(s)}{\mathrm{d}s}{}_bv_x^d \text{,}  &y =\int \dot{y} \, \mathrm{d}t \text{,} \\
	\dot{\varphi}& = \omega_z=  \frac{d\varphi(s)}{\mathrm{d}s}{}_bv_x^d \text{,} &\varphi =\int \dot{\varphi}\, \mathrm{d}t \text{.}\nonumber
\end{align}
\subsection{Slippage-aware planning}
The slippage-aware planner is based on an optimal control formulation that makes use of the pseudo-kinematic model~\eqref{eq:tracked_vehicle_kinematics} leveraging lateral/longitudinal slippage estimates. 
 Optimal control produces trajectories coherent with the given model
while satisfying constraints and boundary conditions and optimizing a performance index. 
We employ a \emph{single shooting} approach to transcribe the optimal control problem into a \gls{nlp}. Specifically, we discretize the vector of sprocket wheel inputs, $\vect{U}=[\vect{u}_0, \dots, \vect{u}_{{N-1}}]^\top$
into $N={T_f}/{dT}$ intervals, where $\vect{u}_k = [\omega_{{w,L}_k}, \omega_{{w,R}_k}]^\top$ and $T_f$  is the duration of the trajectory.
The decision variables are: (1)   the control inputs $\vect{U}$ and (2) the trajectory duration $T_f$. These are mapped into ${}_bv_x$, $\omega_z$ through~\eqref{eq:tracked_vehicle_mapping} and  are given as inputs to~\eqref{eq:control_input_tracked}, while the states  $\vect{x}(k) \in [0, N]$ with $\vect{x}_k = [x_k, y_k, \varphi_k] \in \Rnum^3$  are computed as \emph{dependent} variables from the inputs, along the horizon. Likewise  $\beta_{L,k}$, $\beta_{R,k}$, $\alpha_k$ are  computed from the inputs  $\vect{U}$  via~\eqref{eq:fbeta},~\eqref{eq:falpha}.
The resulting optimization problem is
\begin{subequations}
	\label{eq:nlp_formulation}
	\begin{align}
		\displaystyle{\min_{\vect{U}, T_f }}\quad
		&  w_t T_f + w_s \sum_{k=0}^{N-1}  \left({}_bv_{x,{k+1}} - {}_bv_{x,{k}}\right) + \left(\omega_{z,{k+1}} - \omega_{z,k} \right)  \label{eq:cost_function} \\
		\text{s.t.} \quad & \vect{x}_{0}=\vect{c}_0,      &  \label{eq:initial_condition}\\
		&\vect{x}_{k+1}=f\left(\vect{x}_{k}, \vect{u}_{k}\right),\quad \quad  k \in\mathbb{I}_0^{N-1}, \label{eq:equality}\\
		&h\left(\vect{x}_{k}, \vect{u}_{k}\right) \leq 0,       \quad\quad\quad  k\in\mathbb{I}_0^{N-1}, \label{eq:inequality}
	\end{align}
\end{subequations}
where in the left-hand side of the cost function~\eqref{eq:cost_function},
we minimize the duration  $T_f$ of the trajectory\footnote{Note that for a fair comparison with Dubins curves, 
it is essential to minimize the time in the cost.}, while on the right-hand side we add a regularization term to smooth the variations of ${}_bv_x$ and $\omega_z$. 
The initial condition~\eqref{eq:initial_condition}
is expressed by setting $\vect{x}_0$ equal to the initial state.
The pseudo-kinematic model~\eqref{eq:tracked_vehicle_kinematics} is
integrated in~\eqref{eq:equality} to obtain the states in a single-shooting fashion via a high-order method (\ie{} Runge Kutta 4) starting from the initial state at sample~0.
To reduce the impact of integration errors, we integrate on a finer grid,
performing a number of integration $N_{\text{sub}}$ sub-steps
within two adjacent knots. This has the advantage of improving the
integration accuracy, without increasing the problem size.
For the inequalities~\eqref{eq:inequality}, we encode the following constraints:
\begin{enumerate}
\item Bounds on the input variables: $\vert \omega_{w,L}\vert , \vert \omega_{w,R}
\vert  \leq \omega_{w,\max}$,
\item Bounds on the longitudinal speed: $ {}_bv_{x,{\min}} \leq {}_bv_x \leq {}_bv_{x,{\max}}$.
\item No explicit bound is set on $\omega_z$, as it can be derived from 1. and 2.
\item Ensure the target configuration $\vect{c}_f$ is reached at the end of the trajectory, \ie{} when $t =T_f$
\begin{equation}
	\Vert \vect{x}_N - \vect{c}_f\Vert \leq s
\end{equation}
where  $s$ is a fixed slack variable to ensure feasibility.
\end{enumerate}

\begin{table}[ht!]
	\centering
	\caption{ Optimisation parameters}
	{\small\begin{tabular}{lcc}
    \toprule
			\textbf{Name} & \textbf{Symbol} & \textbf{Value} \\
      \midrule
			Discretisation steps (NLP) & $N$ & 40 \\
			\gls{nlp} interval (\USI{\second}) &  $dt$ & 0.01 \\
			Sub-integration steps & $N_{\text{sub}}$ & 10 \\
			Smoothing weight & $w_s$ & 1 \\
			Time weight & $w_t$ & 1 \\
			Integration method & -- & RK4 \\
			Constr. feas. tolerance & $\epsilon$   & 1e-3 \\
			Max. wheel speed & $\omega_{w,\max}$ & 18 \\
			Max. long. speed & $v_{\max}$  & 0.4 \\
			Target Slack & $s$ & 0.02 \\
      Prediction Horizon & $T_f$ & opt. \\
			\bottomrule
	\end{tabular}}
	\label{tab:params}
\end{table}

\textbf{Initialization} We initialize the optimization variables  leveraging the Dubins solution, namely:  ${}_bv_x^d$, $\omega_z^d$  to initialize the decision variables  $\omega_{w,L}$ $\omega_{w,R}$  via~\eqref{eq:unicycle_mapping} mapping, and $T_{f,0} = L/{}_bv_x^d$ to initialize  the trajectory duration.

%% file: sections/7_results.tex

\section{Experiments on flat terrain}
\label{sec:flat_results}
\begin{figure*}[ht!]
  \centering
  \includegraphics[width=0.8\textwidth]{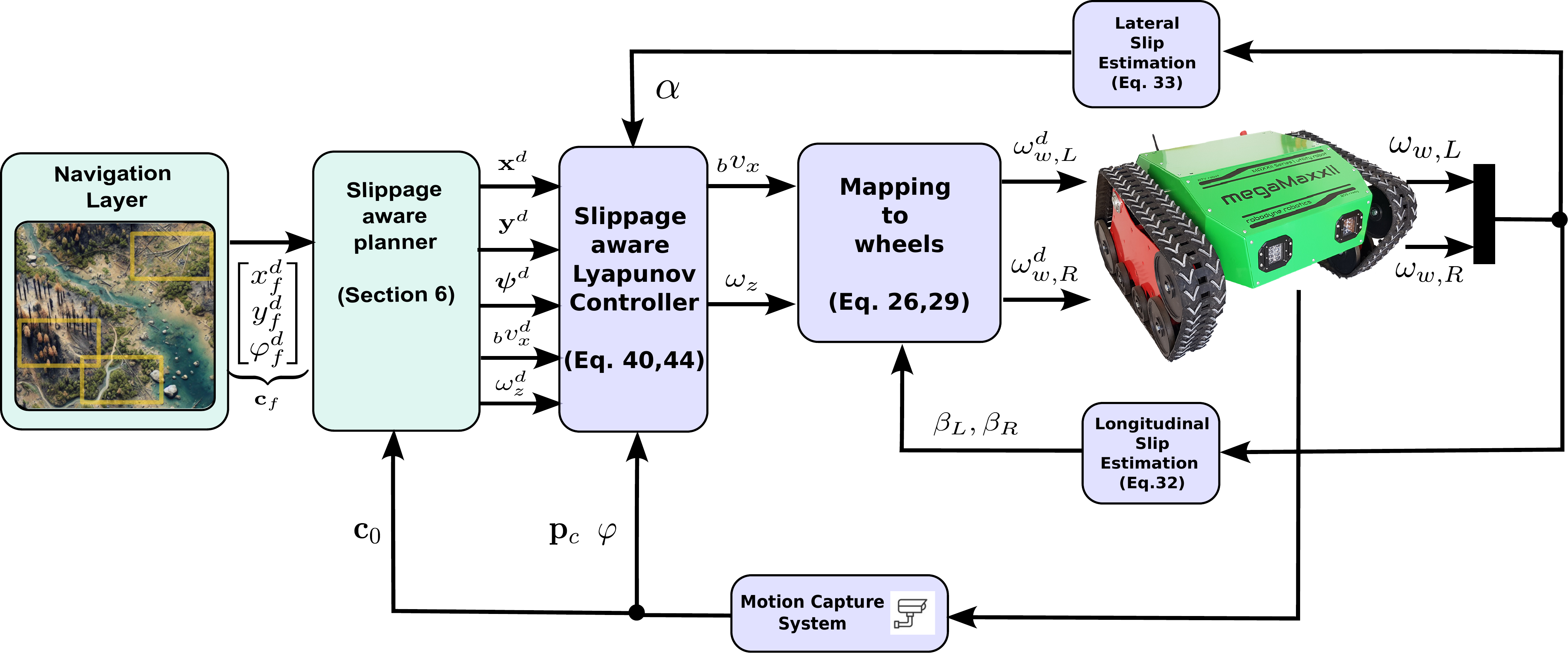}
  \caption{Block diagram of the control pipeline.}
  \label{fig:control_diagram}
\end{figure*}

The block diagram depicted in Fig.~\ref{fig:control_diagram} shows the control and planning pipeline.
Differently from Fig.~\ref{fig:matlab_sim} where we simulate with the pseudo-kinematic model,
in this simulation we use the more physically accurate distributed parameter model~\eqref{eq:2D_distributed_model}.
The  parameters for both simulation models are reported in Table~\ref{tab:sim_params} \DSrev{and resembles the ones of the MAXXII robot.
We report also experimental tests performed with the LIMO robot. Indeed due to physical actuation constraints, the achievable speeds with the  MAXXII robot were too limited to appreciate significant slippages. The physical parameters of the LIMO robot are reported in Table   \ref{tab:limo_params}.}

\begin{table}[ht!]
  \centering
  \caption{Simulation Parameters of Dynamic models }
       \renewcommand{\arraystretch}{1.1}
    \resizebox{\columnwidth}{!}{
    \begin{tabular}{l c c }
    \toprule
    \textbf{Name} \quad & \textbf{Symbol} & \textbf{Value} \\
    \midrule
      Robot mass (\USI{\kilo\gram}) & $m$ & 62 \\
      \gls{com} Height (\USI{\meter})                           & h                                         & 0.25    \\
      Inertia moment (\USI{\kilo\gram\meter\squared}) & $I_{zz}$ & 4.5 \\
      Inertia tensor  (\USI{\kilo\gram\meter\squared})         & ${}_bI$ & $\mathrm{diag}\left(2.18, 2.81, 4.5\right)$      \\ 
      Track width (\USI{\meter}) & $B$ & 0.606 \\
      Track length (\USI{\meter}) & $2\ell$ & 0.7 \\
      Track area (\USI{\meter\squared}) & $A_t$ & 0.07 \\
      Number of Track patches longitudinal (--)     &   -    & 10 \\
      Number of Track patches lateral (--)   &   -    & 4 \\
      Track patch area (\USI{\meter\squared}) & $A_p$ & 0.00175 \\
      Sprocket radius (\USI{\meter}) & $r$ & 0.0856 \\
      Rolling friction coefficient (--)       & $c_r$                                         & 0.025    \\         
      Cohesion  & $c$ (\USI{\pascal})                              &   0 \\
      Shear deformation modulus & $K$ (--)                              &   0.001 \\
      Nominal terrain  stiffness (\USI{\newton\per\meter})& $K_{t}$  &  1e05           \\
      Nominal terrain  damping (\USI{\newton\second\per\meter})   & $D_{t}$    & 0.5e04\\
      Speed dependent stiffness increase  (\USI{\newton\second\per\meter\squared})  & $K_{t_p}$                &           10.5 \\
      Simulation time interval (\USI{\second}) & $dt_\text{sim}$ & 0.001 \\
    \bottomrule
  \end{tabular}}
  \label{tab:sim_params}
\end{table}

\DSrev{ 
\begin{table}[ht!]
  \centering
  \caption{Physical Parameters of LIMO robot }
    \begin{tabular}{l c c }
    \toprule
    \textbf{Name} \quad & \textbf{Symbol} & \textbf{Value} \\
    \midrule
      Robot mass (\USI{\kilo\gram}) & $m$ & 4.36 \\
        Sprocket wheel radius (\USI{\meter}) & $r$ & 0.055 \\
      Track width (\USI{\meter}) & $B$ & 0.172 \\
      Track length (\USI{\meter}) & $2\ell$ & 0.22 \\
    \bottomrule
  \end{tabular}
  \label{tab:limo_params}
\end{table}}

\subsection{User defined reference\DSrev{: simulation}}
\label{sec:user_def_reference}
The goal of this section is to demonstrate that, despite we rely on a simpler pseudo-kinematic model to synthesize the controller, 
this still captures the key terrain interactions represented by a more detailed distributed parameter model, leading to improved tracking performance.
We report in this section the simulation results comparing the performances of
the reference unicycle controller (UC)~\eqref{eq:control_input_unicycle}, with the proposed slippage-aware control
law (SLC)\eqref{eq:control_input_tracked}   for a given ``chicane'' velocity reference defined in~\eqref{eq:open_loop_vel_reference}.
The control law's outputs  ${}_bv_x$ and $\omega_z$ are mapped into desired speeds $\omega^d_{w,L}$,  $\omega^d_{w,R}$ by inverting~\eqref{eq:unicycle_mapping} and~\eqref{eq:tracked_vehicle_mapping} in case of UC and SLC, respectively,
to obtain the wheel speed inputs for the simulator. In the case of SLC controller
we also employ  the wheel speeds to estimate $\beta_L$, $\beta_R$, $\alpha$,  using~\eqref{eq:fbeta},~\eqref{eq:falpha} and $\dot{\alpha}^d$ by differentiation.
Note that the estimation of the slippage parameters from wheel speed can be done using desired ($\omega^d_{w,L}$, $\omega^d_{w,R}$) or actual values  ($\omega_{w,L}$, $\omega_{w,R}$), the latter bringing a more accurate prediction.

\begin{figure}[ht!]
   \centering
   \includegraphics[width=0.7\columnwidth]{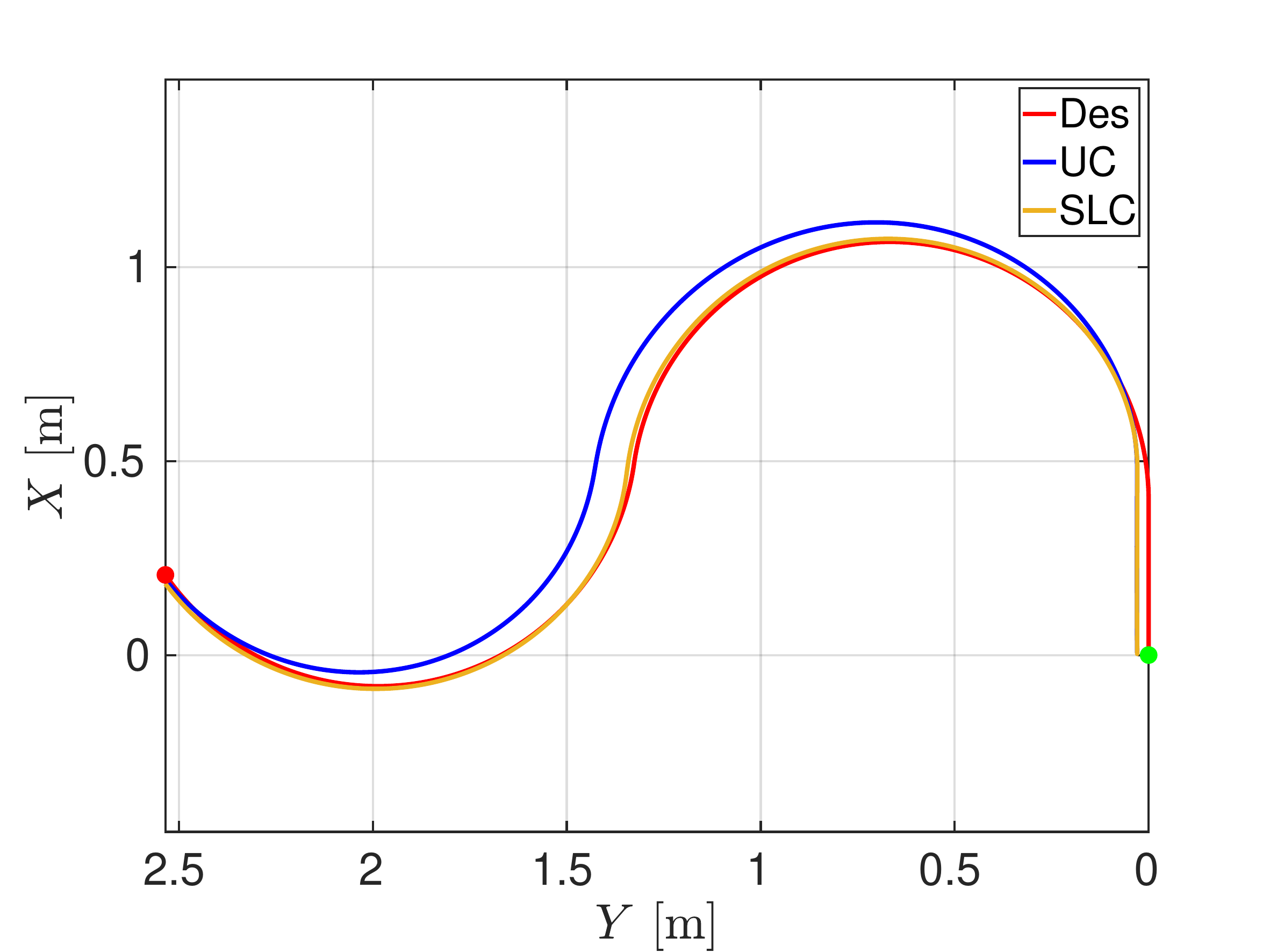}
   \caption{\textbf{Simulation}: Trajectory plot for the closed loop control of the chicane
    reference trajectory \eqref{eq:open_loop_vel_reference}(red line) with the MAXXII model \ref{eq:2D_distributed_model} for flat terrain. The friction coefficient is set to $\mu = 0.1$.  Both the UC Unicycle controller  (blue line)
    and the SLC slippage-aware controller (yellow line) are reported.  Green and red dots are the desired initial and final  configuration, respectively. }
  \label{fig:cloop_xy}
\end{figure}

\begin{figure}[ht!]
  \centering
  \includegraphics[width=0.75\columnwidth]{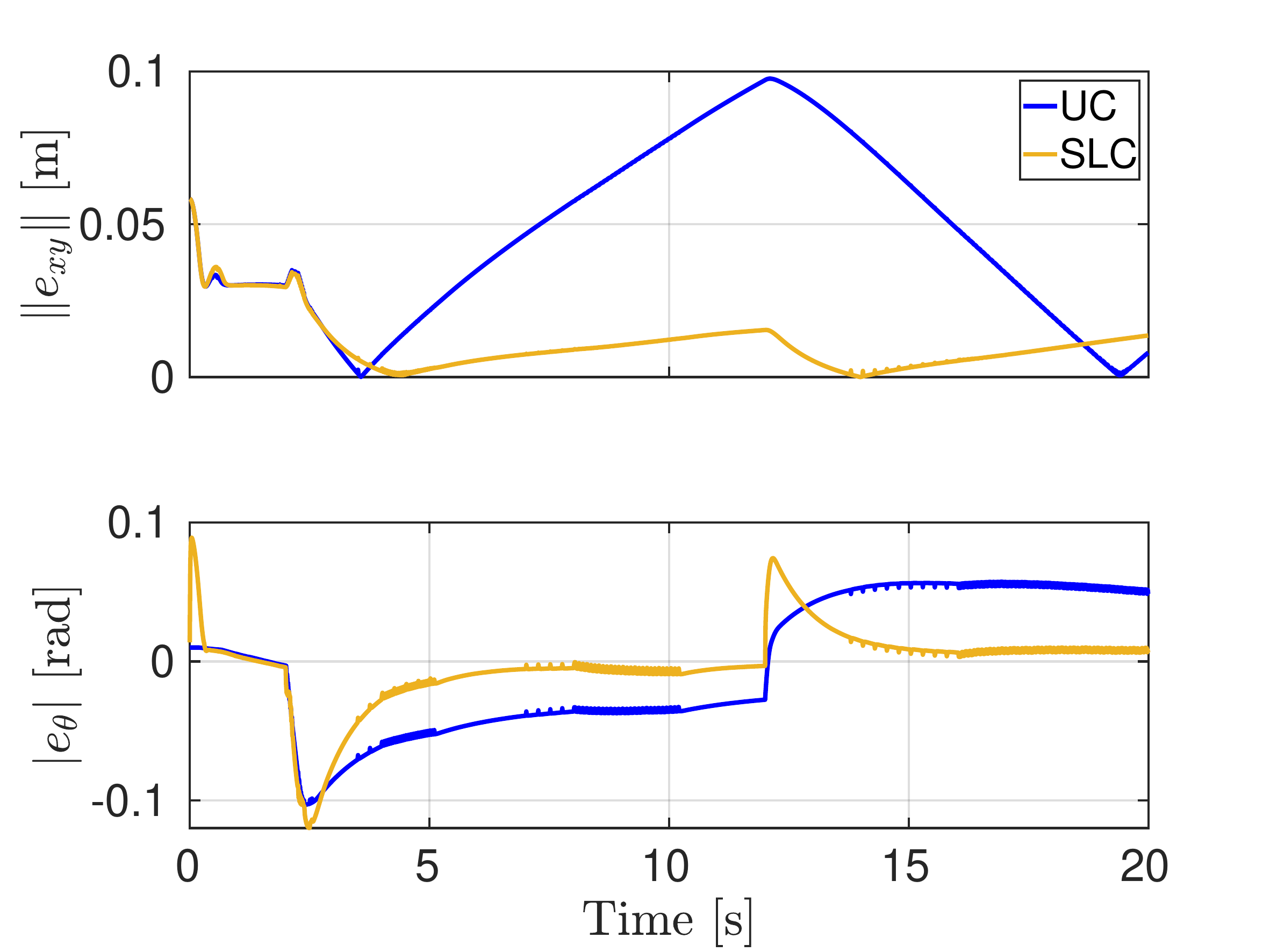}
  \caption{\textbf{Simulation}: Comparison of tracking errors for the closed loop control of the chicane
    reference trajectory \eqref{eq:open_loop_vel_reference} for the (blue)  UC  Unicycle controller and the (yellow) SLC slippage-aware controller: Cartesian error (upper plot) and orientation error  (bottom plot). }
  \label{fig:cloop_track_errors}
\end{figure}

The simulation is part of the Locosim framework~\cite{focchi2023locosim}\footnote{Source code is available
at the following  \href{https://github.com/mfocchi/tracked_robot_simulator.git}{(link)}.}. 
As terrain model, we assume a firm ground (\ie{} cohesion is null)
with a a very low friction coefficient $\mu = 0.1$ to emulate a challenging environment.
As reference trajectory we created a chicane with a deliberately
discontinuous velocity profile in ${}_bv_x^d$ and $\omega_z^d$. 
The robot is meant to accelerate forward in a linear motion then a
sharp left turn at $t=t_1$ is followed by a right turn at $t=t_2$.  The velocity reference is as
follows
%
%
%
{\small{
    \begin{equation}
       \begin{split}
      {}_bv_x^d =\begin{cases}
              v_{\max}^d (t - t_1 )  & 0 \leq t \leq t_1 ,\\
              v_{\max}^d & t_1 \leq t \leq t_{\mathrm{end}} ,
      \end{cases} ,
      \omega_z^d =
      \begin{cases}
              0 & 0 \leq t \leq t_1 , \\
              \omega_{\max}^d & t_1\leq t \leq t_2 , \\
              -\omega_{\max}^d & t_2 \leq t \leq t_{\mathrm{end}} ,\\
      \end{cases} \raisetag{2\normalbaselineskip}
      \end{split}
      \label{eq:open_loop_vel_reference}
    \end{equation}}}
\noindent with $t_1 = \SSI{2}{\second}$, $t_2=\SSI{12}{\second}$ and $t_{\mathrm{end}} = \SSI{20}{\second}$, $v_{\max}^d= \SSI{0.2}{\meter\per\second}$ and $\omega_{\max}^d = \SSI{0.3}{\radian\per\second}$.  The reference states are obtained by integration of~\eqref{eq:unicycle_desired} with \eqref{eq:open_loop_vel_reference} as inputs, 
while the initial robot state has been  set  to $[x_0, y_0, \varphi_0]^\top = [\SSI{0.05}{\meter}, \SSI{0.03}{\meter}, \SSI{0.01}{\radian}]^\top$, deliberately different from the initial reference $[0, 0, 0]^\top$. We also simulate the presence of actuation uncertainties by adding a Gaussian noise $n \sim \mathcal{N}(0,0.02)$ \USI{\radian\per\second} 
to the actuator inputs. We set gains $k_p=10$ and $k_\varphi=1$ for both controllers. Figure~\ref{fig:cloop_xy} 
reports the $X-Y$ plots for the simulation experiments showing the superiority of the SLC controller. Figure~\ref{fig:cloop_track_errors}, instead, shows that the UC controller has a maximum Cartesian error of $\SSI{10}{\centi\meter}$, while the SLC manages to keep it always below $\SSI{3}{\centi\meter}$. The tracking of the orientation is also significantly improved with the slippage-aware controller.

%
\begin{figure}[t]
  \centering
  \includegraphics[width=0.8\columnwidth]{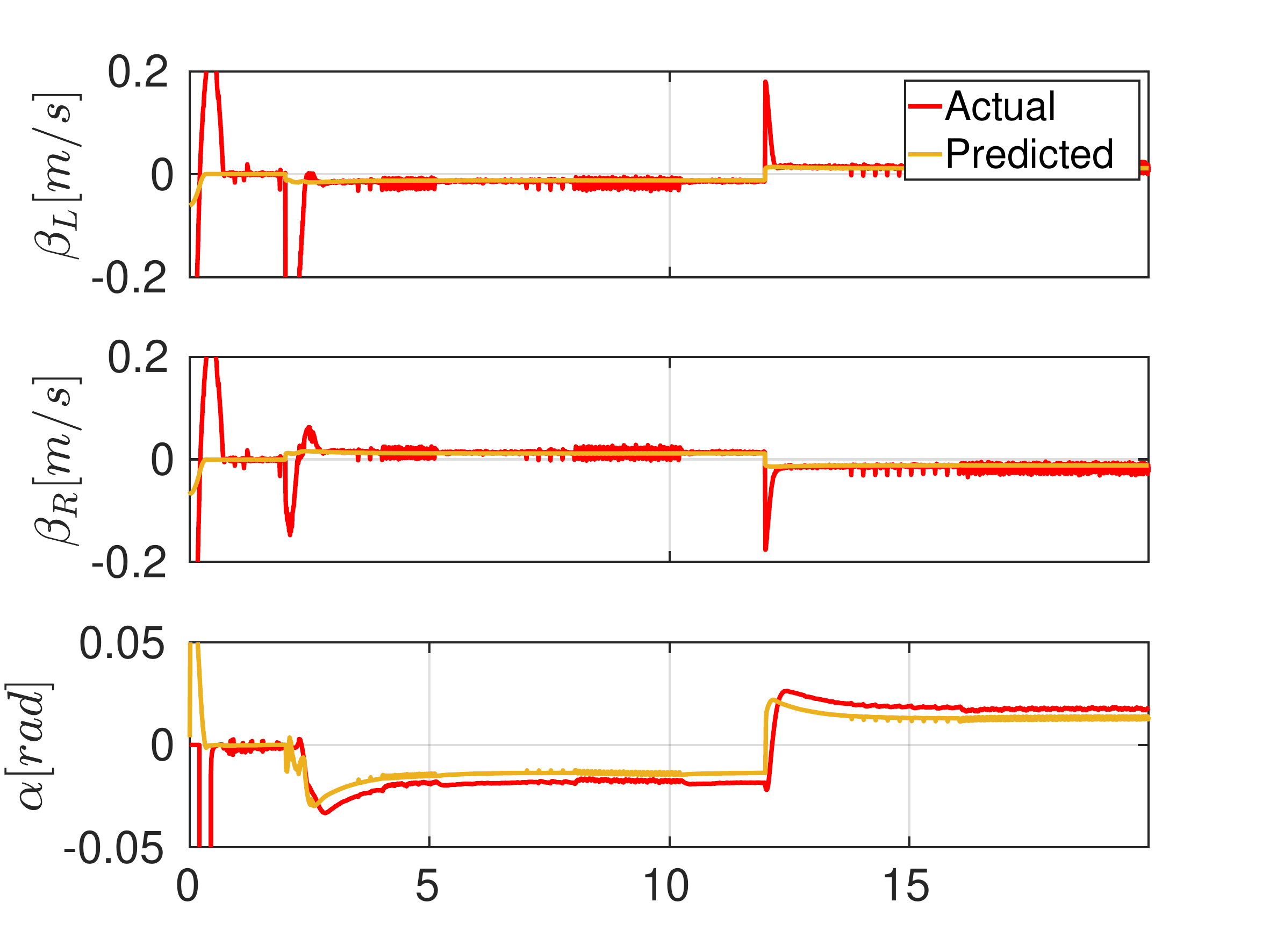}
  \caption{\textbf{Simulation}: Longitudinal slippage for (upper plot) left track (middle plot) right track
     and (bottom plot) lateral slippage. The red plots are the actual
    values while the yellow plot are the  values predicted with~\eqref{eq:falpha},\eqref{eq:fbeta}, used in
    the slippage-aware controller. }
  \label{fig:cloop_plot_slippage_pred}
\end{figure}
\begin{figure}[ht!]
   \centering
   \includegraphics[width=1.0\columnwidth]{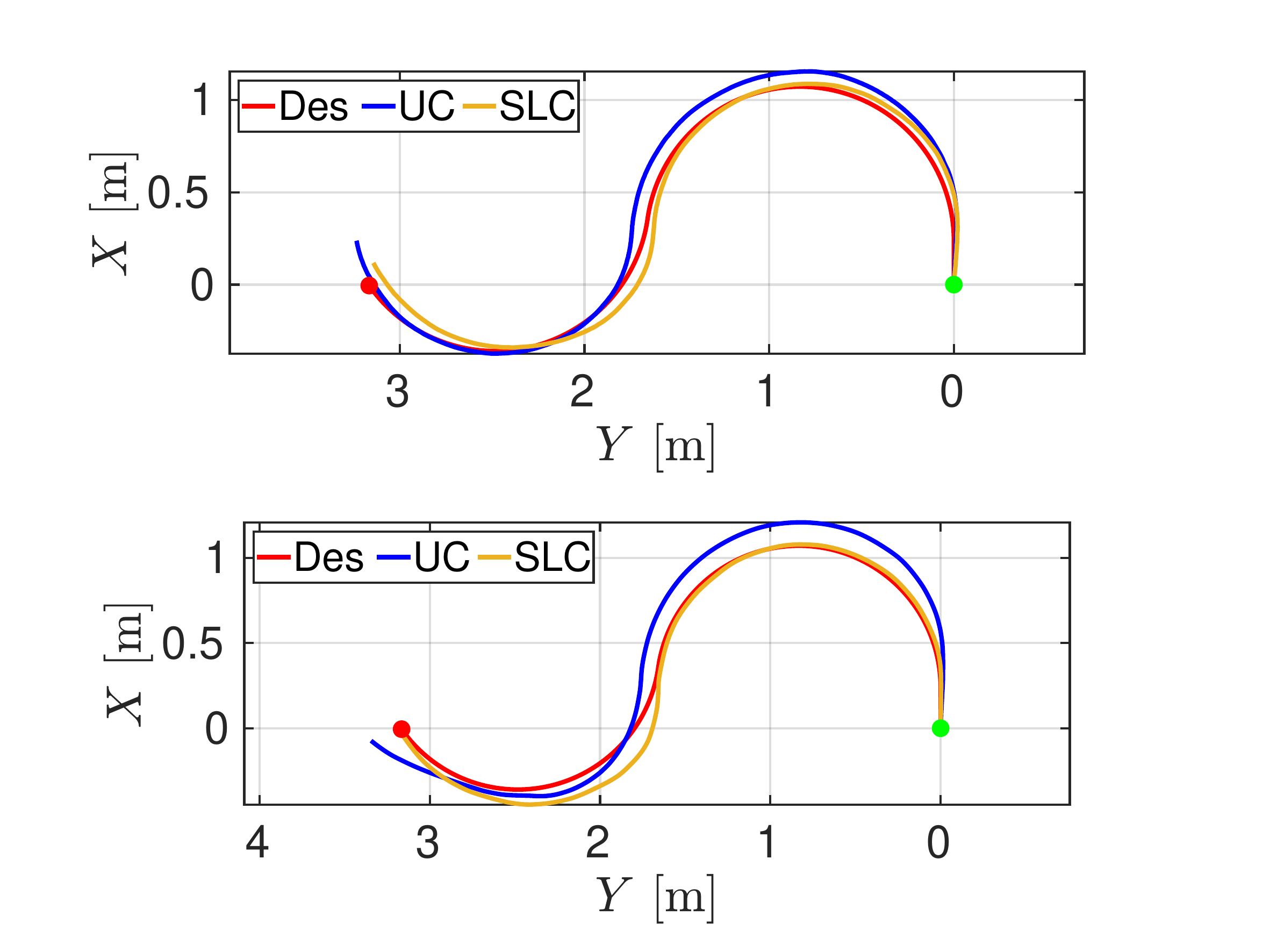}
   \caption{\DSrev{\textbf{Experiments}: Trajectory plot for the
       closed loop control of the chicane reference trajectory (red
       line) in~\eqref{eq:open_loop_vel_reference} with the LIMO robot
       in the experimental arena: (upper plot) parquet floor, (lower
       plot) slippery terrain.  Both the UC Unicycle controller (blue
       line) and the SLC slippage-aware controller (yellow line) are
       reported. Green and red dots are the desired initial and final
       configuration, respectively. The parameters $v_{\max}^d= 0.4$
       (slippery)/0.5 (parquet) m/s and
       $\omega_{\max}^d = \SSI{0.5}{\radian\per\second}$ were set to
       achieve sufficient slippage to appreciate the increased
       performance of the SLC controller. }}
  \label{fig:cloop_xy_exps}
\end{figure}

Figure~\ref{fig:cloop_plot_slippage_pred} shows the results of the
prediction of  $\beta_L$(upper plot), $\beta_R$(middle plot) and $\alpha$
(bottom plot) parameters.
The prediction for $\beta_L$ and $\beta_R$ is accurate good except when
there are velocity discontinuities, which reveal a more complex dependency on acceleration. 
This  behavior is reasonable, as longitudinal slippage depends on inertia, and abrupt changes in wheel speed (e.g., due to nonsmooth variations in $\omega_z$) can induce slippage. Our function approximator was trained using only velocity terms and therefore could not capture these effects.
The prediction of $\alpha$ is also correct 
(its sign is opposite to the steering speed $\omega_z$) just showing a slight
under-compensation.
\DSrev{
\subsection{User defined reference: experiments}
\label{sec:hw_exps}
We repeated the previous chicane experiment on the indoor arena with
the LIMO robot (Fig.~\ref{fig:robots}(right)) in two different terrain
condition: parquet and slippery terrain.  The slippery terrain is
achieved spreading soap on a plastic sheet attached to the floor.  The
parquet flooring in the indoor arena exhibited a friction coefficient
significantly higher than $0.1$. Consequently, we set higher target
velocities ($v_{\max}^d = \SSI{0.5}{\meter\per\second}$ and
$\omega_{\max}^d = \SSI{0.5}{\radian\per\second}$) than those used in
simulation to induce sufficient slippage for evaluating the
performance of our SLC controller. Conversely, on the soapy terrain,
due to its increased slipperiness, we reduced the longitudinal
velocity slightly to $v_{\max}^d =
\SSI{0.4}{\meter\per\second}$. 
Before performing the experiments, we run the identification of the
function approximators in Section \ref{sec:slippage_identification}
for both the parquet and the slippery terrain.  Due to lab space
limitations, we did not sample the wheel speed uniformly because this
would make the robot unpredictably drift away, but we set a variation
of $\vert \omega^d \vert\in [0, 2]$~rad/s with $0.3$ increments,
performing right and left spiral turns with angular velocity changing
for different (forward) longitudinal speed in the set
$v^d \in [0.23, 0.32, 0.4, 0.5, 0.6, 0.7]$~m/s. For all these
experiments we recorded the desired wheel speeds $w_L^d$, $w_R^d$ and
the estimated values of $\alpha$, $ \beta_L$,$\beta_R$. For the
experiment, a Motion Capture System was employed to estimate the pose
of the robot by means of markers attached to its base. To avoid
initial slippage of the tracks the desired longitudinal speed was
smoothly ramped up from $0$ to the desired value $v^d$.  To reduce the
effect of the noise, we passed the timeseries through a non-causal
filter before the training.  Due to the reduced number of samples, it
was necessary to up-sample the dataset with the \gls{rbf}.  In
Fig.~\ref{fig:cloop_xy_exps}, we report the $X-Y$ plots for the real
experiments.  The SLC controller tracks the reference trajectory more
accurately, whereas the UC controller overshoots the final reference
position on the parquet surface and accumulates a significant
orientation error on the slippery terrain. Throughout the trajectory,
the Cartesian tracking error is also larger in the UC case.  }

\begin{figure}[ht!]
  \centering
  \includegraphics[width=1.0\columnwidth]{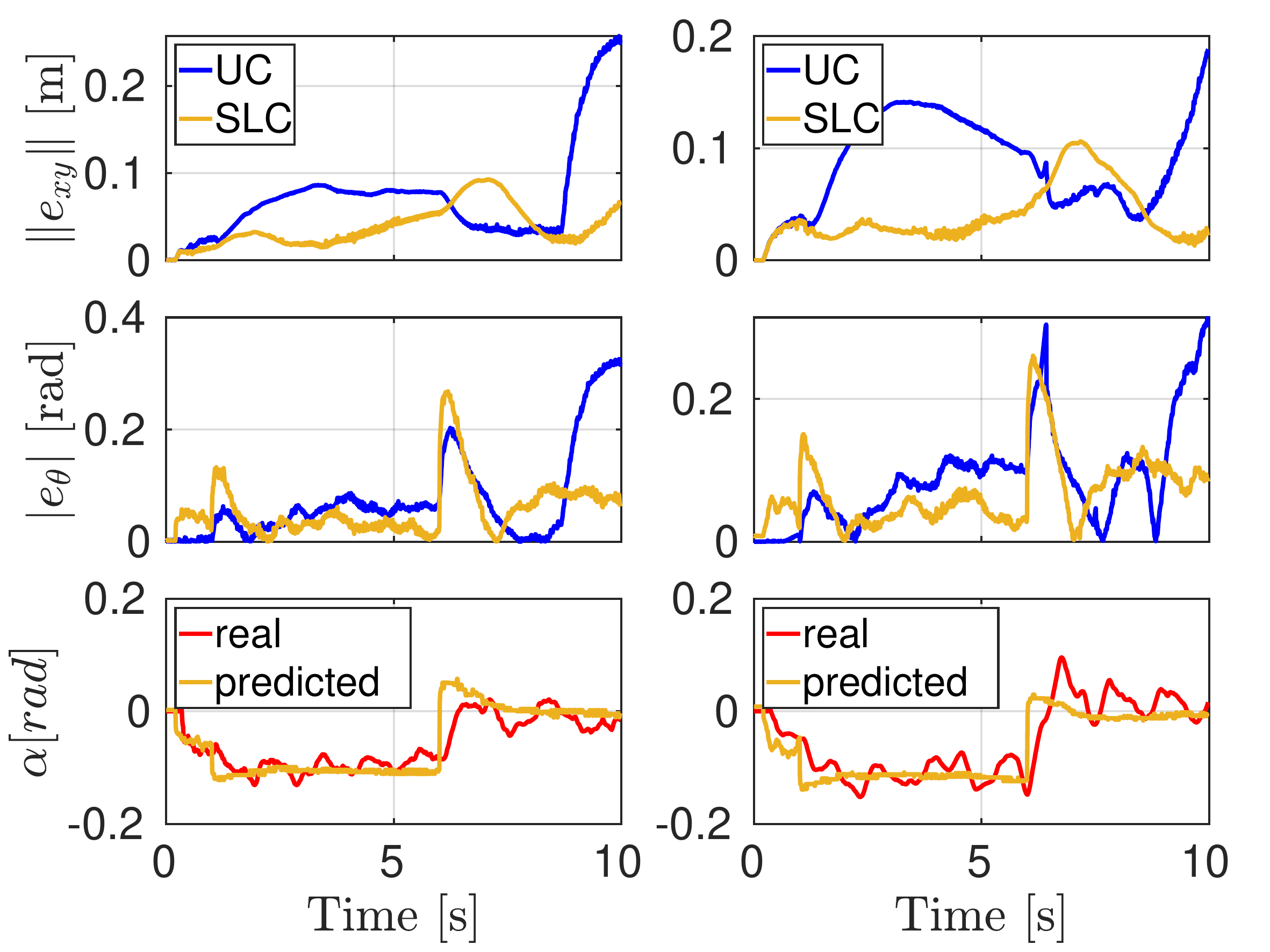}
  \caption{\DSrev{\textbf{Experiments:} Comparison of tracking errors
      for the closed loop control of the chicane reference
      trajectory~\eqref{eq:open_loop_vel_reference} with the LIMO
      robot in the experimental arena: (left plots) parquet floor,
      (right plot) slippery terrain.  The blue line is the UC Unicycle
      controller while the yellow line the SLC slippage-aware
      controller: upper plots show the Cartesian error while middle
      plot the orientation error. Finally, the bottom plots illustrate
      the estimated~\eqref{eq:alpha_estimate} and
      predicted~\eqref{eq:falpha} side slippage used in the SLC
      controller.}}
  \label{fig:cloop_track_errors_real}
\end{figure}

\DSrev{ Quantitatively Figure~\ref{fig:cloop_track_errors_real}(upper
  and middle plot), shows that the SLC controller achieves a 48$\%$
  decrease of the average Cartesian error (from
  $\SSI{7.2}{\centi\meter}$ to $\SSI{3.8}{\centi\meter}$), and a
  24$\%$ of the average orientation error (from 4.2 degs to 3.2 degs)
  in the parquet floor case, and a 55$\%$ and a 22$\%$ reduction for
  Cartesian and orientation error in the slippery terrain case,
  respectively, showing that the overall tracking performance is
  significantly improved with the slippage-aware controller.}
\DSrev{
The lower plot shows that the predicted $\alpha$ closely matches the value estimated from the Motion Capture System using \eqref{eq:alpha_estimate}.
It is worth noting that the amount of slippage is comparable on both the parquet and slippery terrains, as the higher slipperiness of the latter is compensated by the lower operating speed.
Experimental results are also reported in the accompanying video\footnote{The accompanying video can be found at this  \href{https://youtu.be/TyeZ0vRkI04}{(youtu.be/TyeZ0vRkI04)}}.
}
\subsection{Planning strategies evaluation}
\label{sec:plan_eval}
In this section we want to compare the performance of the  different planning strategies presented in Section~\ref{sec:planning}
in  relation with the standard UC and the proposed  SLC controller.
To appreciate the differences we set a higher desired longitudinal speed of \SSI{0.4}{\meter\per\second}. For this reason we need to increase also
the friction coefficient $\mu$ that is now set to a less slippery value of 0.4.

\begin{table}[ht!]
  \centering
  \caption{Planning Strategy Evaluation Experiments}
  {\small\begin{tabular}{ccc}
    \toprule
    \multirow{2}{*}{\textbf{Planner type}} & \multicolumn{2}{c}{\textbf{Controller}} \\ \cline{2-3}
    & Unicycle & Slip. aware \\
    \midrule
    Dubins                  & UC-DP  & SLC-DP \\ 
    Clothoids               & UC-CP  & SLC-CP \\ 
    Slip. aware (fix. speed) & UC-SLP & SLC-SLP \\ 
    Slip. aware (var. speed) & x      & SLC-SLP (var) \\
    \bottomrule
  \end{tabular}}
  \label{tab:experiments}
\end{table}

\begin{figure}[th!]
    \centering
    \includegraphics[width=0.49\columnwidth]{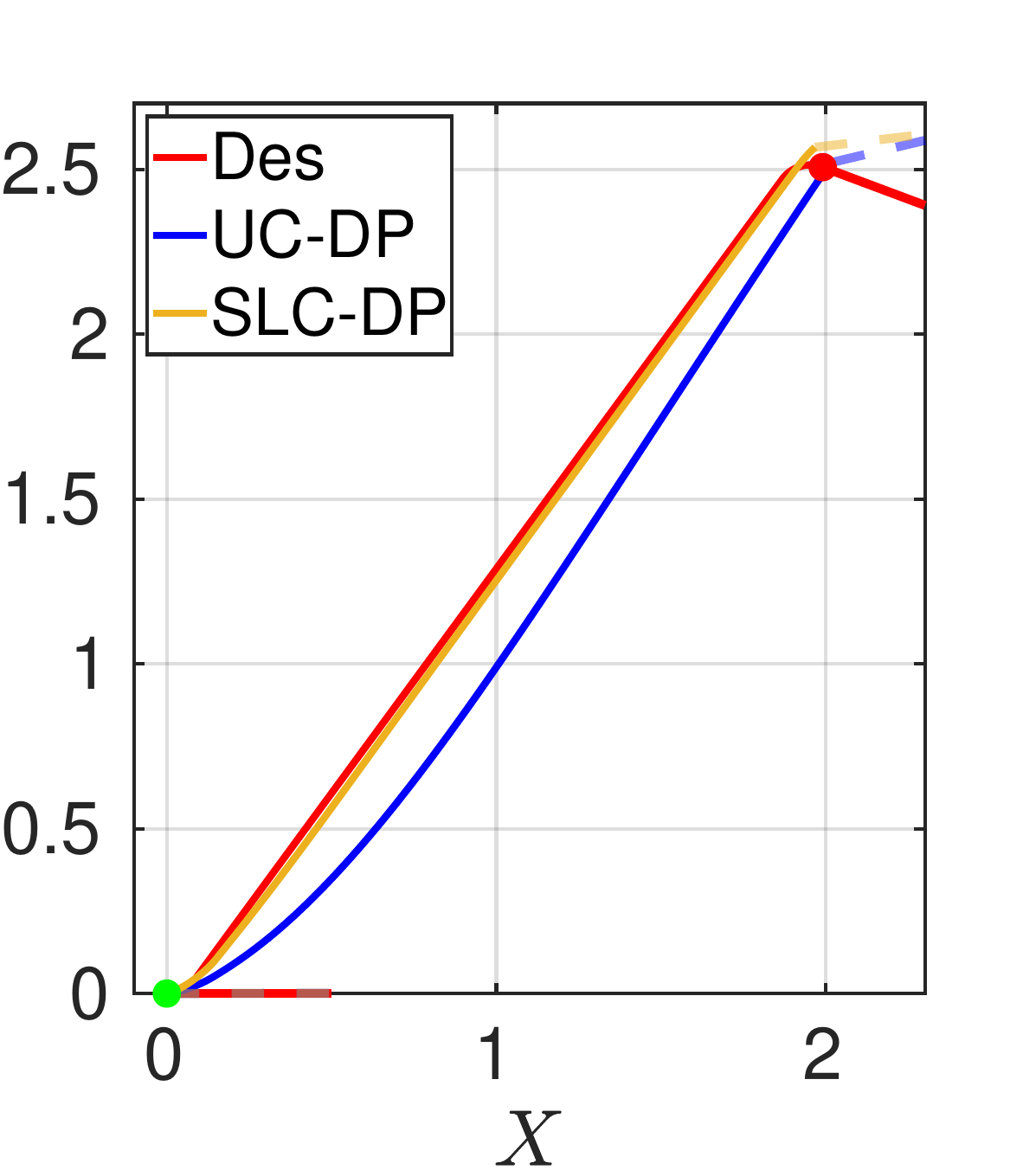}
    \includegraphics[width=0.49\columnwidth]{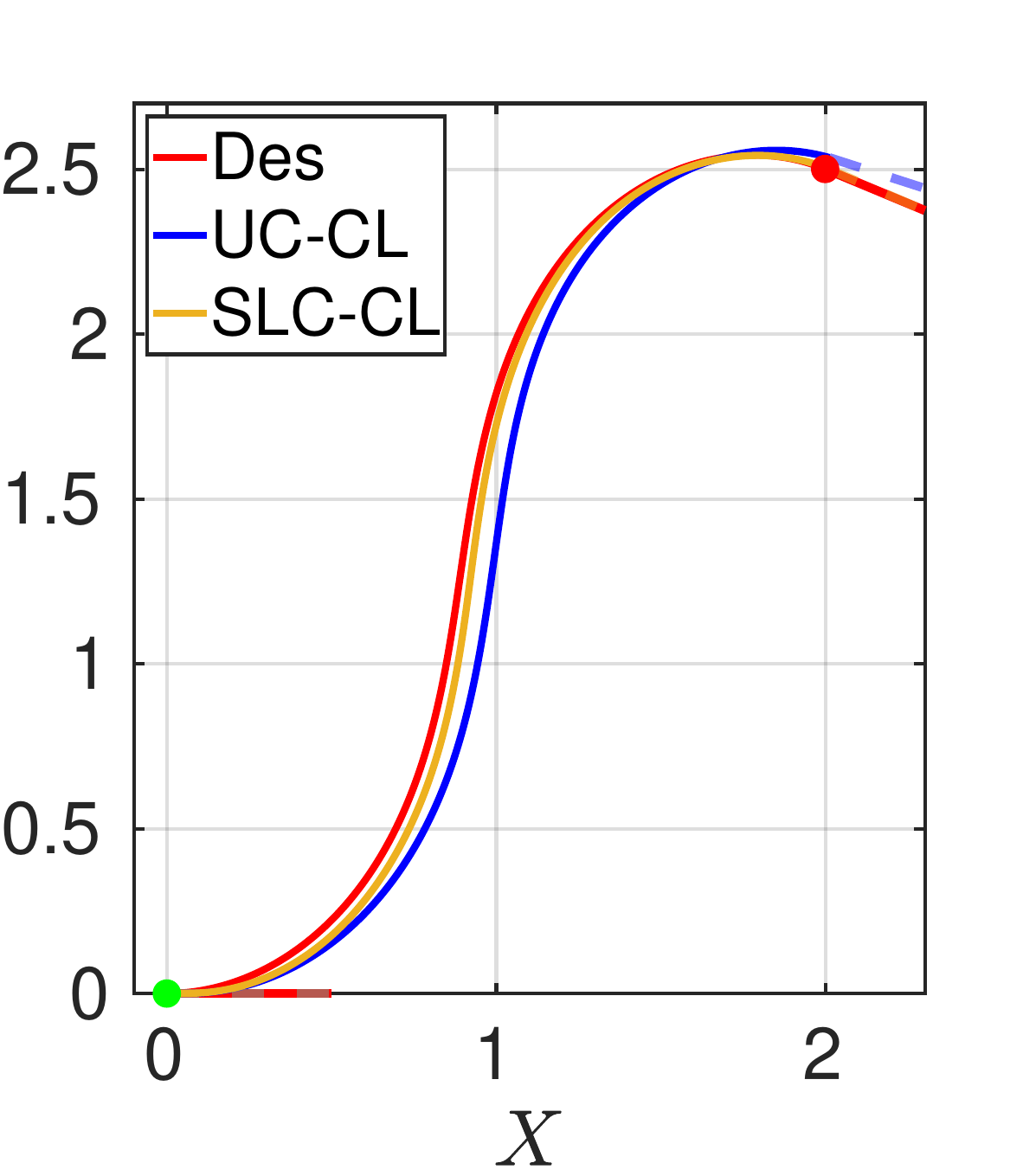}
    \includegraphics[width=0.49\columnwidth]{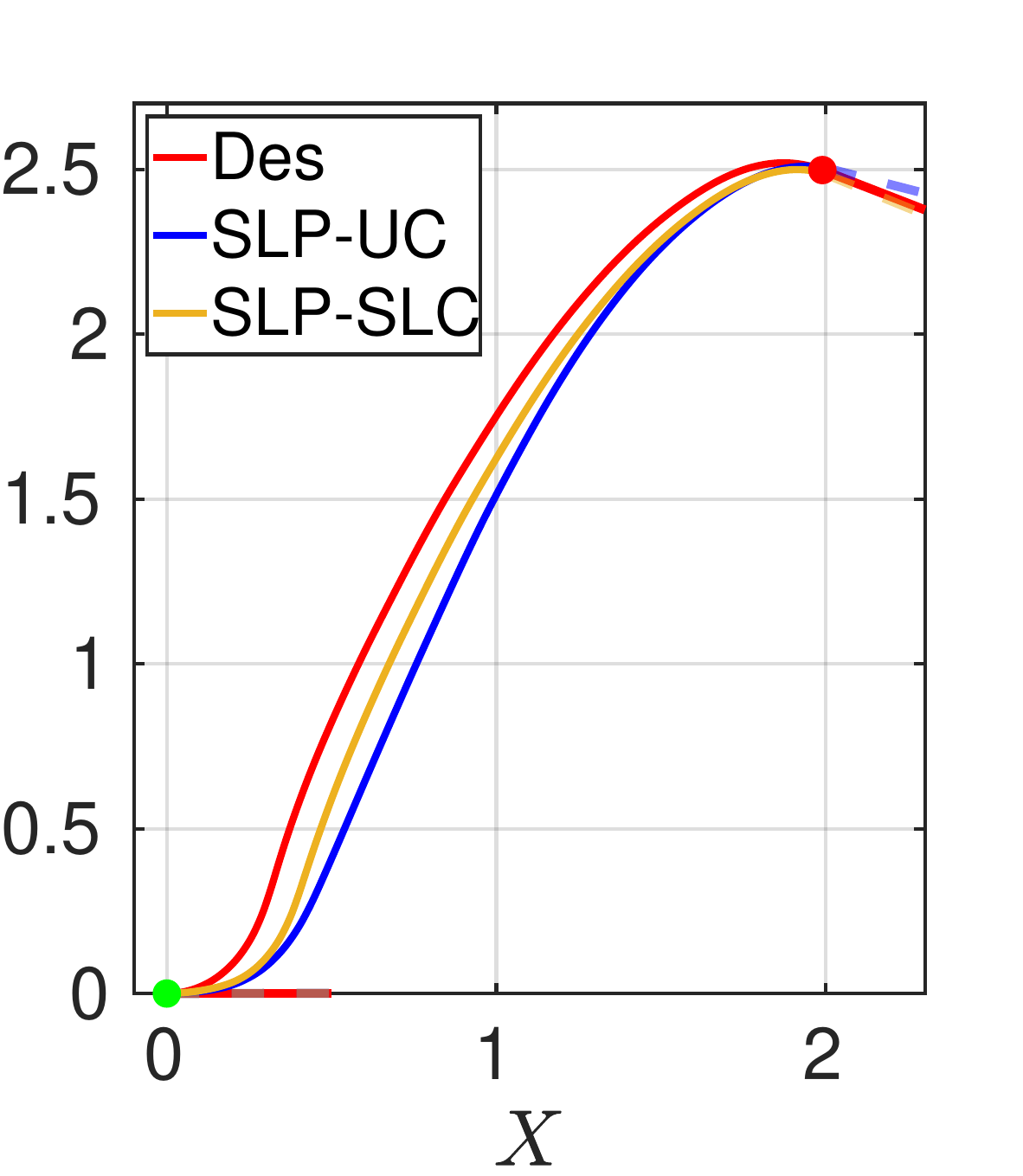}
    \includegraphics[width=0.49\columnwidth]{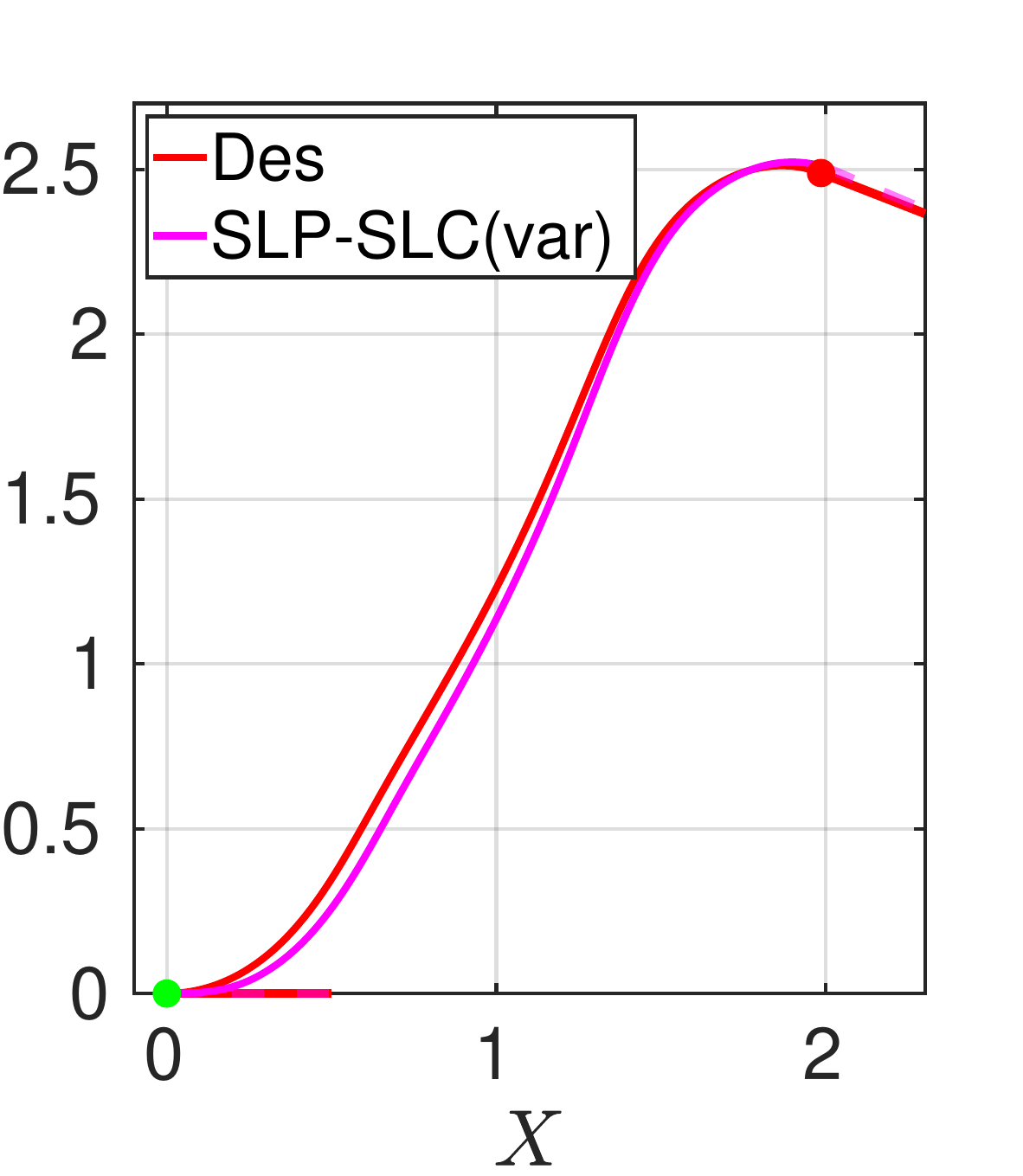}
    \caption{(upper left) Bird eye view of the trajectory planned  with Dubins tracked with the  unicycle controller UC-DP (blue) and the slippage-aware controller SLC-DP 
    (yellow).   (Upper right) Bird eye view of the trajectory planned  with Clothoids tracked with the  unicycle controller UC-CP (blue) and the slippage-aware controller SLC-
    CP (yellow).        (bottom left) Bird eye view of the trajectory planned  with the Slippage-aware Optimal planner tracked with the  unicycle controller UC-SLP (blue) and the  slippage-aware controller SLC-SLP (yellow).  (Bottom right)  the Slippage-aware Optimal planner where the fixed speed constraint is relaxed,  tracked    with the slippage-aware controller SLC-SLP (var)  (magenta).    
            Green and red dots are the desired initial and final  configuration, respectively. The desired initial and final orientation is highlighted with a solid line, the actual with dashed translucid line. }
    \label{fig:bird_eye_traj}
\end{figure}

\begin{figure}[th!]
  \centering
  \includegraphics[width=0.9\columnwidth]{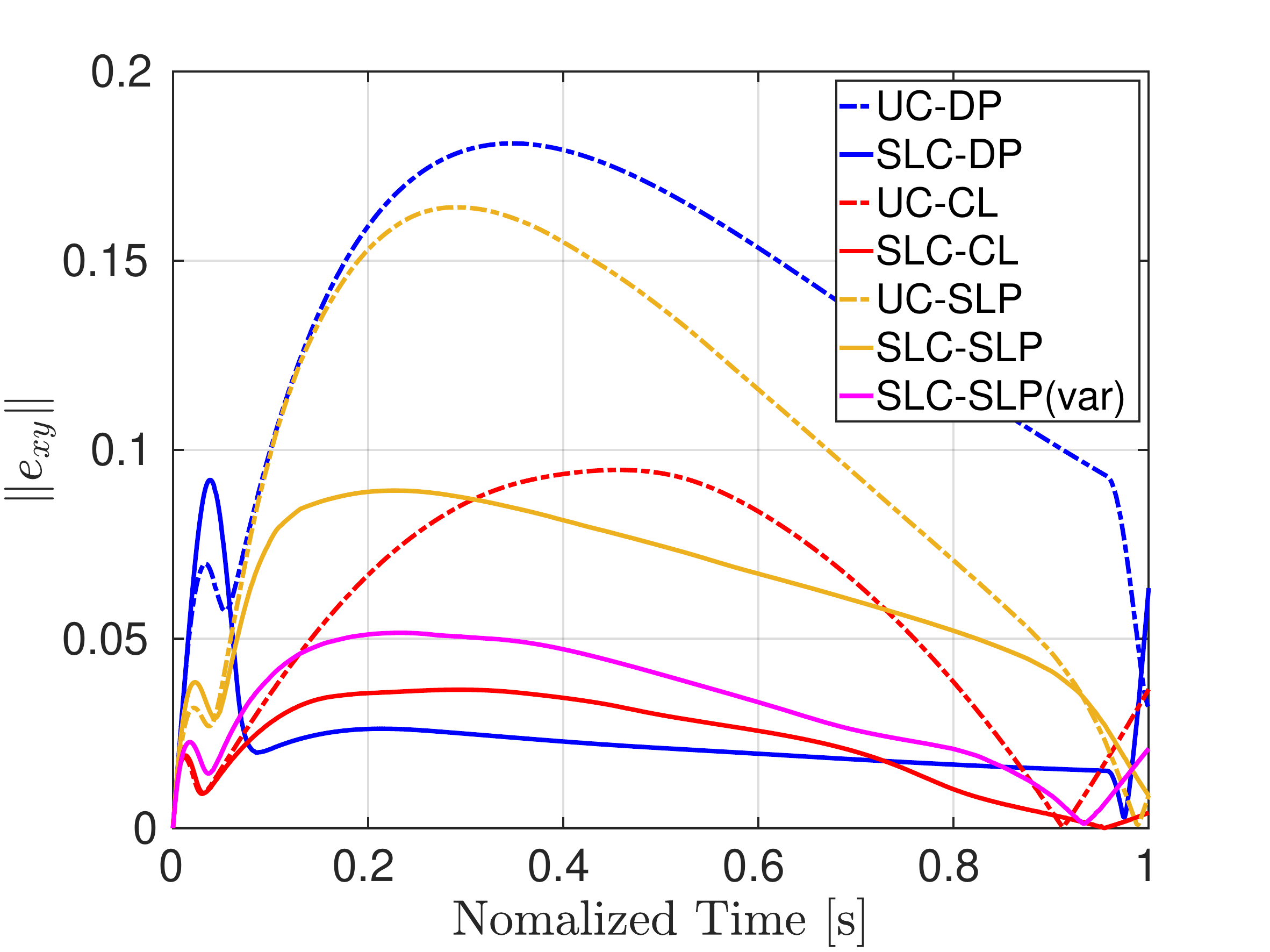}
  \caption{Tracking (Cartesian) error for the different experiments. Time is normalized for a fair comparison.}
  \label{fig:planning_tracking_errors}
\end{figure}

We  perform the tests summarized in Table~\ref{tab:experiments}. 
In the case of the optimal control SLP and SLP (var) planners, the optimisation has been implemented in Matlab using the \texttt{fmincon} function.
To improve performance, we used a C++ implementation of the \gls{ocp} while the communication with the Locosim framework, where the simulation models are implemented,  is through ROS.
We set a starting configuration $\vect{c}_0= [0, 0, 0]^\top$
and a target configuration $\vect{c}_{f}= [2.0,  2.5, -0.4]^\top$. The speed is constant by definition in the case of the Dubins. In the case of the  Clothoid planner and the slippage aware optimal planner (UC-SLP, SLC-SLP),
we set artificial constraints  $v_{\max}^d=v_{\min}^d = \SSI{0.4}{\meter\per\second}$ to have a fair comparison.
Then we show the impact of relaxing this constraints in the case of SLC-SLP(var) where we set $v_{\min}^d = 0$ and $v_{\max}^d= \SSI{0.4}{\meter\per\second}$
letting the optimizer free to vary the longitudinal speed along the trajectory, while still  encouraging  a forward motion.
A bird-eye (XY) view of the results for (left) Dubins and (right) Clothoids  are shown in Fig.~\ref{fig:bird_eye_traj} (upper plots)
while the results for the optimal planner in case of (left) fixed and (right) variable longitudinal velocity are shown in Fig.~\ref{fig:bird_eye_traj}(bottom plots).
The Cartesian tracking errors are shown in Fig.~\ref{fig:planning_tracking_errors}. In this figure time has been normalized
for a meaningful comparison because the trajectories can have different durations.

Fig.~\ref{fig:bird_eye_traj}~(upper left) shows that the Dubins planner does a very abrupt curve to reach the final orientation 
at the end of the trajectory. Even if inside the actuation limits that change in orientation cannot be tracked, because of slippage, 
by none of the controllers, with the robot ending with a big orientation error. 
However, the SLC controller does a better job managing to track better the trajectory along the way. 
Inspecting Fig.~\ref{fig:planning_tracking_errors} we can conclude that,  
in all the cases, a smaller tracking error along the trajectory, is achieved by the use of the  slippage-aware controllers (solid lines).
The worst results are obtained by using Dubins with the unicycle controller.
Intermediate accuracy is with the slippage-aware controller paired with Dubins (SLC-DP) or by the slippage-aware planner paired with the unicycle controller (UC-SLP).
This shows that taking into account slippage either in control of planning improve performances.
The best result, in terms of reducing the final error are for the combination of slippage-aware planner/controller (SLC-SLP),
with even lower errors if the longitudinal velocity is left as a free variable (SLC-SLPvar) because the robot can smoothly accelerate
and slow down at the beginning and the end, resulting in lower values of the longitudinal slippage.
The Clothoid planner behaves also very well, achieving the best results in combination
with the slippage-aware controller,  comparable with the SLC-SLPvar. The trajectory durations for the Dubins and the SLC-SLPvar planner are 8.12 s and 8.4 s, respectively, while for the Clothoid planner is slightly bigger (9.07 s).

\subsubsection{Statistical analysis}
To demonstrates that these trends are valid not only for a specific target configuration,
we repeated the tests for several configurations sampled in a annular region whose radius ranges from 2 to 4 m around the robot. 
We generated the sampling using the Halton low-discrepancy deterministic sampling technique, which ensures good coverage even with small number of samples
\footnote{In contrast to non  deterministic methods (\ie{} random sampling methods),  such as Monte Carlo sampling with a uniform distribution that tend to produce clustered points,the Halton technique distributes samples uniformly, ensuring that the
points are sufficiently close to one another without leaving any region under-sampled.}. 
Since the mapping is nonlinear, we avoid sampling in polar coordinates and instead use the Halton sampling in $[0,1]\times3$.
The first two columns represent the X,Y coordinates of the target points, which can be scaled to cover a square of
edge $R=\SSI{4}{\meter}$. Points outside the inscribed annular region are then removed.
The third column is directly mapped to  $[0, 2\pi]$ and represents the target orientation $\varphi^d_f$.
%

We generate 100 randomized target configurations following the procedure described above. 
For each target $\vect{c}_f^d$ in the set, we run a simulation starting from $\vect{c}_0^d=(0,0,0)$ and compute the integral 
of the tracking error along the trajectory. We report the mean and standard deviation of the computed integrals 
across all samples, repeating the procedure for each planner/controller combination, as summarized in Table~\ref{tab:statistical_analysis}.
The results presented in this table serve to evaluate how effectively the robot can track a trajectory for a given controller and planner configuration.
The reader should be aware that the planning durations can be different, therefore we also report the average plan duration considering the Dubins as the nominal one $T_{f,{nom}}$.
The Clothoids planner has, on average, an 83\% longer planning duration, while the Optimal planner takes 31\% longer compared to the nominal one.
\begin{table}[ht!]
  \centering
  \caption{ Statistical analysis }
    {\small
  \begin{tabular}{l c c c c }
    \toprule
    \textbf{Exp}     & $\Vert e_{xy}\Vert$  & $\Vert e_{\varphi}\Vert$    & $T_{f,{nom}}$            &      Non conv. \\
    \midrule
    UC-DP            & $0.20  \pm 0.096$   & $0.24  \pm 0.089$          &        1                & NA   \\   
    SLC-DP           & $0.089 \pm 0.062$    & $0.40  \pm 0.12$            &        1                 & NA        \\ 
    UC-CP         & $0.090 \pm 0.054$   & $0.056 \pm 0.044$        &        1.83               & NA   \\ 
    SLC-CP       & $\mathbf{0.03  \pm 0.022}$   & $\mathbf{0.038\pm 0.046}$&   1.83                &NA  \\ 
    UC-SLC       & $0.11  \pm 0.046$   & $0.18  \pm 0.062$            &        1.31              & 22  \\ 
    SLP-SLC       & $0.055 \pm 0.04$   & $0.14  \pm 0.043$            &        1.31               & 22   \\ 
    \bottomrule
  \end{tabular}}
  \label{tab:statistical_analysis}
\end{table}
In the case of optimal control planner, the convergence failure is also reported, where the number of iterations is limited to 500.
Inspection of the table confirms the outcome shown in Fig.~\ref{fig:planning_tracking_errors}, revealing that
the worst results occur with the unicycle controller paired with the Dubins planner (UC-DP), where slippage effects are neglected at both the control and planning level.
The tracking performance of the unicycle controller significantly improves when paired with the slippage-aware planner (UC-SLP).
In all cases, using the slippage-aware controller enhances performance compared to the standard unicycle controller.
The optimal control approach failed to converge in 22 out of 100 experiments (22\%).
The analysis confirms that the best results are achieved when the slippage-aware controller is paired with either the slippage-aware planner
(SLC-SLP) or the clothoid planner (SLC-CP). This suggests that the Clothoid planner offers comparable performance without incurring in  high computational cost of optimal control  making it a preferable alternative and a promising approach
for enhancing sample-based navigation algorithms, which typically rely on Dubins maneuvers.
On the other hand, with respect to planning with Clothoids, optimal control  offers distinct advantages:  (1) brings lower trajectory duration, (2) produces references consistent with the tracked vehicle considering slippage effects, by  incorporating a more physically accurate slippage-aware model than just an unicycle. (3) it enables to handle variable speed profiles, (4)  It allows enforcing actuator constraints directly on wheel speeds. Point (3),(4) ensure that the result is a physically feasible trajectory.
On the downside, optimal control methods may suffer  from (possible) lack of convergence, the  solution is dependent on the initialization and  can remain 
trapped in suboptimal local minima (local optimality),  the presence of integration errors and a higher computational cost. 
The issue of local optimality can be partially alleviated by warm-starting the solver with the simpler, possibly infeasible trajectory generated using Dubins or Clothoid methods.

\subsubsection{Different friction coefficients}
\begin{figure}[ht!]
  \centering
  \includegraphics[width=1.0\columnwidth]{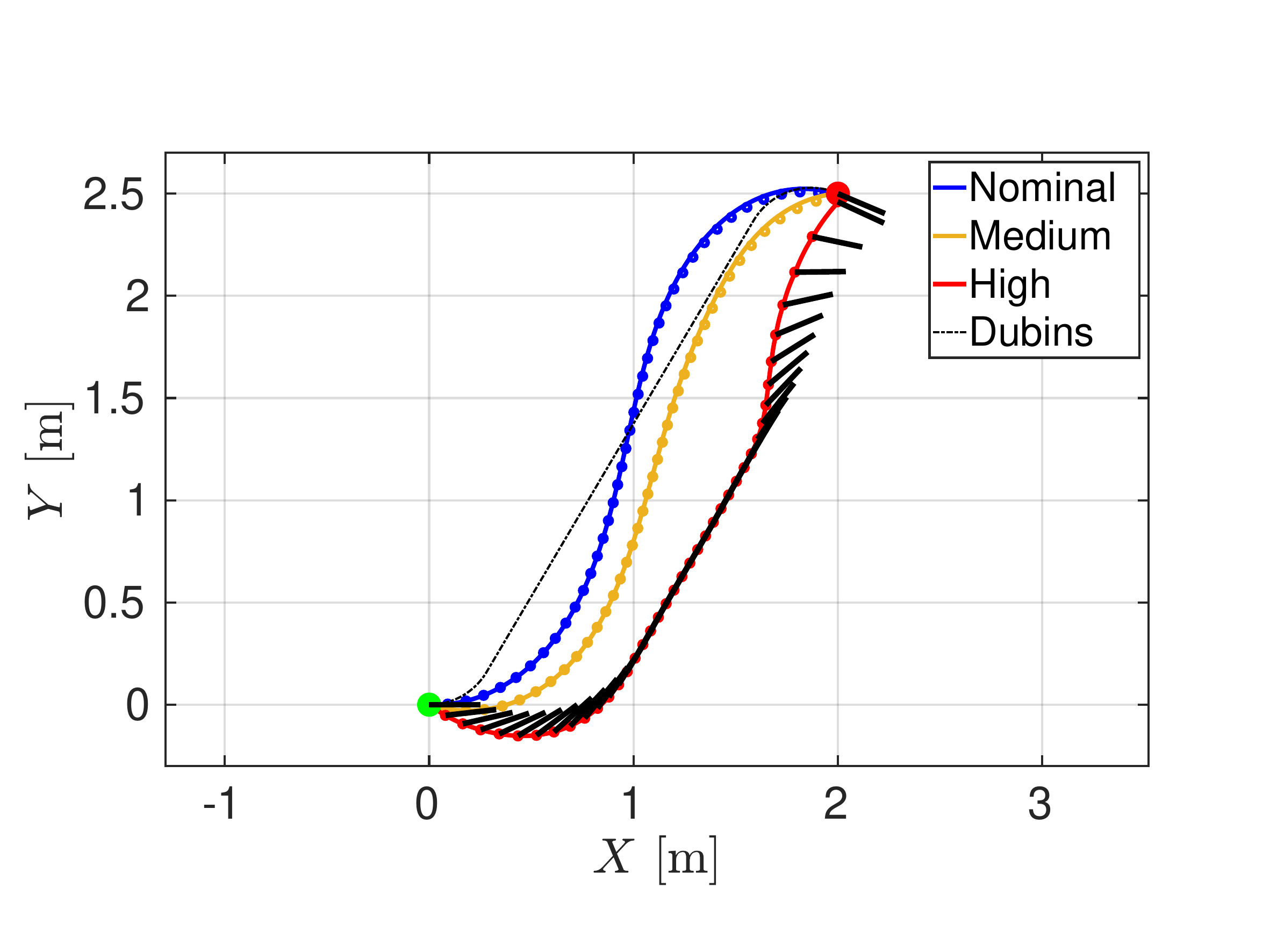}
  \caption{Planning with optimal control for terrain of increasing slipperiness: nominal (blue), medium (yellow) and high (red) slippage. In dashed black the Dubins's curve is reported for reference. For  comparison the time is normalized in [0,1] for the 3 tests. In the high slippage case the desired  orientation is also highlighted with black arrows along the trajectory.  }
  \label{fig:friction_tests_traj}
\end{figure}

Figure~\ref{fig:friction_tests_traj} illustrates how different values of the slippage coefficient—nominal (same as in the previous test), medium, and high—affect the reference trajectory generated by the optimal planner. The function approximators used in the pseudo-kinematic model has to be trained for the different levels of slippage.
To emphasize the effects of slippage, the upper bound on longitudinal speed $v_{\max}^d$ has been increased from 0.4 to \SSI{0.8}{\meter\per\second}, while the lower bound remains at \SSI{0}{\meter\per\second} to encourage forward motion. The Dubins curve is also included in the figure for comparison. The planner results in the robot turning with lower curvatures under higher slippage conditions because it requires lower centripetal force causing lower slippages.
The optimization results yield trajectories of different durations: 5.2s, 4.9s, and 4.12s
for normal, medium, and high slippage, respectively. Interestingly, the shortest duration
corresponds to the plan that employs the high-slippage model. This occurs because the
optimizer chooses to maximize speed while leveraging slippage for turning, effectively reducing the total trajectory time.
%
\begin{figure}[ht!]
  \centering
  \includegraphics[width=1.0\columnwidth]{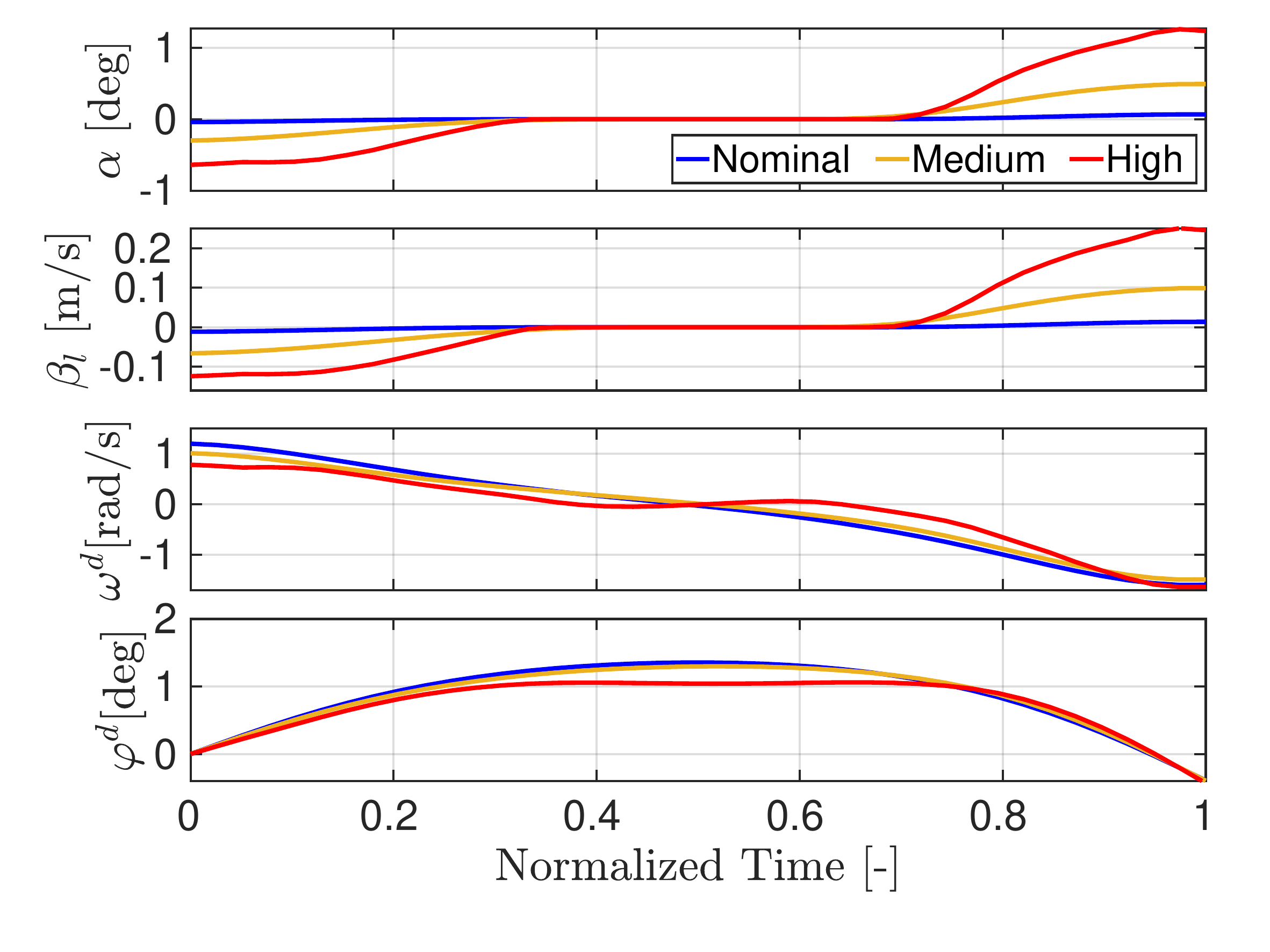}
  \caption{Plot of the lateral $\alpha$  (first plot) and longitudinal  $\beta$   (second plot) slippage parameters for  nominal (blue), mid (yellow,) and high (red) slippage. 
  Plot of the planned angular velocity $\omega_z^d$ (third plot) and robot orientation $\varphi^d$ (fourth plot).}
  \label{fig:friction_tests_slippage_params}
\end{figure}

Finally, the estimates of the slippage parameters, that has been used in the slippage-aware controller are shown in Fig.~\ref{fig:friction_tests_slippage_params}. The upper plot is the lateral slip $\alpha$ whose sign is opposite with with the sign of $\omega_z^d$(third plot). The second plot is the $\beta_l$ (having $\beta_r$ an analogue trend but mirrored).
In coherence with the  expectations, the slippages are higher for more slippery terrain. The fourth plot shows that the target orientation is achieved in all  cases. 
%
\subsubsection{Exponential function approximation}
The average computation time to solve the optimization problem using the optimal control planner is 18 seconds. The bottleneck is due to the multiple evaluation of the regressors. 
When less accurate results are acceptable, an exponential approximation that considers only the functional
dependency on the turning radius $R = {}_bv_x/\omega_z$, can be used in place of the decision trees in~\eqref{eq:fbeta}
and~\eqref{eq:falpha} to estimate the slippage parameters, significantly reducing the optimization time.
We propose the following exponential parametrization
\begin{equation}
  \alpha, \beta_L, \beta_R = f_{\alpha, \beta_L, \beta_R}(R) = \begin{cases}
    -c_1 e^{-c_2 R} & R \geq 0 \\
    c_1 e^{c_2 R} & \text{otherwise}
  \end{cases} \text{,}
  \label{eq:exp}
\end{equation}
where $c_1 >0$ and $c_2 > 0$ are two constants that embed the terrain
specificity and that are identified based on the type of track and
terrain.  In this case, we employ the exponential function~\eqref{eq:exp}
to fit the data for $\alpha$, $\beta_L$, and $\beta_R$, using the curvature radius as input.
With this exponential approximation, the computation time is reduced to 1.64 seconds.
Table~\ref{tab:comp_time} presents the average computation times for various planning strategies,
demonstrating that both the Dubins and Clothoid planners, which feature closed-form solutions,
have computation times that are orders of magnitude lower.
\begin{table}[ht!]
  \centering
  \caption{Planner Computation Time Comparison}
  {\small\begin{tabular}{l c  } \toprule  \textbf{Planner} &  \textbf{Avg. comp. time (\USI{\second})} \\ \midrule
        Dubins Planner (DP)       & 0.06  \\
        Clothoids Planner  (CP)     & 0.022  \\
    SLP: Exp. approx                         &  1.64     \\
    SLP: Dec. Trees, Side Slip. est.             &  7.2     \\
    SLP: Dec. Trees, Side+Long Slip. est.       & 18  \\
    \bottomrule
  \end{tabular}}
  \label{tab:comp_time}
\end{table}
\section{Experiments on slopes}
\label{sec:slope_results}
\begin{figure*}[ht!]
  \centering
  \includegraphics[width = 0.65\textwidth]{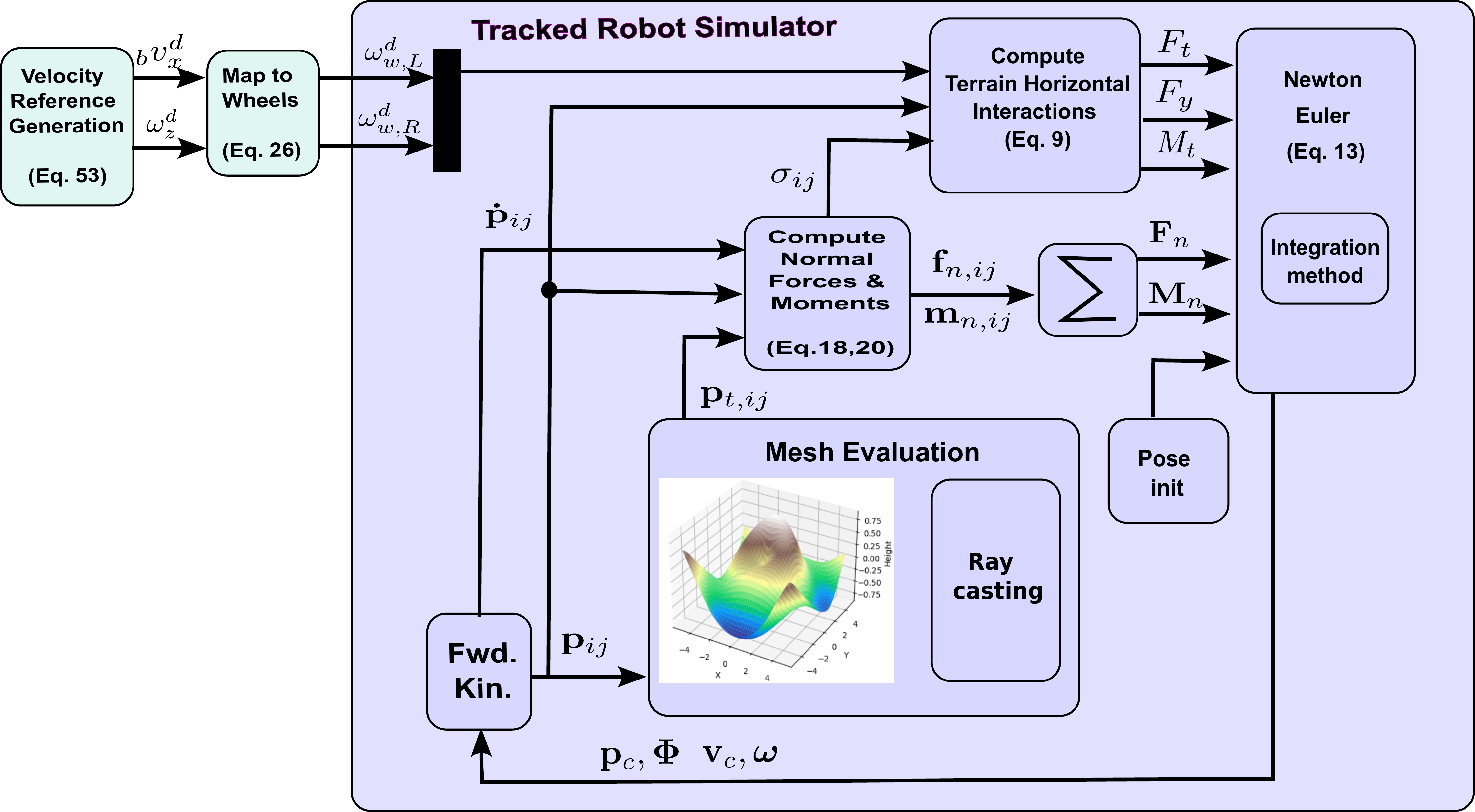}
  \includegraphics[width = 0.3\textwidth]{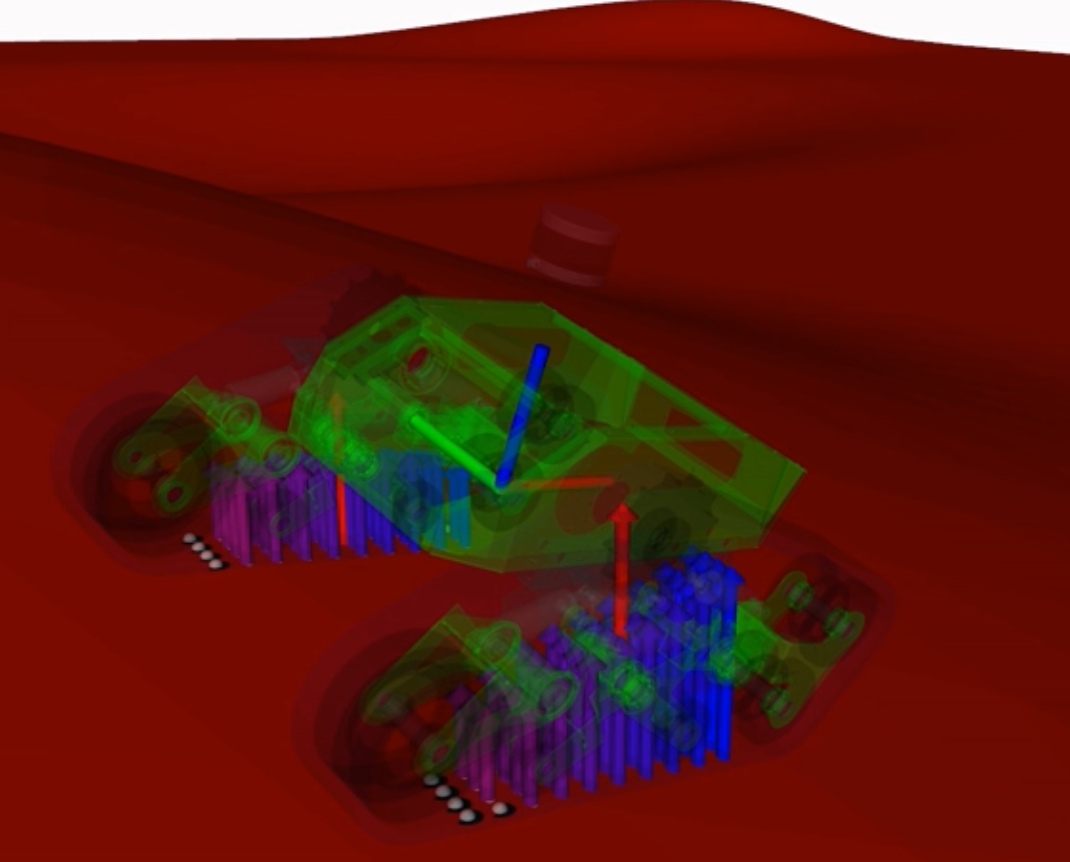}
  \caption{(left) Block diagram of the distributed parameter dynamic simulator, highlighting variables and computational steps. (right) Snapshot of a down-slope simulation with dynamic model with nonuniform loading. The blue arrow represents the normal terrain forces $\vect{f}_{n,ij}$, color shifting towards red means softening of the terrain due to speed. Because of the downward slope, the loading is shifted toward the front of the tracks. The resultant forces for each track are also shown with red arrows. The base frame is also highlighted as solid lines, while the robot chassis is translucid.}
  \label{fig:block_simulator}
\end{figure*}

For experiments on slopes we employ, for the simulation, the
distributed parameter model \eqref{eq:3D_distributed_model} whose parameters
are reported in Table~\ref{tab:sim_params}. A schematic open loop diagram of the  simulator, 
highlighting the computational steps, is presented in Fig. \ref{fig:block_simulator}~(left).
The sloped terrain is discretized into a mesh with triangular
shapes using the Open3d~\cite{open3d} library and  a ray casting algorithm is employed \cite{ray_casting_open3d}
to query the elevation of the terrain  for a certain (x,y) tuple representing the robot position.
This is obtained  finding the intersection between a vertical ray passing
through (x,y, $b$) and the mesh, where $b=-10$ m is the baseline distance from which the rays are casted.
To initialize properly the simulation it is fundamental to initialize the robot state to a pose that is consistent with the terrain inclination at 
the starting position (i.e a pose in which the robot does not penetrate the terrain).
This means  evaluating also of the terrain normal at the starting position to get the initial values for the roll, pitch, yaw angles.
At each simulation loop, first the normal force  distribution is computed via~\eqref{eq:terrain_forces} querying to the \textit{mesh evaluation} routine, the  projection of the 
track patches on the terrain (for the \textit{actual} robot state). Then then normal pressure $
\sigma_{ij}$  on the track patches is computed via~\eqref{eq:sigma_distributed}. Subsequently, the shear displacements are computed from 
the input wheel speeds via~\eqref{eq:shear_displacement} and finally the horizontal terrain interactions 
with~\eqref{eq:terra_mechanics_interactions2}. All forces are then used to compute linear and  angular accelerations by 
inverting~\eqref{eq:3D_distributed_model} and forward integrating via the chosen integration method to obtain the updated robot state. 
Properly setting the torsional damping is crucial to avoiding simulation instabilities. We tested several integrators of different orders and observed that the simulation becomes unstable when either the damping or the simulation time step $dT$ is increased. Additionally, although the Runge-Kutta integrator and backward Euler method are more computationally demanding, they did not appear to offer significant improvements in simulation stability.

\subsection{Nonuniform loading test}
\begin{figure}
  \centering
  \includegraphics[width=0.33\columnwidth]{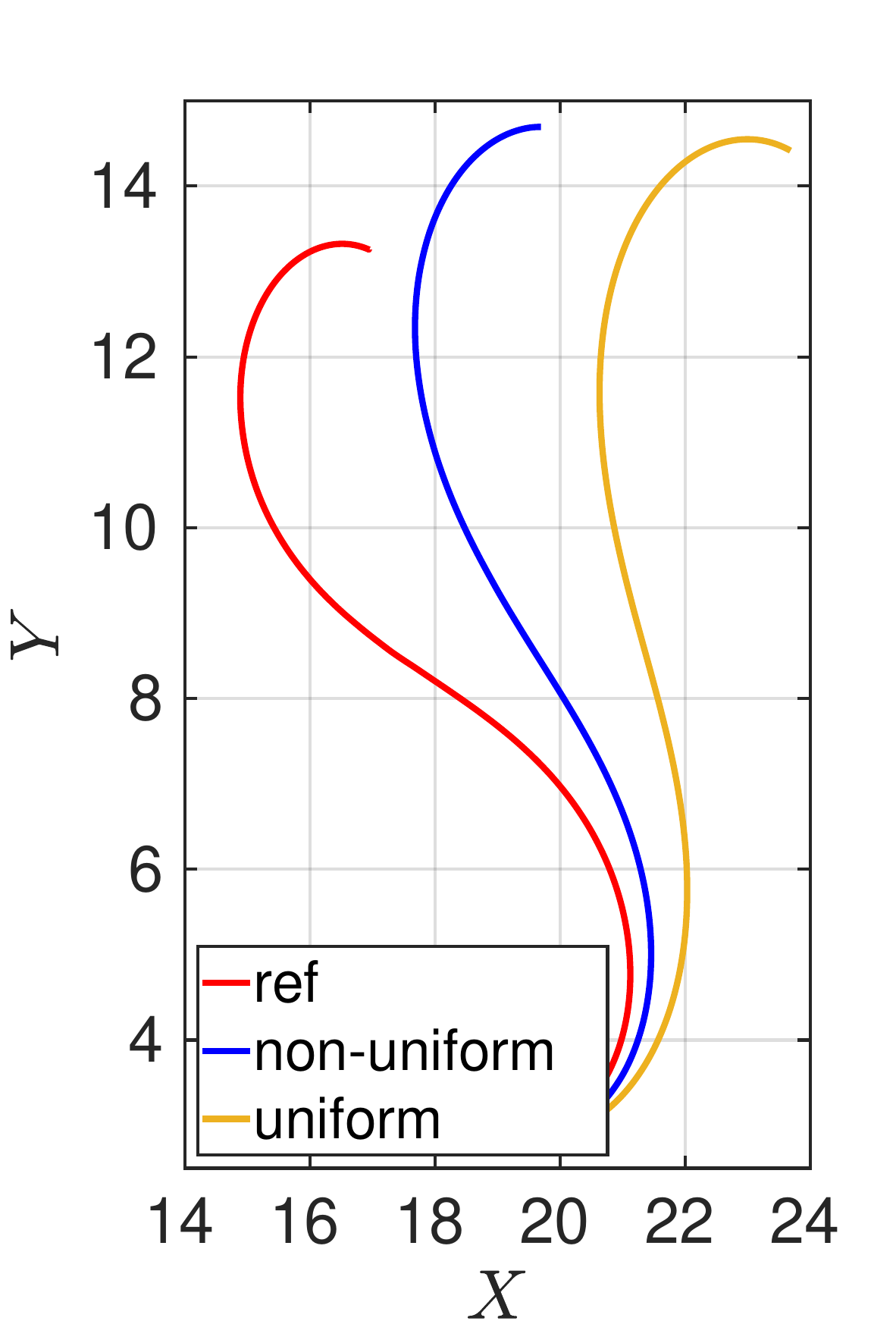}
  \includegraphics[width=0.66\columnwidth]{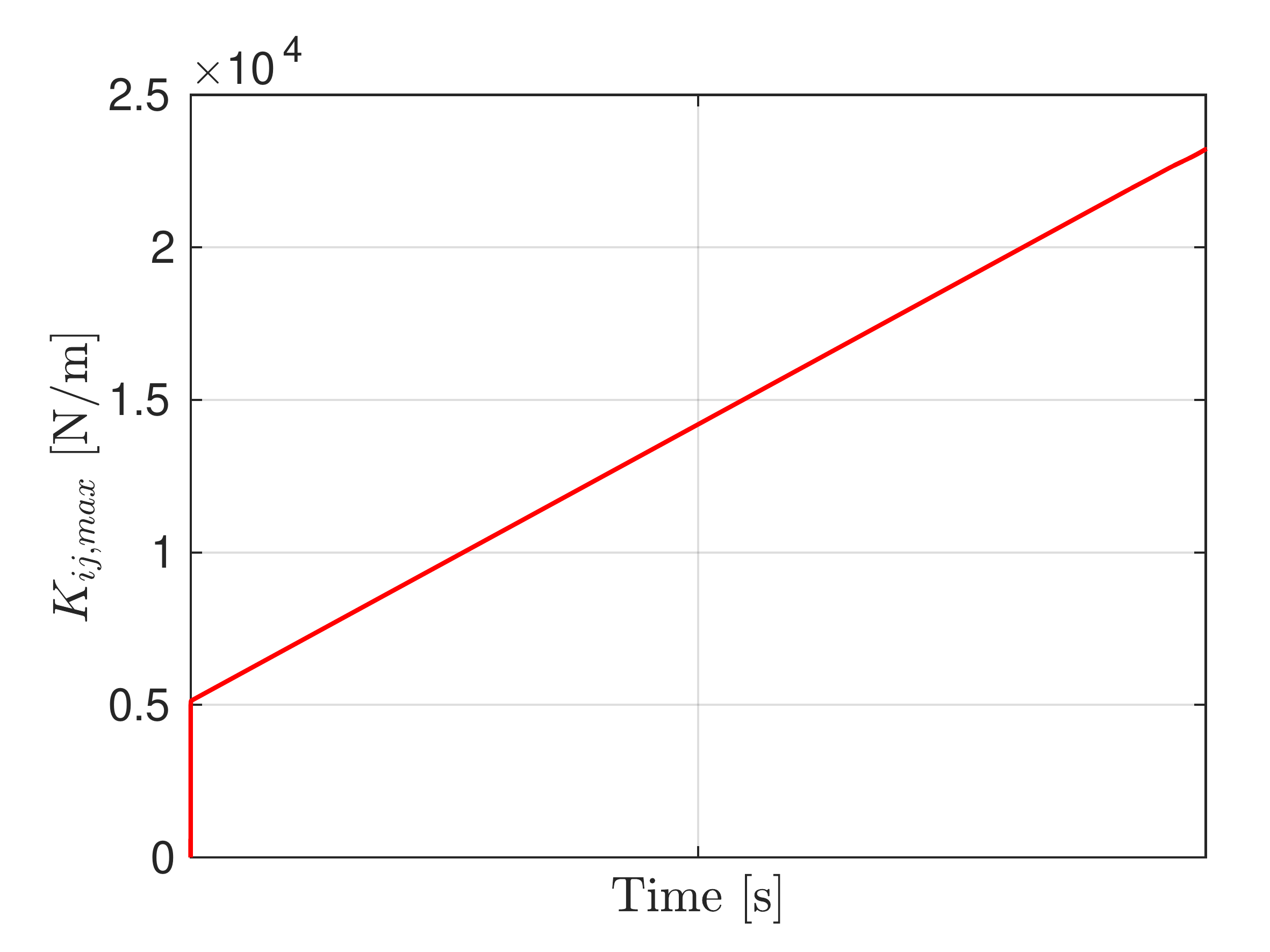}
  \caption{(left) Open loop simulation comparison of distributed model~\eqref{eq:3D_distributed_model} 
  with nonuniform (blue) and uniform (yellow)  load distribution. Desired  trajectory (red) is given just for reference.
  (right) Simulation of terrain speed-dependent stiffness variation. The red plot is the maximum value of 
  the stiffness for the patches of the left track as a function of desired robot speed ranging from 0 to 1.4 rad/s.}
  \label{fig:slopes_open_unif_nonunif_comparison}
\end{figure}

To appreciate the  simulator capability of modeling \textit{nonuniform} load distribution, in the accompanying video, we show simulations of the robot moving in open-loop on a
3D sloped terrain with stiffness $K_{t} = \SSI{100}{\kilo\newton\per\meter}$ and damping
$D_{t} = \SSI{50}{\kilo\newton\second\per\meter}$ and friction coefficient $\mu=0.6$. The sloped terrain has been randomly generated with Blender.
We give reference velocity commands to the robot generating a chicane reference as in \eqref{eq:open_loop_vel_reference} buth with higher longitudinal speed $v_{\max}^d=\SSI{0.8}{\meter\per\second}$ and angular velocity $\omega_{\max}^d=\SSI{0.6}{\radian\per\second}$ 
for $t_{\mathrm{end}} = \SSI{20}{\second}$ of simulation.
The forces $\vect{f}_{n,ij}$ on each terrain patch are highlighted with blue vectors applied 
at each patch location (see Fig.~\ref{fig:block_simulator}~(right)).
The resultant of the forces for each track is also  highlighted in red.
To avoid  the simulation getting too slow
a good compromise with accuracy has been found discretising the tracks into 10x4 patches. 
The accompanying video shows realistic loading behaviour according to the
relative orientation of the robot w.r.t. the slope: \ie{} the downhill part of the track gets more loaded while the uphill part gets unloaded. 
Despite the non negligible speed, gravity still dominates the loading and the effects due to centrifugal forces 
(\eg{} higher loading of the outer track in a turn) is not visible. 
The height $h$ of the \gls{com} influences load transfer, resulting in different track loading for the same robot orientation 
at different heights (not shown here for the sake of space).
We repeated the experiments also  in the assumption of \textit{uniform} load distribution and compared the results in 
Fig.~\ref{fig:slopes_open_unif_nonunif_comparison}(left). 
The plot shows a vertical difference of 4 m over a 15.6 m distance, corresponding to a cumulative error of 25.6\% when using the uniform load approximation compared to the non-uniform case. In terms of computational time, the simulation loop takes 0.49 ms on average (on a average of 2~$10^6$ calls) under the uniform load assumption, while it requires 11 ms—two orders of magnitude higher—for the non-uniform load case. The performance gain in the uniform load scenario stems from the fact that the terrain mesh is evaluated only once.
The tests have been performed on a  AMD Ryzen 7 7700X\SSI{4.5}{\giga\hertz} machine.
This highlights a fundamental trade-off between model fidelity and computational efficiency: while more accurate models can capture complex terrain-vehicle interactions more precisely, they may be infeasible for time-sensitive applications. On the other hand, simplified models—although less precise—can offer sufficiently accurate predictions at a fraction of the computational cost.

%

\subsection{Speed dependent stiffness test}
The stiffening behavior of the terrain with speed is visible only at high speed and/or soft terrain.
To show this effect we set a linearly increasing speed from 0 to
\SSI{1.4}{\meter\per\second} with the robot moving in a straight motion on a soft terrain with
nominal stiffness $K_t=\SSI{5000}{\newton\per\meter}$. We set a speed dependent stiffness gain $K_{t_p}=10.5$ $Ns/m^2$.
The video shows that the robot starts to pitch when speed increases due to the increased stiffness in the front of the tracks. The stiffness increase is highlighted by the color of the arrows representing the ground forces, with the color changing from blue towards red when going from higher to lower stiffness.
We report in Fig.~\ref{fig:slopes_open_unif_nonunif_comparison}~(right) how the  the maximum value of the  stiffness for the left track linearly changes with the vehicle speed, from the nominal value of \SSI{5000}{\newton\per\meter} for ${}_bv_x^d=\SSI{0}{\meter\per\second}$ to a higher value  of \SSI{23000}{\newton\per\meter} for ${}_bv_x = \SSI{1.4}{\meter\per\second}$.
\subsection{Closed loop tracking tests}
\begin{figure}[!ht]
  \centering
  \includegraphics[width=0.9\columnwidth]{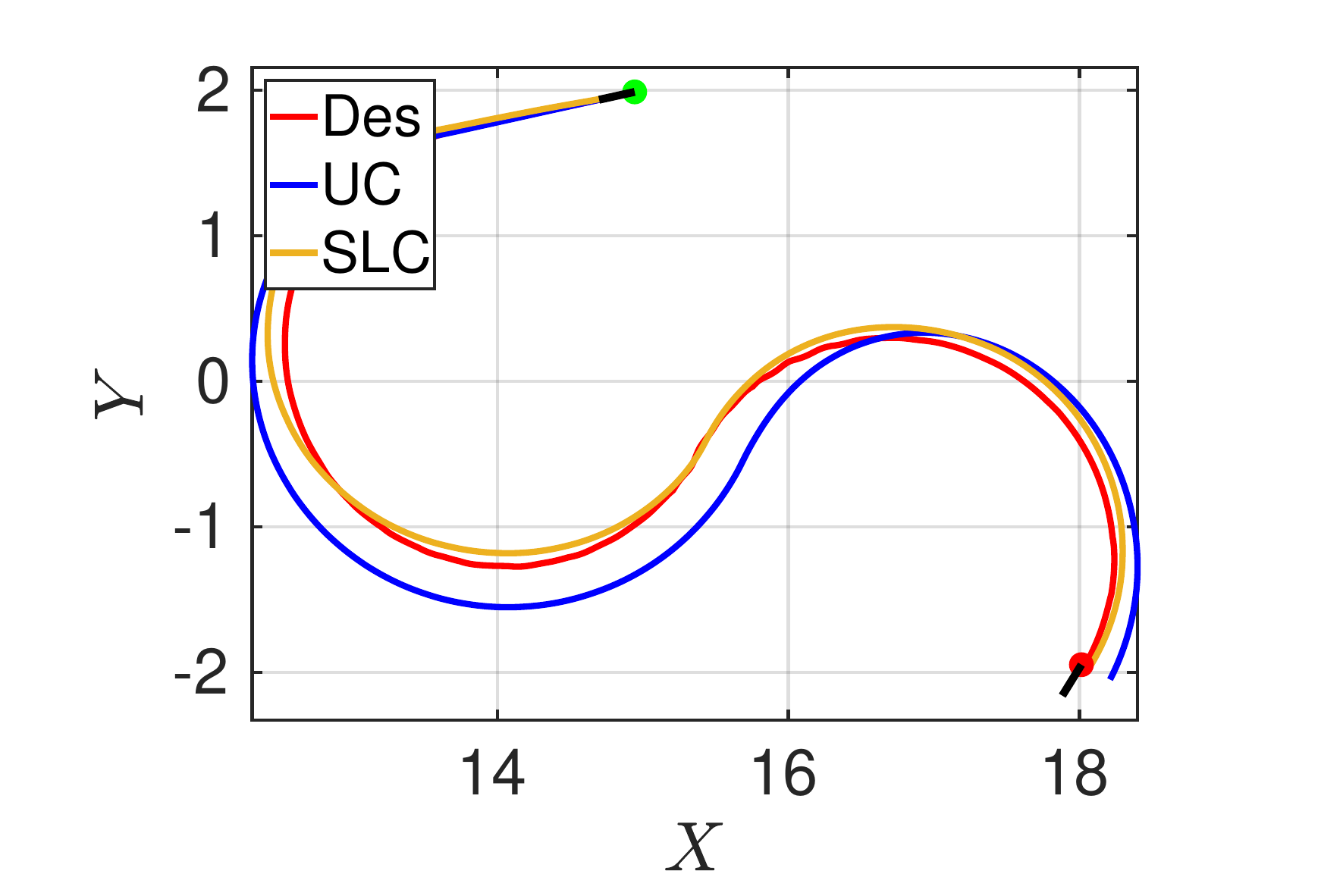}
  \caption{ Trajectory plot for the closed loop control of the chicane reference
        trajectory (red line) on slopes. Both the UC controller (blue line) and the SLC
        controller (yellow line) are reported. Friction coefficient is set to $\mu= 0.6$.
        Green and red dots are the initial and final configuration respectively.
        The initial and final orientation is highlighted with a black solid line.}
  \label{fig:slopes_cloop_xy}
\end{figure}

\begin{figure}[!ht]
  \centering
  \includegraphics[width=0.9\columnwidth]{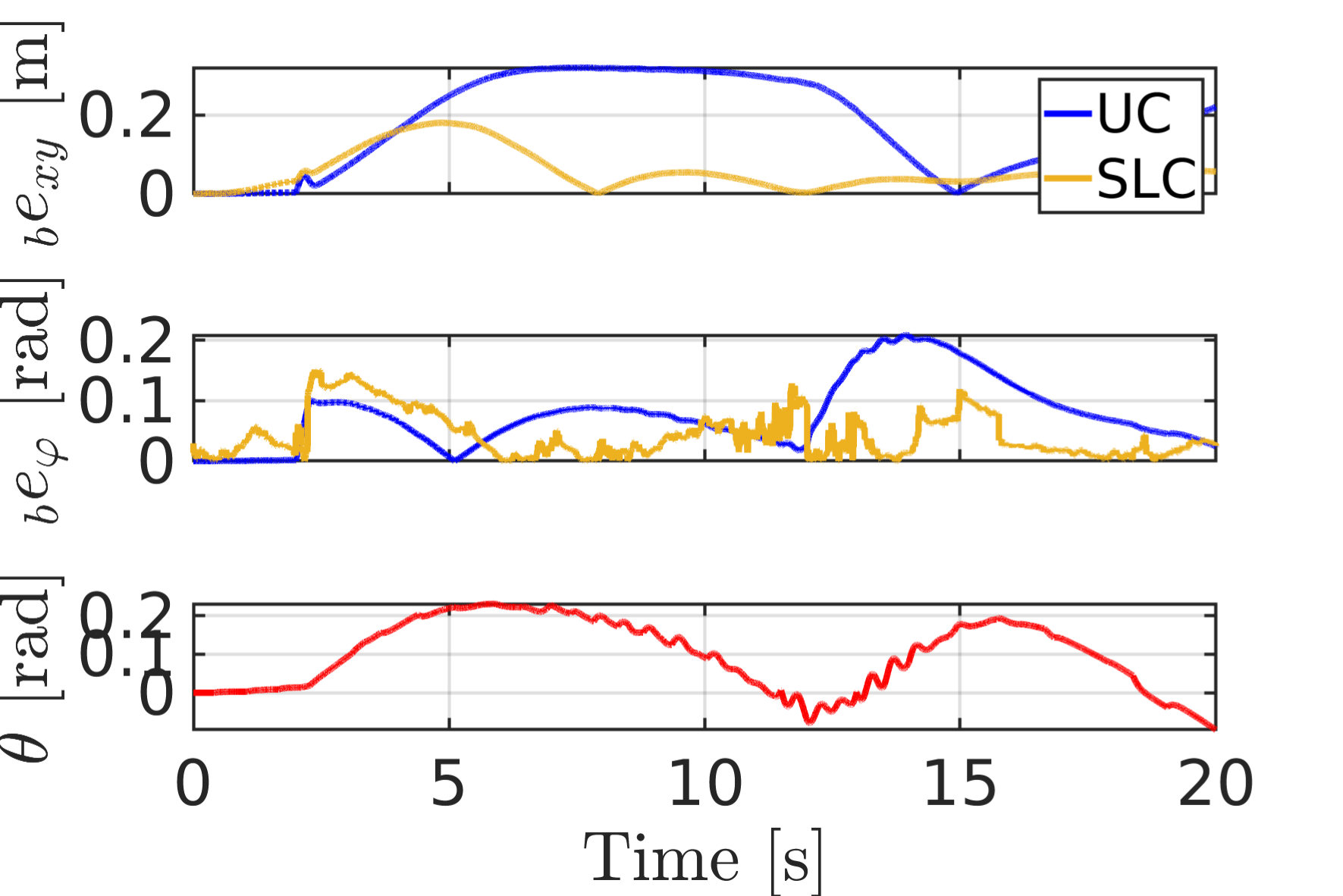}
  \caption{Plots of the tracking error (in local frame) for the closed loop control of the chicane reference 
        trajectory  \eqref{sec:user_def_reference} on slopes. (upper)  ${}_be_{xy}$ Cartesian error, (middle)  ${}_be_{\varphi}$ orientation error, for the UC (blue)   
        and SLC (yellow) controller. (bottom) Plot of the robot pitch representing the slope inclination.}
  \label{fig:slopes_cloop_errors}
\end{figure}

In this section (dual of Section \ref{sec:user_def_reference} for sloped terrain) we test in closed loop the slippage-aware controller  extended to the 3D case introduced in Section~\ref{sec:3D_control}, with function approximators~\eqref{eq:falpha_fbeta3D} trained for slopes. The reference to be tracked is the usual user-defined chicane reference trajectory~\eqref{eq:open_loop_vel_reference}, the simulator model is the one for nonflat terrain~\eqref{eq:3D_distributed_model}.
With respect to the flat terrain case we increased the friction coefficient from $\mu=0.1$ to $\mu=0.6$ to avoid catastrophic slippage. To have comparable slippages we adjsusted the speed parameters of the chicane to  $v_{\max}^d=\SSI{0.6}{\meter\per\second}$ and $\omega_{\max}^d=\SSI{0.4}{\radian\per\second}$, respectively.
The initial configuration is set to $\vect{c}_0 = [15, 2, -2.9]^\top$\,\USI{\meter}, a location where the robot is on a \textit{downward} slope. 
Fig.~\ref{fig:slopes_cloop_xy} is a bird-eye (XY) view showing of the reference trajectory (red) and the plots of the actual trajectories in the UC (blue) and SLC (yellow) case. 
The tracking results comparing the proposed slippage-aware controller SLC with the reference unicycle controller UC are reported in Fig.~\ref{fig:slopes_cloop_xy} and~\ref{fig:slopes_cloop_errors}.  Fig.~\ref{fig:slopes_cloop_errors} reports the plots of the metrics used to evaluate the controller performance, which in this case  are the tracking errors mapped in body coordinates ${}_be_{xy}$ and ${}_be_{\varphi}$  (upper and middle plot). 
The \textit{mean} and the \textit{final} Cartesian errors are both around \SSI{20}{\centi\meter} in the case of UC controller, while they are reduced to 5.9 cm and 5.6 cm, respectively, in the case of the SLC controller over a 13 m long trajectory. The bottom plot shows the robot  pitch $\theta$ indicating the  slope inclination is continuously changing, ranging from -0.1 to \SSI{0.2}{\radian}.
In the accompanying video, we also demonstrate an application of the SLC controller within a navigation
framework that generates planned trajectories to explore a sloped terrain and reach user-defined goals.

%% file: sections/8_conclusions.tex

\section{Conclusions}
\label{sec:conclusions}
In this paper, we have proposed a framework that enables reliable navigation of tracked vehicles across challenging terrains.
Our first contribution is the development of a simulation model that
provides a physically accurate representation of ground contact
forces, both on flat and sloped terrains. This model offers a faithful
digital twin of the system. It has been validated against experimental
data and is suitable for realistic simulations across a wide range of
operational scenarios. However, the complexity of this model
discourages its direct use in control design.

To address this, we have also developed a simpler pseudo-kinematic
model, which condenses the sophisticated nuances of terramechanics
interaction into a small set of parameters. This control-oriented
model accounts for both longitudinal and lateral slippage. We have
devised an effective machine learning procedure to estimate these
parameters, and its accuracy has been confirmed through experimental
validation.
Our analysis of the simulation model suggests that these parameters
are directly influenced by the velocities of the tracks. This insight
has been pivotal to our second and third contributions: a package of
algorithmic solutions for trajectory control and path planning.

For trajectory control, we have proposed a Lyapunov-based controller,
extended with feed-forward compensation. Our analysis reveals that
this controller effectively mitigates slippage effects with minimal
computational overhead, thereby enhancing the tracking capabilities of
tracked vehicles. 
\DSrev{The superior tracking performance, relative to a state-of-the-art unicycle controller, is validated through real-world indoor experiments conducted with a LIMO tracked robot operating on both a parquet floor and a low-friction soapy surface.}

The application of our control strategy allows the closed-loop system
to approximate unicycle kinematics, paving the way for the adoption of
simple and efficient planning algorithms based on Dubins
manoeuvres~\cite{Laumond1998}. Equally straightforward is the use of a
planning solution based on the analytical interpolation of $G^1$
curves.

While these solutions offer simplicity and numerical efficiency, they
do not necessarily ensure high levels of accuracy or optimality. For
applications where such performance is a stringent requirement, we
have also proposed a motion planning approach based on numerical
optimisation, which can be warm-started using the Dubins-based
solution.

To conclude, we evaluated two families of planning strategies. The first comprises approximate methods that rapidly generate reasonable trajectories without feasibility guarantees, namely, they do not account for slippage effects or actuation limits. As a result, the responsibility of compensating for these shortcomings is shifted to the controller. These strategies are widely used in the literature~\cite{Laumond1998} and are typically based on simplified models. For instance, the Dubins model assumes a virtual unicycle with no constraints on angular velocity, allowing for discontinuous angular acceleration. Slightly more advanced models, such as those based on Clothoids, enforce a linear variation in curvature, better approximating car-like behavior with continuous steering dynamics.
The second family consists of optimal control approaches, which leverage the proposed pseudo-kinematic model to compute feasible, high-quality trajectories that explicitly account for physical constraints. While this results in significantly more accurate and consistent plans, it comes at the cost of increased computational time. This can be  reduced accepting less accurate function approximators based on exponential parameterizations to estimate the slippage parameters.
The analysis in Section~\ref{sec:plan_eval} shows that Clothoid-based planners can achieve satisfactory performance but remain suboptimal compared to more sophisticated planner based on numerical optimization. This highlights an inherent trade-off between the level of optimality and the computational reactivity of the planner, allowing users to choose the most appropriate method based on their application's requirements.

\paragraph{Future work}

This research has provided valuable insights into the dynamics of
tracked vehicles, laying the groundwork for future investigations. A
first important direction for our forthcoming research is a broader
experimental validation of the models across different types of
vehicles and a wide range of terrains, weather conditions, and
slopes. Such conditions are difficult to replicate in a laboratory
setting and require a sophisticated and portable setup for data
collection in the wild. \DSrev{To achieve this objective, we aim to implement relative visual 
odometry techniques as an alternative to the Motion Capture System, thereby enabling 
accurate motion estimation without external tracking infrastructure and enhancing 
the system’s autonomy by reducing dependency on external positioning equipment.}

A second research direction is the investigation of trajectory control
algorithms capable of delivering high levels of performance for
high-dynamics tracked vehicles. The experimental data we have
collected align well with our theoretical findings, but were obtained
using a heavy and slow platform, suitable for typical agricultural
applications.
Additionally, we are exploring new motion planning solutions that can
offer a better trade-off between performance and execution time than
the alternatives considered in this paper.

Finally, we plan to apply visual foundation models to estimate ground
parameters for our model, following the approach of the Magic-VFM
framework~\cite{lupu2025magicvfm}.

%% file: sections/9_appendix.tex
\subsection{Derivative of a vector in a moving frame}
\label{sec:derivative_in_moving_frame}
When computing the dynamics in \eqref{eq:2D_distributed_model} and \eqref{eq:3D_distributed_model} in the body frame, the effect of the body's rotation must be taken into account in the acceleration computation. Consider two coordinate systems: the inertial frame $\mathcal{W}$, and the body frame $\mathcal{B}$ which is rotating w.r.t. $\mathcal{W}$ with angular velocity $\boldsymbol{\omega}$. 
The components of a vector $\vect{v}$ expressed in the inertial frame $\mathcal{W}$ are  linked to the components ${}_b\vect{v}$ of the same vector expressed in the body frame by
\begin{equation}
    \vect{v} = {}_w\vect{R}_b~{}_b\vect{v} 
    \label{eq:inertial}
\end{equation}
where ${}_w\vect{R}_b$ is the rotation matrix representing the orientation of   $\mathcal{B}$ w.r.t.  $\mathcal{W}$. 
The derivative of eq. \eqref{eq:inertial} is:
\begin{equation}
   \frac{d\vect{v}}{dt} = {}_w\vect{R}_b \frac{d{}_b\vect{v} }{dt}    + {}_w\dot{\vect{R}}_b {}_b\vect{v}
    \label{eq:inertial_der1}
\end{equation}
\noindent Now, from mechanics, is well known that
\begin{equation}
    {}_w\dot{\vect{R}}_b = \boldsymbol{\omega} \times {}_w\vect{R}_b = \vect{S}(\boldsymbol{\omega})  {}_w\vect{R}_b
    \label{eq:inertial_der2}
\end{equation}
\noindent where $\vect{S}(.)$ is the skew-symmetric matrix  associated to the  cross product.  
By simple algebraic passages, exploiting the property $\vect{R} \vect{S}(\vect{x}) \vect{R}^T = \vect{S}(\vect{R}\vect{x})$ of skew symmetric matrices, we get
\begin{align}
\begin{split}
   {}_w\dot{\vect{R}}_b &=  \vect{S}( {}_w\vect{R}_b~ {}_b\boldsymbol{\omega})~ {}_w\vect{R}_b\\
                                   &=  {}_w\vect{R}_b ~\vect{S}(  {}_b\boldsymbol{\omega}) {}_w\vect{R}_b^T ~ {}_w\vect{R}_b\\
                                   &= {}_w\vect{R}_b ~\vect{S}(  {}_b\boldsymbol{\omega})   
\end{split}
\label{eq:inertial_der3}
\end{align}
\noindent by substituting \eqref{eq:inertial_der3} in \eqref{eq:inertial_der1} we obtain
\begin{align}
  \left( \frac{d\vect{v}}{dt}\right)_{\text{in}} &= {}_w\vect{R}_b \frac{d{}_b\vect{v} }{dt}    + {}_w\vect{R}_b {}_b ~\boldsymbol{\omega} \times  {}_b\vect{v}  
    \label{eq:inertial_der4}
\end{align}
\noindent where 
\begin{align}
    \vect{{}_b\boldsymbol{\omega}} = \mat{ 0 & -{}_b\omega_z &  {}_b\omega_y \\  
                              {}_b\omega_z  & 0 &  -{}_b\omega_x \\
                              -{}_b\omega_y& {}_b\omega_x &  0 }\nonumber    
\end{align}
if $\vect{v}$ is the body velocity then $\left( \frac{d\vect{v}}{dt}\right)_{\text{in}} $ is called absolute or total acceleration,  where the first term of \eqref{eq:inertial_der4} is due to the translational motion while  the  second term is  due to the rotational velocity of the frame $\mathcal{B}$. Now, by  expressing everything in the body (moving) frame we get:
\begin{align}
\left( {}_b\dot{\vect{v}}\right)_{\text{in}} =  {}_w\vect{R}_b^T   \left( \frac{d \vect{v}}{dt}\right)_{\text{in}}&=  {}_b\dot{\vect{v}}     +  {}_b\boldsymbol{\omega} \times  {}_b\vect{v}  
    \label{eq:inertial_der5}
\end{align}
\noindent Where we have the linear terms that appear in \eqref{eq:3D_distributed_model}, a similar reasoning applies to the angular terms. In the flat terrain case, by explicitly writing the cross product, setting ${}_bv_z = 0$, and considering only the $x$ and $y$ components, we recover the terms shown in Eq. \eqref{eq:2D_distributed_model}.

\begin{align}
\begin{split}
  ({}_b\dot{v}_x )_{\text{in}} &= {}_b\dot{v}_x     - \omega_z    {}_bv_y \\
  ({}_b\dot{v}_y )_{\text{in}} &= {}_b\dot{v}_y     + \omega_z   {}_bv_x  
\label{eq:inertial_der6}
\end{split}
\end{align}